\documentclass[final]{article}

\usepackage{soul}
\usepackage{graphicx}
\usepackage[linesnumbered]{algorithm2e}
\usepackage{todonotes}
\usepackage{url}
\usepackage{booktabs}
\usepackage{amsfonts}
\usepackage{caption}
\usepackage{subcaption}
\usepackage{mathbbol}
\usepackage{multirow}
\usepackage{graphicx, array, blindtext}
\usepackage{amsmath}
\usepackage{sidecap}
\usepackage{sidecap}
\usepackage{caption,subcaption}
\usepackage{pbox}
\usepackage{arydshln}
\newtheorem{proposition}{Proposition}

\DeclareMathOperator*{\argmin}{arg\,min}
\def\defterm#1{\textbf{#1}}
\def\bs#1{\boldsymbol{#1}}
\def\set#1{\bs{#1}}

\def\reals{\mathbb{R}}

\newcommand{\graph}{\G}
\newcommand{\gengraph}{\hat{\graph}}

\newcommand{\numnodes}{N}
\newcommand{\unode}{i}
\newcommand{\vnode}{j}

\newcommand{\xinstance}{\set{X}}

\newcommand{\gprob}{p}
\newcommand{\prior}{p}
\newcommand{\linstance}{\set{z}}
\newcommand{\ldim}{t}

\newcommand{\ndim}{d} 

\newcommand{\parameter}{\theta}

\newcommand{\ourModelAcronym}{-MM}
\newcommand{\nodeNum}{N}
\newcommand{\nodem}{i}
\newcommand{\noden}{j}
\newcommand{\G}{G}
\newcommand{\A}{\set{A}} 
\newcommand{\D}{D} 
\newcommand{\X}{\set{X}} 

\newcommand{\prob}[1]{\Tilde{#1}}

\newcommand{\V}{V} 
\newcommand{\dhk}{\textbf{d}}

\newcommand{\eparameters}{\varphi} 
\newcommand{\dparameters}{\set{\psi}} 

\newcommand{\kthFeature}{u}
\newcommand{\Featuredim}{l}
\newcommand{\featureNUM}{m} 
\newcommand{\kweight}{\sigma}

\newcommand{\parameters}{\set{\parameter}}
\newcommand{\elbo}{\textsc{ELBO}~}
\newcommand{\featureFunction}{\phi}
\newcommand{\ofeature}{\set{F}}

 \newcommand{\training}{\A}
 \newcommand{\localloss}{\mathcal{L}^{0}}
 \newcommand{\globalloss}{\mathcal{L}^{1}}
 \newcommand{\tradeoff}{\gamma}
 \newcommand{\samplesize}{n}
 \newcommand{\sindex}{j}
 \newcommand{\mse}{\mathit{MSE}}

\newcommand{\embedder}{\mathcal{E}}

\usepackage[utf8]{inputenc} 
\usepackage[T1]{fontenc}    
\usepackage{hyperref}       
\usepackage{url}            
\usepackage{booktabs}       
\usepackage{amsfonts}       
\usepackage{nicefrac}       
\usepackage{microtype}      
\PassOptionsToPackage{numbers, compress}{natbib}
\usepackage{neurips_2022}

\title{Micro and Macro Level Graph Modeling for Graph Variational Auto-Encoders}
\author{%
  Kiarash Zahirnia, Oliver Schulte\thanks{Supported by NSERC Canada Discovery Grant R611341}, Parmis Naddaf, Ke Li\\
 School of Computing Science, Simon Fraser University, Canada\\
  \texttt{\{kzahirni, oschulte, pnaddaf, keli\}}@sfu.ca \\
}

\begin{document}

\maketitle

\begin{abstract}
Generative models for graph data are an important research topic in machine learning. Graph data comprise two levels that are typically analyzed separately: node-level properties such as the existence of a link between a pair of nodes, and global aggregate graph-level statistics, such as motif counts.
This paper proposes a new multi-level framework that {\em jointly} models node-level properties and graph-level statistics, as mutually reinforcing sources of information.  
We introduce a new {\em micro-macro} training objective for graph generation
that combines node-level and graph-level losses. 
We utilize the micro-macro objective to improve graph generation with a GraphVAE
, 
a well-established model based on graph-level latent variables, that provides fast training and generation time for medium-sized graphs. Our experiments show that adding micro-macro modeling to the GraphVAE model improves graph quality scores up to 2 orders of magnitude on five benchmark datasets, 
while maintaining the GraphVAE generation speed advantage.
\end{abstract}

\section{Introduction: multi-level graph modeling}\label{Introduction}
\begin{figure}[h]
     \centering
     \begin{subfigure}[b]{.3\textwidth}
         \centering
         \includegraphics[width=\textwidth]{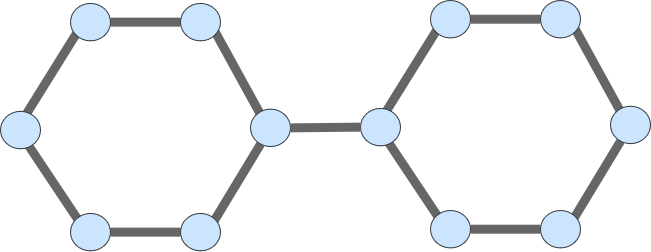}
                  \caption{Original Graph}
         \label{fig:original}
     \end{subfigure}
     \hfill
     \begin{subfigure}[b]{0.3\textwidth}
         \centering
         \includegraphics[width=\textwidth]{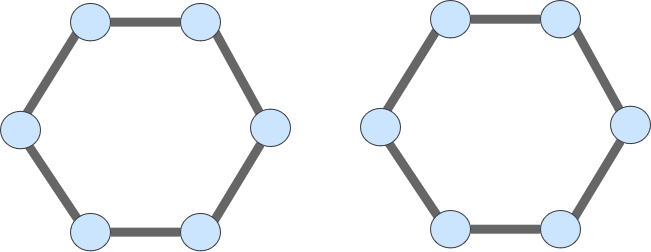}
                  \caption{Generated Graph 1}
         \label{fig:sample1}
     \end{subfigure}
     \hfill
     \begin{subfigure}[b]{0.3\textwidth}
         \centering
         \includegraphics[width=\textwidth]{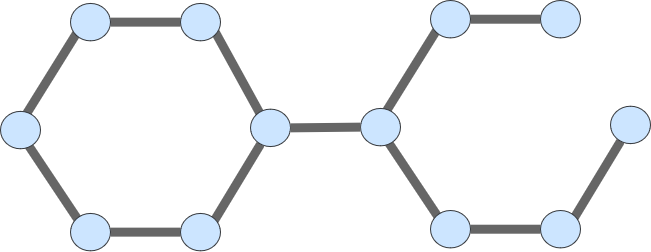}
                  \caption{Generated Graph 2}
         \label{fig:sample2}
     \end{subfigure}
     \caption{To illustrate the difference between local and global properties. The two right graphs~(\ref{fig:sample1}),~(\ref{fig:sample2}) score the same in terms of number of edges reconstructed from the left graph~(\ref{fig:original}), a local node-level property.  
    However the right graph~(\ref{fig:sample2}), is structurally more similar to~(\ref{fig:original}), containing the same number of connected components, a global graph-level statistic.}
      \label{fig:crossentropyExample}
\end{figure}

Many datasets contain relational information about entities and their links that can be represented as a graph. 
Deep generative learning on graphs has become a popular research topic~\cite{hamilton2020graph}, with applications including molecule design~\cite{DBLP:journals/jmlr/SamantaDJGCGG20}, and recommendation~\cite{do2019matrix}. It is common in graph analysis to distinguish two levels of information~\cite{bib:chakrabarti2012graph, hamilton2020graph}:
(1) local node-level properties, such as the existence of a link between two nodes or the attribute of a node~\cite{kipf2016variational,arga,graphite}, and (2) global graph-level statistics (graph statistics for short) that depend on the entire graph, such as 
node degree distribution or motif counts.
Most deep Graph Generative Models (GGMs) are trained with an objective based on local properties (e.g., maximizing the probabilities of  observed edges).  Figure \ref{fig:crossentropyExample} illustrates the difference between local and global structure. 
Different edges have different roles in the graph global structure. Some edges play a critical role for the connectivity/community structure, while others are less important. 
Previous GGM training is generally based on a likelihood objective that 
decomposes into individual edge likelihoods  \cite{you2018graphrnn,liao2019efficient,dai2020scalable}, which does not discriminate edges with different roles.
Graph global level information on the other hand can capture how different edges have different importance  
in the graph structure (see also Section \ref{sec:kernels}).  Whereas the sparsity of typical graphs causes difficulties for training objectives based on local information only~\cite{qiao2020novel,kipf2016variational}, graph statistics are typically dense. 

This paper proposes a new perspective on the node/graph level dichotomy: a principled probabilistic framework that incorporates both local  and global graph properties. From the framework we derive a
novel micro-macro (MM) objective function for training a GGM to match {\em both} local and global properties. The term {\em micro-macro} originates from the philosophy of science and refers to scientific frameworks that treat local (``micro'') resp. global (``macro'') properties of a system as equal  targets~\cite{sep-models-science}.
\paragraph{Advantages} Micro-macro modeling increases {\em graph realism} and {\em user control.} (1) Compared to 
objective functions 
that are based on predicting local properties, matching graph statistics serves as a {\em regularizer} 
that increases the realism of the generated graph structures. 
This is sufficient to generate realistic graphs with a fast all-at-once edge-parallel model.
(2) 
\citet{o2021evaluation} note that different graph statistics are important for different applications. For example for a large payment graph recording economic transactions, a macro economist may be mainly interested in the average price level of a goods basket. For a central bank managing a payment system, the total number of transactions may be more important.
\citet{o2021evaluation} therefore advise selecting a GGM based on the graph statistics of interest in the target application. However,  
this entails searching through the space of GGMs and their hyperparameters to find a good match with target graph statistics. In our MM framework, the user only needs to specify the target graph statistics and learning will automatically select graph models that match them.

\paragraph{Approach} We introduce a joint probabilistic model over both local properties and global graph statistics. 
An MM objective that can be used with graph encoder-decoder models is derived as an \elbo from the joint probability model. Our experiments focus on the GraphVAE (Graph Variational Auto-Encoder) model where the encoder outputs a graph-level latent posterior $\linstance$ (graph embedding) and the decoder maps the graph embedding $\linstance$ to a soft adjacency matrix representing link probabilities~\cite{DBLP:conf/icann/SimonovskyK18}. A graph-level embedding supports modeling graph-level statistics. We show how the recent   calibrated Gaussian variational auto-encoder \cite{rybkin2021simple} can be adapted to model graph statistics with divergent scales. The implementation and datasets are provided at 
\url{https://github.com/kiarashza/GraphVAE-MM}.
%

\paragraph{Evaluation}

Evaluating GGMs has become a research topic of its own~\cite{thompson2022evaluation, o2021evaluation}; see the related work, section below. Proposed metrics quantify how similar a set of generated graphs is to a set of observed graphs. Our assessment focuses on evaluation metrics based on Graph Neural Networks
(GNN-based metrics)
~\cite{thompson2022evaluation}. 
To our knowledge, this is the most recent published evaluation method, with state-of-the-art (SOTA) performance shown in extensive experiments. For GraphVAEs, adding micro-macro modeling improves both the realism and the diversity 
of the generated graphs, up to an order of magnitude on GNN-based metrics. This is sufficient to reach very competitive graph quality compared to SOTA auto-regressive benchmark models~\cite{DBLP:conf/nips/LiaoLSWHDUZ19,you2018graphrnn,dai2020scalable}. At the same time, one-shot graph generation with GraphVAEs is faster than sampling from auto-regressive models. Our experiments on medium-sized graphs show that even with the overhead of micro-macro modeling, GraphVAEs maintain their generation speed advantage over auto-regressive approaches. While our evaluation focuses on  GraphVAEs, we discuss how micro-macro modeling can be applied to other GGMs to leverage graph-level statistics.


{\em Contributions.} Our main contributions can be summarized as follows. 

\begin{itemize}
\item A new joint probabilistic model over both local graph properties and global graph-level statistics.
\item Deriving a joint \elbo as a new micro-macro objective function for training graph encoder-decoder models.
\item An adaptation of the GraphVAE architecture \cite{DBLP:conf/icann/SimonovskyK18} to learn graph embeddings using the joint micro-macro objective.
\end{itemize}

 \section{Related work} \label{sec:related}

 
 {\em Graph Statistics as Modeling Targets.} To our knowledge, this is the first work to develop a GGM with the objective of matching graph statistics. 
Work on graph moment matching in network statistics ~\cite{orbanz2014bayesian, bickel2011method}  has studied theoretical properties of graph moment estimators with increasing graph size. Maximizing the likelihood of exponential random graph models~\cite{harris2013introduction} such as Markov Logic Networks is equivalent to maximizing entropy subject to matching a set of graph statistics~\cite{kuvzelka2018relational}. 
 
 
 {\em Graph Statistics for Evaluating Generative Models.} The ability of generative models to match an empirical graph statistic distribution has been assessed in several research papers, which 
  supports our approach of including them in the training objective. \citet{farnadi2017soft} compare  
 relational models in terms of matching (expected) motif counts.  ~\citet{you2018graphrnn} and ~\citet{DBLP:conf/nips/LiaoLSWHDUZ19} use 4 graph statistics to measure the realism of generated graphs. 
 
The question of how to evaluate a GGM has been studied in recent papers~\cite{o2021evaluation,thompson2022evaluation}. An evaluation metric quantifies how similar a set of generated graphs $\gengraph_{1},\ldots,\gengraph_{m_1}$ is to a set of test graphs $\graph_{1},\ldots,\graph_{m_2}$. 
 From the graphs we can compute corresponding descriptors (e.g., statistics) $\hat{x}_{1},\ldots,\hat{x}_{m_1}$ and $x_{1},\ldots,x_{m_2}$. The similarity of two descriptor sets can be quantified using Maximum Mean Distance (MMD).  
 \citet{o2021evaluation}  raise several difficulties with this approach. 
 (1) Given a list of graph statistics, we obtain a list of MMD scores, rather than a single score that can be used to rank GGMs.  
 (2) The statistics-based MMD scores are sensitive to hyper-parameters; different settings can lead to different model rankings even for a single graph statistic. (3) In perturbation studies where observed graphs are corrupted in a controlled manner, MMD scores do not correlate well with degree of perturbation.  \citet{thompson2022evaluation} develop a recent proposal to address these issues. They utilize a {\em reference embedding network} GNN $\embedder$. The embedder $\embedder$ is obtained from pretraining or from random weights and is therefore independent of any of the models under evaluation. Given $m_1$ generated graphs and $m_2$ test graphs, $\embedder$ provides embeddings $\hat{e}_{1},\ldots,\hat{e}_{m_1}$ and $e_{1},\ldots,e_{m_2}$  which can be compared using a  vector metric such as MMD. 
 Extensive evaluation shows that the GGM scores provided by the reference embedding approach correlate well with perturbation degree and capture the realism and diversity of generated graphs. To our knowledge, the 
 GNN-based approach is the state of the art for evaluating GGMs, so {\em we use it as our main evaluation metric.}

{\em Graph Generation Architecture.} Our novel contribution is to develop a new kind of objective function, not a new kind of GGM. We therefore follow an AB design where we utilize an existing architecture as is and change only the training objective. Developing GGMs is an on-going topic of research; overviews can be found in ~\cite{zhou2020graph,hamilton2020graph}. 
Two major GGM groups are all-at-once and auto-regressive methods~\cite{hamilton2020graph}. 
All-at-once generation decodes a latent variable  $\linstance$ to generate a (soft) adjacency matrix $\prob{\A}$. Auto-regressive methods~\cite{DBLP:conf/nips/LiaoLSWHDUZ19,you2018graphrnn,khademi2020deep} generate a graph incrementally. All-at-once methods are faster at graph generation, but tend to generate less realistic graphs. They can use node-level representations 
 or a graph-level representation (embedding).
We use a graph-level representation for two reasons: (1) A natural fit with modeling graph-level statistics. (2) It is known from previous work that node-level representations, while useful for many applications, do not capture enough graph structure to generate realistic graphs.
%
We utilize the well-known GraphVAE 
architecture for computing graph embeddings~\cite{DBLP:conf/icann/SimonovskyK18},\cite[Ch.9 Sec.1.2]{hamilton2020graph}.

For a generative model of graph statistics, we adapt  calibrated Gaussian variational auto-encoder \cite{rybkin2021simple}, which has been developed as a generative model for i.i.d. data, but not previously been applied to graph modeling.

\section{Data model and micro-macro objective}\label{theory}
An attributed graph is a pair $\G=(V,E)$ comprising a finite set of $\numnodes$ nodes and edges where each node  is assigned an $\ndim$-dimensional attribute $\xinstance_{\unode}$. An attributed graph can be represented by an $\numnodes\times \numnodes$ adjacency matrix $\A$ with $\{0,1\}$ entries, together with an $\numnodes \times \ndim$ node feature matrix $\X$. 
Following \citet{DBLP:conf/nips/MaCX18}, we view the observed adjacency matrix as a sample from an underlying probabilistic adjacency matrix $\prob{\A}$ with $\prob{\A}_{\nodem,\noden} \in [0,1]$. The sampling distribution for independent edges  is given by
\begin{equation} \label{eq:sample-edges}
    p(\A|\prob{\A}) = 
    \prod_{\nodem=1}^{\numnodes} \prod_{\noden=1}^{\numnodes} \prob{\A}_{\nodem,\noden}^{\A_{\nodem,\noden}} (1-\prob{\A}_{\nodem,\noden})^{1-\A_{\nodem,\noden}}.
\end{equation}
%
A \defterm{descriptor function} $\featureFunction$ maps an adjacency matrix $\A$ to a $\Featuredim$-dimensional \defterm{graph statistic} such that $\featureFunction(\A)  \in \reals^{\Featuredim}$~\cite{o2021evaluation}. We use also matrices as graph statistics. We consider only descriptor functions that are permutation-equivariant~\cite[Ch.5]{hamilton2020graph}.
A graph statistic represents a higher-order graph property.
Examples include a node degree histogram or motif counts; for discussion and further examples see~\cite[Sec.2.1.2]{hamilton2020graph} and Section~\ref{sec:kernels} below.

\subsection{Micro-macro objective} 

We assume a finite list of descriptor functions $\featureFunction_1,\ldots,\featureFunction_{\featureNUM}$ that define \defterm{target statistics}. 
For a fixed attribute matrix $\X$, a \defterm{micro-macro (MM) loss} is of the form  
\begin{equation*} \label{eq:micro-macro}
 \mathcal{L}_{\parameters}(\training) =
 \localloss_{\parameters}(\training) + \tradeoff \globalloss_{\parameters}(\ofeature_{1},\ldots,\ofeature_{\featureNUM})
 \end{equation*}
 
 where $\training$ is the training graph and $\ofeature_{\kthFeature},\kthFeature=1,\ldots,\featureNUM$ is a random variable defined by applying the $\kthFeature$-th descriptor function to the training graph. The hyperparameter $\tradeoff$ controls the balance between capturing micro and macro aspects of the training graph. The micro loss $\localloss$ decomposes into losses for each node or pair of nodes. Well-known examples include
 the cross-entropy loss and max-margin loss. 
 In this paper we work with negative log-likelihood losses and a variational approximation based on an encoder-decoder architecture:
   \begin{align}
  \localloss_{\dparameters}(\training) = - \ln{\gprob^{0}_{\dparameters}}(\training) =\textcolor{black}{ -\ln{\int p(\training|\prob{\training}_{\linstance}) \prior(
  \linstance) d\linstance} \label{eq:local-loss}}\\
\globalloss_{\dparameters,\set{\kweight}}(\ofeature_{1},\ldots,\ofeature_{\featureNUM}) = - \sum_{\kthFeature=1}^{\featureNUM} \frac{1}{|\ofeature_{\kthFeature}|} \ln{\gprob^{1}_{\dparameters,\set{\kweight}}(\ofeature_{\kthFeature})} 
\\
\gprob^{1}_{\dparameters,\set{\kweight}}(\ofeature_{\kthFeature}) =
\int \mathcal{N}(\ofeature_{\kthFeature}|\featureFunction_{\kthFeature}(\prob{\training}_{\linstance}),\kweight_{\kthFeature}^{2} I) \prior(\linstance) d\linstance \label{eq:statistics-model}
 \end{align}
 %
 using the following notation.
 \begin{itemize}
    \item  $\linstance_{1 \times \ldim}$ specifies a latent $\ldim$-dimensional graph embedding with prior distribution $p(\linstance)$.
    \item $\prob{\training}_{\linstance}$ is a probabilistic adjacency matrix computed as a trainable deterministic decoder function of graph embedding $\linstance$. The decoder parameters are denoted as $\dparameters$. 
    The edge reconstruction probability $p(\training|\prob{\training}_{\linstance})$ is computed as in Equation~\eqref{eq:sample-edges}.
    \item The conditional distribution of each graph statistic is modeled as a Gaussian $\mathcal{N}$ with diagonal variance parameter $\kweight_{\kthFeature}^{2}$:
     \[\gprob^{1}_{\dparameters,\set{\kweight}}(\ofeature_{\kthFeature}|\linstance) = \mathcal{N}(\ofeature_{\kthFeature}|\featureFunction_{\kthFeature}(\prob{\training}_{\linstance}),\kweight_{\kthFeature}^{2} I).\]  The Gaussian mean is computed by applying the $\kthFeature$-th descriptor function to the reconstructed (soft) adjacency matrix.
     Even with a Gaussian conditional distribution, the {\em marginal} distribution over graph statistics, $\gprob^{1}_{\dparameters,\set{\kweight}}(\ofeature_{\kthFeature})$, can in principle fit any distribution \cite{DBLP:journals/corr/KingmaW13}.
    \item $|\ofeature_{\kthFeature}|$ is the dimensionality of target statistic $\ofeature_{\kthFeature}$. 
    
 \end{itemize}
 
Figure~\ref{fig:PGM} shows a generative model diagram for Equations~\eqref{eq:local-loss}--\eqref{eq:statistics-model}. 
Normalizing each graph statistic by its dimension is important because their scales can diverge widely. For example, transition matrices have $\numnodes^2$ entries, while the number of triangles is a single scalar. The following proposition provides an \elbo for an MM loss.
 
  \begin{proposition}
 \label{pre:1}
 Let $\mathcal{L}_{\parameters}(\training)$ be a micro-macro loss defined by Equations~\eqref{eq:local-loss}--~\eqref{eq:statistics-model}. Then 
 \begin{align}
     \mathcal{L}_{\parameters}(\training) \le E_{\linstance \sim 
q_{\eparameters}(\linstance|\training,\xinstance)}
\big[-\ln p(\training|\prob{\training}_{\linstance}) - \tradeoff \sum_{\kthFeature=1}^{\featureNUM} \big( \frac{1}{|\ofeature_{\kthFeature}|} \ln{\mathcal{N}(\ofeature_{\kthFeature}|\featureFunction_{\kthFeature}(\prob{\training}_{\linstance}),\kweight_{\kthFeature}^{2} I)} \big] \label{eq:mm-objective} \\ 
+ (1+\tradeoff m) KL(q_{\eparameters}(\linstance|\training,\xinstance)||p(\linstance)) 
\notag
 \end{align}
 \
  where $q_{\eparameters}(\linstance|\training,\xinstance) =  q(\linstance|\training,\X,\ofeature_{1},\ldots,\ofeature_{\featureNUM})$ is an approximate posterior distribution with parameters $\eparameters$.
 \end{proposition}
 The proof is in the Appendix section~\ref{sec:proof1}. The basic idea is to combine ELBOs for each individual loss term. Adding a hyperparameter $\beta$ to control the relative importance of the KL-divergence (as in the $\beta$-VAE~\cite{higgins2016beta}), 
we obtain the following {\em training objective} for a set of observed training graphs $\A^{1},\ldots,\A^{\samplesize}$:
 \begin{align} 
\label{eq:micromacro-elbo}
 \argmin_{\eparameters,\dparameters,\set{\kweight}} \sum_{\sindex=1}^{\samplesize} 
     E_{\linstance \sim 
q_{\eparameters}(\linstance|\training^{\sindex},\X)}
\big[-\ln p(\training^{\sindex}|\prob{\training}_{\linstance}) - \tradeoff  \sum_{\kthFeature=1}^{\featureNUM}  \frac{1}{|\ofeature_{\kthFeature}|} \ln{\mathcal{N}(\ofeature^{\sindex}_{\kthFeature}|\featureFunction_{\kthFeature}(\prob{\training}_{\linstance}),\kweight_{\kthFeature}^{2} I)} \big]  \\ + \beta KL(q_{\eparameters}(\linstance|\training^{\sindex},\X)||p(\linstance)) \notag  
\end{align}
In this objective, {\em the graph statistic reconstruction loss, $-\tradeoff  \sum_{\kthFeature=1}^{\featureNUM}  \frac{1}{|\ofeature_{\kthFeature}|} \ln{\mathcal{N}(\ofeature^{\sindex}_{\kthFeature}|\featureFunction_{\kthFeature}(\prob{\training}_{\linstance}),\kweight_{\kthFeature}^{2} I)}$, acts as a regularizer with respect to the edge reconstruction loss} $-\ln p(\training^{\sindex}|\prob{\training}_{\linstance})$.

 \subsection{Implementation} We implement the micro-macro \elbo\eqref{eq:micromacro-elbo} utilizing graph neural networks as follows. 
 
 \begin{itemize}
 \item The prior $p(\linstance)$ is a standard normal distribution.
   \item We limit the encoder input to the observed training graph, so 
         $q(\linstance|\training, \X ,\ofeature_{1},\ldots,\ofeature_{\featureNUM}) =q_{\eparameters}(\linstance|\training, \X).$
     Therefore we can use any standard graph encoder with parameters $\eparameters$. The encoder deterministically maps an input graph 
     $(\training,\X)$ to a posterior distribution $q_{\eparameters}(\linstance|\training, \X)$. 
     \item The decoder deterministically maps a latent representation $\linstance$ to a probabilistic graph $\prob{\training}_{\linstance}$ with parameters $\dparameters$. 
     Any standard graph decoder can be used.
     \item For each graph-level statistic, the diagonal co-variance parameter is computed using the optimal $\sigma$-VAE method from the calibrated Gaussian framework~\cite{rybkin2021simple}. This uses the maximum likelihood estimate given the estimated means
     with respect to a minibatch of size $B$:
\begin{align*}
   {\kweight_{\kthFeature}^{2}} = \frac{1}{B} \sum_{j=1}^{B} \mse( \featureFunction_{\kthFeature}(\prob{\training}_{\linstance^{j}}),\ofeature^{\sindex}_{\kthFeature})), \textit{~~~~~~~~~~}\mse(\set{x},\set{\mu}) = \frac{1}{|\mu|} \sum_{l=1}^{|\mu|} (\set{x}_{l} - \set{\mu}_{l})^{2}.
\end{align*}
 \end{itemize}
 
\begin{figure}[h]
  \centering
  \resizebox{.60\textwidth}{!}{
     \centering
          \begin{subfigure}[b]{.23\textwidth}
         \centering
         \includegraphics[width=1\textwidth]{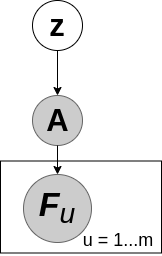}\caption{\label{fig:PGM}}
     \end{subfigure}
     \hfill
     \begin{subfigure}[b]{.75\textwidth}
         \centering
         \includegraphics[width=1\textwidth]{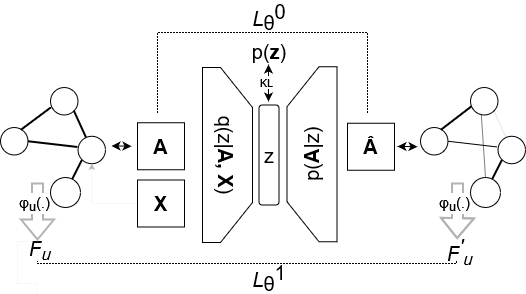}\caption{\label{fig:VAE}}
     \end{subfigure}
   }
     \caption{(a) Micro-macro generative model. Observed  variables in gray. (b) GraphVAE-\ourModelAcronym~ architecture. \label{fig:GenerativeMoldel}}
\end{figure}

 Figure \ref{fig:VAE} plots the architecture. The advantages of the calibrated Gaussian model for generating graph statistics are as follows. (1) In experiments on non-relational data, the calibrated Gaussian model achieves state-of-the-art performance.
 (2) Learning a variance parameter eliminates hyperparameters compared to user-assigned weights for each graph statistic. (3) The variance values can be interpreted as quantifying the empirical uncertainty of a graph feature.
 For example, if all training graphs exhibit a similar number of triangles, the corresponding variance parameter will be small. Table~\ref{table:variances} in the Appendix  illustrates this phenomenon in benchmark datasets.

\section{Graph statistics} \label{sec:kernels}
The micro-macro \elbo\eqref{eq:micromacro-elbo} requires user-specified input graph descriptors for computing target graph statistics. In our experiments, we utilize several types of {\em default statistics} for regularizing graph embeddings that have the following advantages. (1) Known from prior research to be generally important for graph modeling across different domains. (2) Easy to interpret. (3) Differentiable with respect to the entries in a reconstructed soft adjacency matrix. In an application, the default statistics can be combined with other statistics of interest. 

Graph descriptors used in network analysis and graph kernels 
comprise two main groups~\cite{hamilton2020graph}: (1) Summaries of local node-level properties. We utilize a  descriptor derived from the degree histogram, and the number of triangles in a graph.
 (2) Higher-order proximity relations between nodes. We utilize $s$-step transition probability matrices,   
for $s=1,\ldots 5$~\cite{cao2015grarep,DBLP:conf/nips/0007WX0N18}. These descriptors are differentiable with respect to the entries in a reconstructed soft adjacency matrix $\prob{\A}$, which we view as representing a   weighted undirected graph. 

\paragraph{Degree Histogram.} The \textbf{degree} of node $v_\unode$ is given by $d(v_\unode) = \sum_{\vnode} \prob{\A}_{\unode\vnode}$. We adapt the permutation-invariant differentiable histogram layer (DHL)~\cite{wang2016learnable}.
 
The DHL is based on a soft assignment of points to bins given bin centers and widths. Binning the soft degrees forms a soft degree histogram of graph $\prob{\A}$ comparable to the hard degree histogram. In detail, the bin centers are the possible (hard) node degrees $b=0,\ldots,\numnodes$. All widths are uniformly set to 0.1 (based on experimentation). Then we have 
\begin{eqnarray*}
\dhk_{\prob{\A}}(b) \equiv \sum_{\unode = 1}^{\numnodes} a(v_{\unode},b), \textit{~~~~~~~~~~}a(v_{\unode},b)=\max \{0, 1 - 0.1 \cdot |d(v_\unode) - b| \}.  
    \label{eq:degree}
\end{eqnarray*}
Thus the membership $a(v_{\unode},b)$ of a node in a bin $b$ ranges from 0 to 1 and decreases with the difference between the node's expected degree and the bin center. The DHL assigns to each bin the sum of nodes membership in the bin. 
%
     In a naive implementation, the computational cost of finding the $\dhk_{\prob{\A}}(b)$ vector
     is $O(\numnodes^2)$. In our experiments, we use $\numnodes$ parallel processors for a near constant time to obtain the soft degree histogram.
    
     \paragraph{$S$-Step Transition Probability Kernel.} $P^s(\prob{\A})$ is the $\nodeNum \times \nodeNum$ $s$-step transition probability matrix derived from adjacency matrix $\prob{\A}$ such that $[P^s(\prob{\A})]_{\nodem,\noden}$ is the probability of a transition from node 
     $\nodem$ to node $\noden$ in a random walk of $s$ steps started from ${\nodem}$.
     The transition matrix $P^s(\prob{\A})$ can be computed as $        P^s(\prob{\A}) = (D(\prob{\A})^{-1}\prob{\A})^{s}$  where $\D(\prob{\A})$ is a diagonal matrix with $\D(\prob{\A})_{\unode \unode}= d(v_\unode)$.
          
     The GraRep system learns node representations that reconstruct the $s$-step transition probabilities, using random walks and matrix factorization~\cite{cao2015grarep}. 
     The transition probability matrix 
    is usually dense and encodes the connectivity information of the graph. 
     An important difference to the adjacency matrix is that whereas adding or removing an edge changes only one adjacency, it can and often does result in a substantive change in many transition matrix elements. The descriptor $P^s(\prob{\A})$  thus measures the importance of an edge in the graph structure (cf. Figure~\ref{fig:crossentropyExample}). 
 In our experiments, we compute the transition probability matrix exactly with a runtime cost of $O(\numnodes^3)$. 
%

\paragraph{Triangle Count.}
The number of triangles in a simple graph  $\A$ is computed by $\mathit{Tri}(A) = \sum_{i}{(A^3)}_{ii}$

  with a runtime cost of $O(\numnodes^3)$. The number of triangles is a fundamental graph statistic in network science~\cite{bera2020count} with many applications in graph mining~\cite{arifuzzaman2013patric}. For example it has been used to detect spamming activity and
assess content quality in social networks~\cite{becchetti2008efficient}, to uncover thematic structure in the world-wide web~\cite{eckmann2002curvature} and for query planning 
in databases~\cite{bar2002reductions}. It is also used extensively in exponential random graph models~\cite{harris2013introduction}. 

\section{Empirical evaluation}

We describe our baseline methods and benchmark datasets, then report comparison results.

\subsection{Comparison methods} 
\label{publicRep}
We examine the effect of micro-macro modeling on a GraphVAE architecture. Our AB methodology is to keep the architecture the same and train the model using the joint MM \elbo\eqref{eq:micromacro-elbo} that combines both local and global graph properties. We also compare the MM GraphVAE with popular GGMs. As the architectures are not new, we describe them briefly. Our experiments used the public repositories and recommended hyperparameters by original creators; the Appendix  contains further details.

\textbf{GraphVAE.~}
A popular model that transforms a graph-level latent variable to generate a soft adjacency matrix \cite{DBLP:conf/icann/SimonovskyK18}\cite[Ch.9.1.2]{hamilton2020graph}. 
The encoder utilizes a Multi-layer Graph Convolutional Network (GCN) followed by a graph level readout and a Fully Connected Layer (FCL) that outputs $(\mu, \sigma)$.
The decoder utilizes  a network with FCLs that outputs entries of a probabilistic adjacency matrix $\prob{\A}$. To evaluate the reconstruction probability, we use a BFS ordering of nodes~\cite{liao2019efficient}\cite[Ch.9.1.2]{hamilton2020graph}.\\
\textbf{GraphVAE-\ourModelAcronym.~} GraphVAE trained with the micro-macro objective~\ref{eq:micromacro-elbo}. \textbf{Our new method.}\\
\textbf{GraphRNN.~}  Auto-regressive method that generates the adjacency matrix incrementally.  Each step generates one entry in the GraphRNN design, or one column in the GraphRNN-S design~\cite{you2018graphrnn}.
\\
\textbf{GRAN.~} Auto-regressive method that
generates a block of nodes and associated edges at a step
~\cite{DBLP:conf/nips/LiaoLSWHDUZ19}.
\\
\textbf{BiGG.~} Auto-regressive method that leverages graphs sparsity to avoid generating the full adjacency matrix
~\cite{dai2020scalable}. To our knowledge BiGG achieves SOTA graph quality.

\subsection{ Benchmark datasets}
\label{hyper3}
 Our design closely follows previous experiments on generating realistic graph structures \cite{you2018graphrnn,DBLP:conf/nips/LiaoLSWHDUZ19}.
We utilize 3 synthetic, and 2 real-world datasets for the main results.
The synthetic    Grid and Lobster are from previous studies~\cite{you2018graphrnn,DBLP:conf/nips/LiaoLSWHDUZ19}, Triangle Grid is introduced in this paper. Protein and 
 ogbg-molbbbp are real-world datasets from biology with information about proteins and molecules respectively. 
 The Appendix, section \ref{sec:datasets}, contains further details, as well as results for 3 more real-world datasets.
 
 {\em Train/Test Split.} Following previous experiments~\cite{you2018graphrnn, li2015gated,dai2020scalable} we randomly split the graphs sets into train (70\%), validation (10\%) and test (20\%) sets.  We use the same train and test graph sets for all models. To evaluate a trained model, we generate $T$ new  graphs to compare with the $T$ graphs in the test set (cf. Section~\ref{sec:related}). Appendix Figure~\ref{fig:test}  illustrates the evaluation process.
 
 Network modeling at the micro-level might be used for surveillance of individuals and communities. Therefore ethical network data collection must take into account issues of consent, privacy, and bias. 
  None of the datasets used in this research study contain personally identifiable information, or offensive/harmful content about either individuals or communities. The  social impact of our work we expect to be on balance more positive than negative, because our macro-level model enhances the understanding of global network structure, not the targeting of individuals. 
\subsection{Evaluation} 
We empirically verify the  effectiveness of micro-macro modeling through different evaluation metrics: Qualitative and quantitative evaluation of graph quality, generation time, and training time.

 \label{sec:quality}

\begin{figure}[htbp]
     \centering
      \rotatebox{90}{Lobster}
          \begin{subfigure}[b]{.13\textwidth}
         
         \centering
         
         \includegraphics[width=\textwidth]{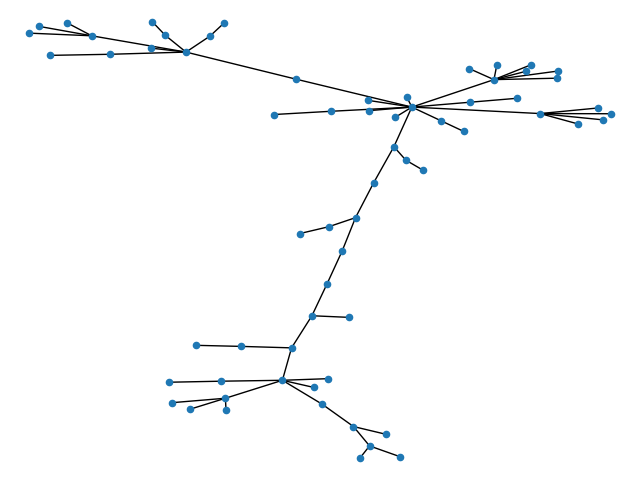}
     \end{subfigure}
          \hfill
          \begin{subfigure}[b]{.13\textwidth}
         \centering
         \includegraphics[width=\textwidth]{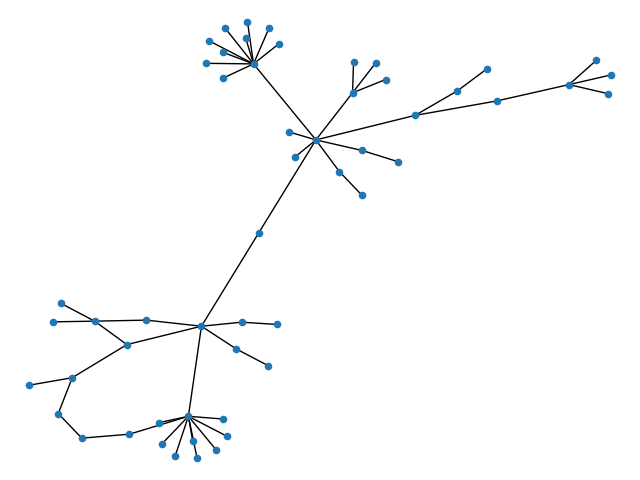}
     \end{subfigure}
          \hfill
          \begin{subfigure}[b]{.13\textwidth}
         \centering
         \includegraphics[width=\textwidth]{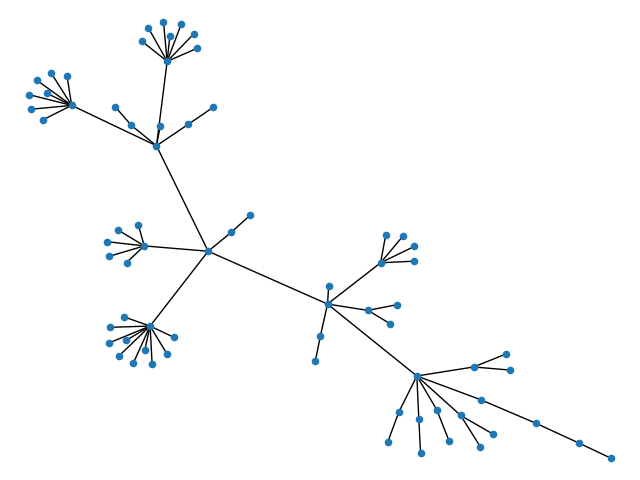}
     \end{subfigure}
        \hfill
        \begin{subfigure}[b]{.13\textwidth}
         \centering
         \includegraphics[width=\textwidth]{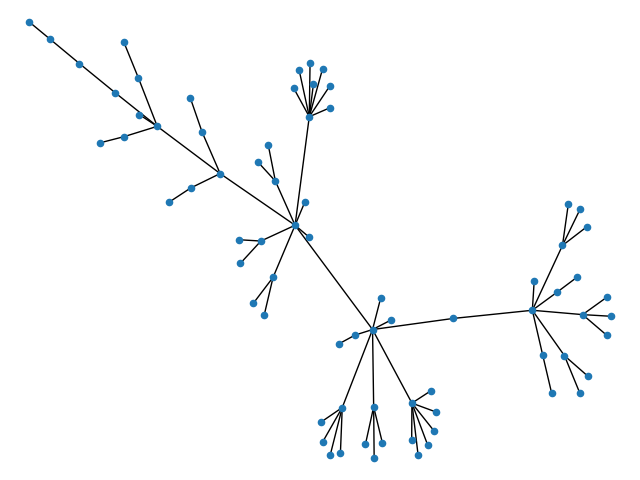}
     \end{subfigure}
        \hfill
    \begin{subfigure}[b]{0.13\textwidth}
        \centering
        \includegraphics[width=\textwidth]{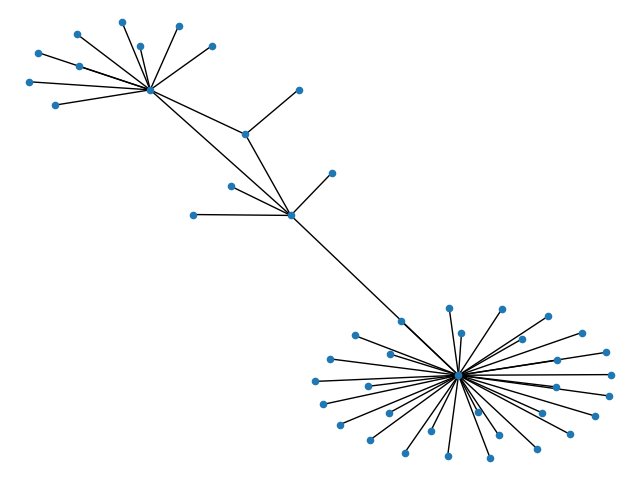}
    \end{subfigure}
    \hfill
     \begin{subfigure}[b]{0.13\textwidth}
         \centering
         \includegraphics[width=\textwidth]{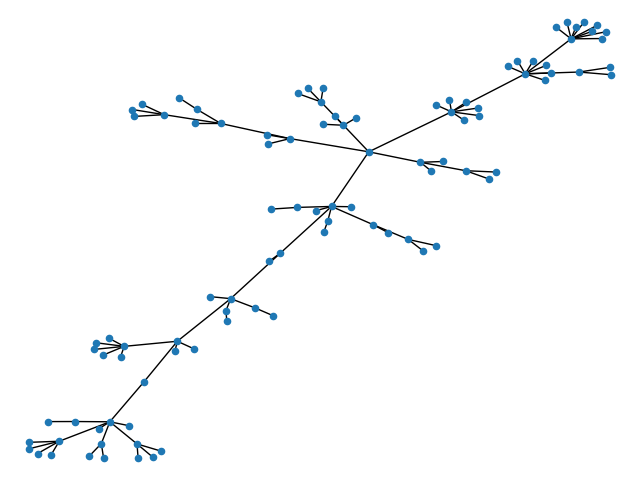}
     \end{subfigure}
          \hfill
     \begin{subfigure}[b]{0.13\textwidth}
         \centering
         \includegraphics[width=\textwidth]{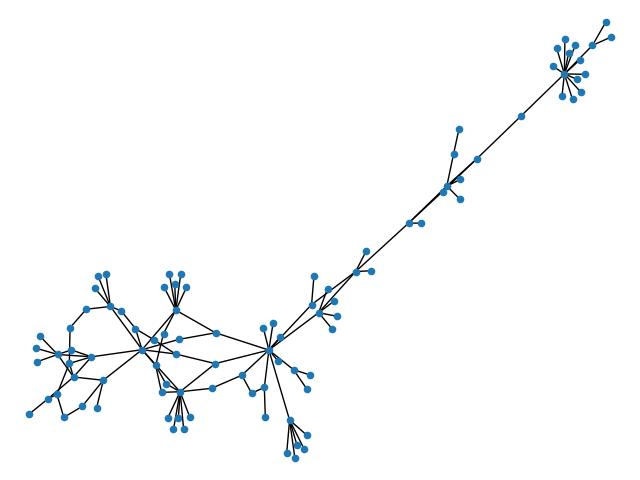}
     \end{subfigure}
      \vfill
     \rotatebox{90}{Lobster}
         \begin{subfigure}[b]{.13\textwidth}
         
         \centering
         \includegraphics[width=\textwidth]{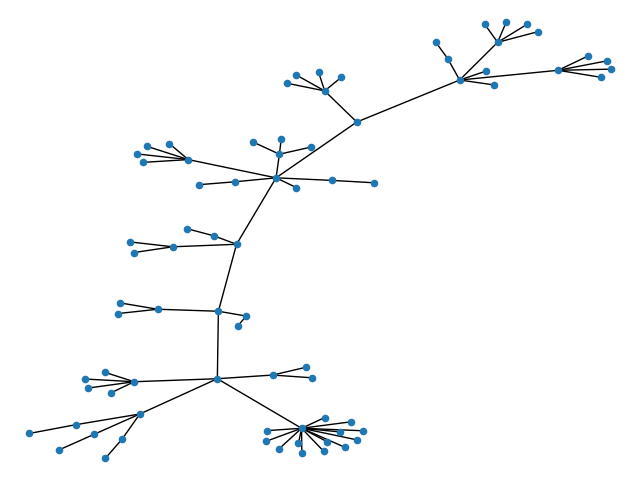}
     \end{subfigure}
          \hfill
     \begin{subfigure}[b]{0.13\textwidth}
         \centering
         \includegraphics[width=\textwidth]{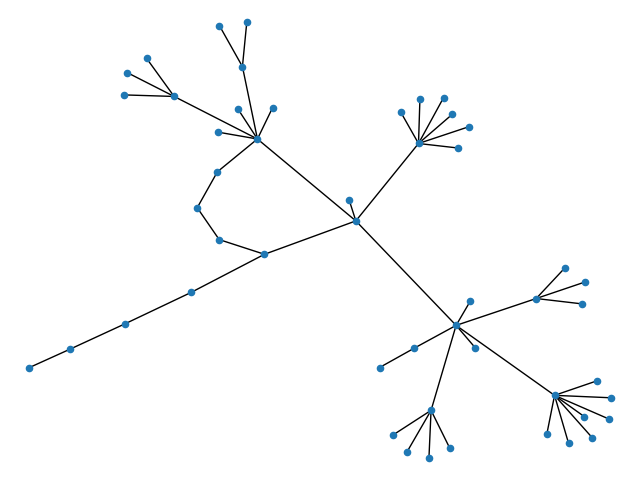}
     \end{subfigure}
          \hfill
     \begin{subfigure}[b]{0.13\textwidth}
         \centering
         \captionsetup{font={footnotesize}}
         \includegraphics[width=\textwidth]{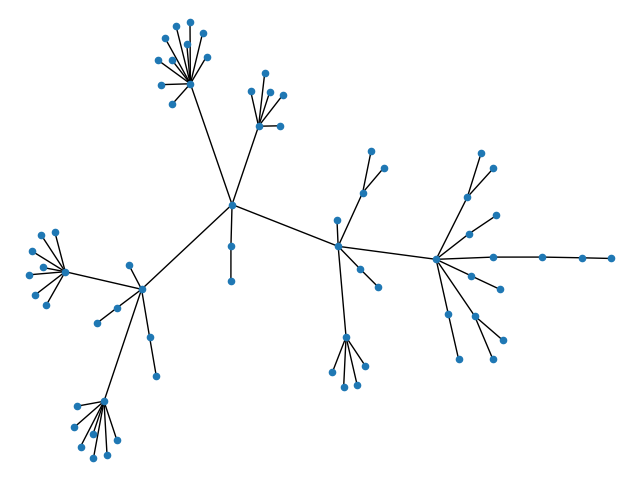}
     \end{subfigure}
          \hfill
                  \begin{subfigure}[b]{.13\textwidth}
         \centering
         \includegraphics[width=\textwidth]{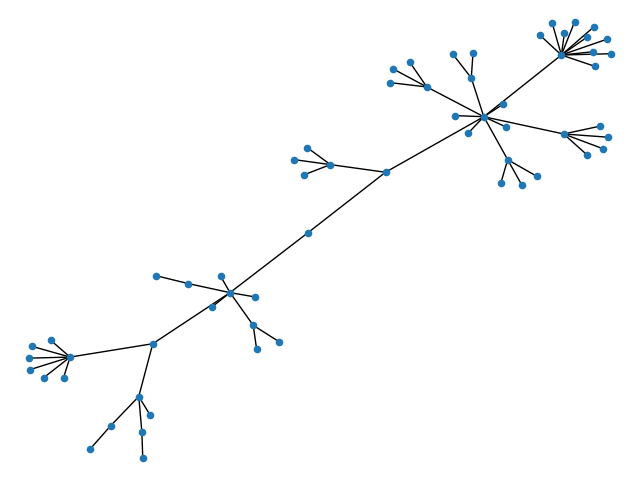}
     \end{subfigure}
        \hfill
\begin{subfigure}[b]{0.13\textwidth}
    \centering
    \includegraphics[width=\textwidth]{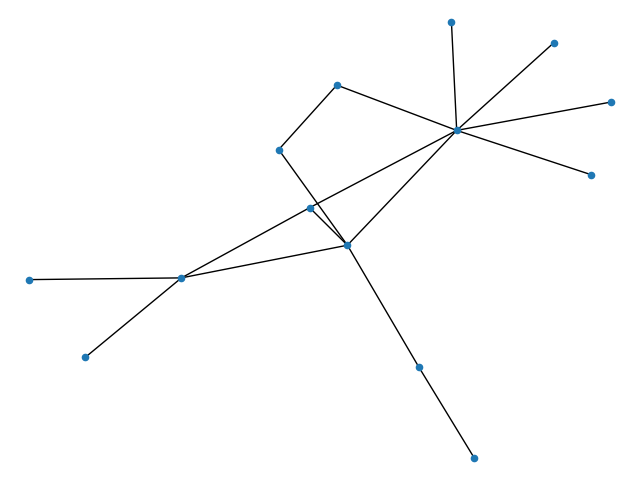}

\end{subfigure} 
     \hfill
     \begin{subfigure}[b]{0.13\textwidth}
         \centering
         \includegraphics[width=\textwidth]{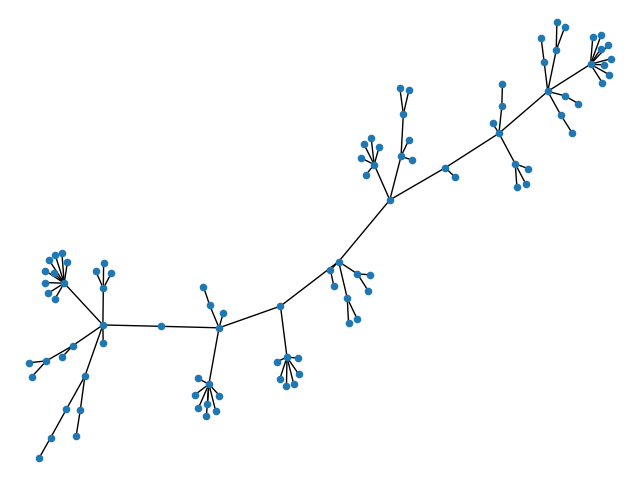}
     \end{subfigure}
              \hfill
     \begin{subfigure}[b]{0.13\textwidth}
         \centering
         \includegraphics[width=\textwidth]{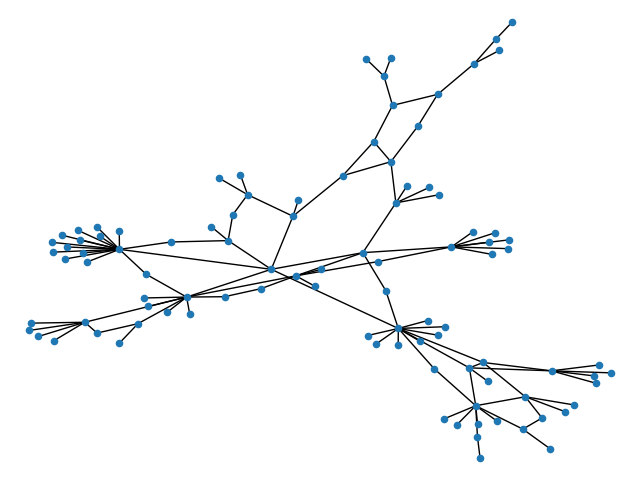}
     \end{subfigure}
     \vfill
     \rotatebox{90}{\quad Grid}
     \begin{subfigure}[b]{.13\textwidth}
      
         \centering
    
         \includegraphics[width=\textwidth]{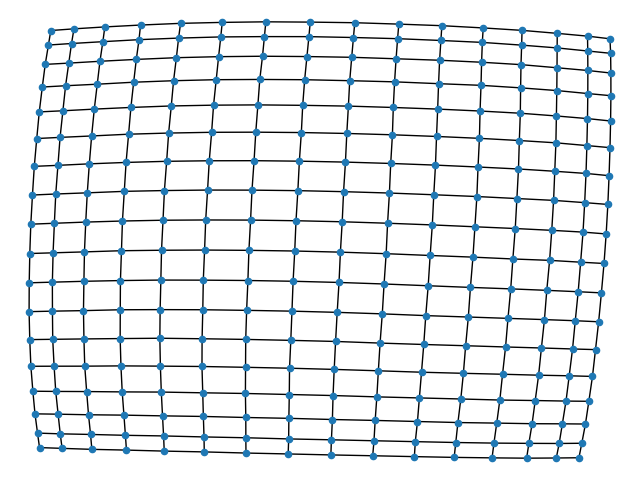}
     \end{subfigure}
          \hfill
          \begin{subfigure}[b]{.13\textwidth}
         \centering
         \includegraphics[width=\textwidth]{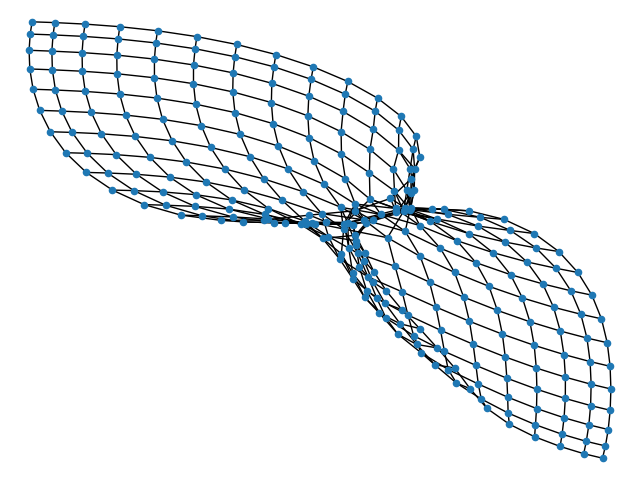}
     \end{subfigure}
          \hfill
          \begin{subfigure}[b]{.13\textwidth}
         \centering
         \includegraphics[width=\textwidth]{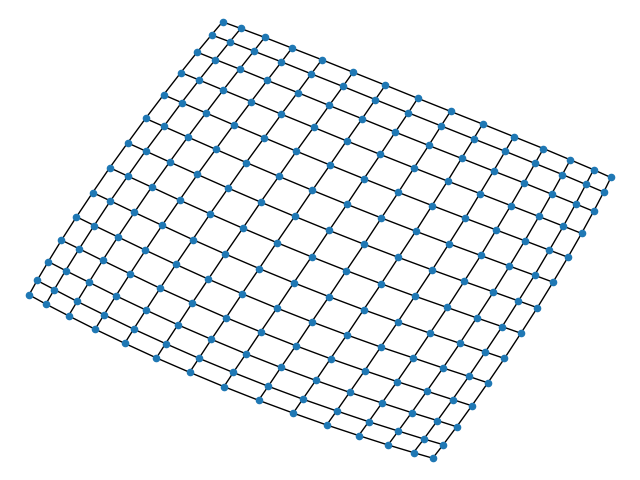}
     \end{subfigure}
        \hfill
        \begin{subfigure}[b]{.13\textwidth}
         \centering
         \includegraphics[width=\textwidth]{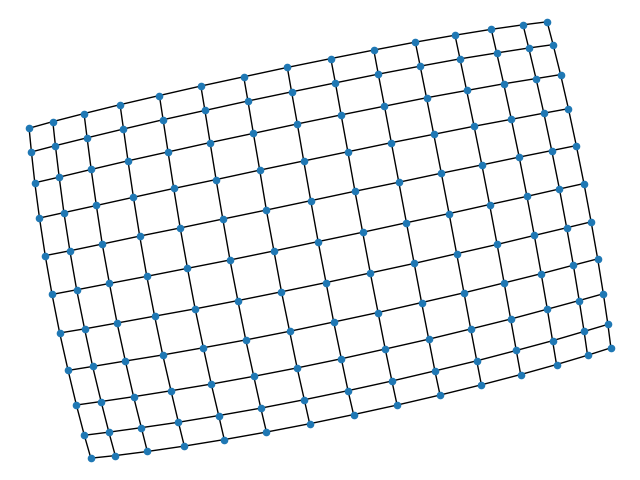}
     \end{subfigure}
        \hfill
    \begin{subfigure}[b]{0.13\textwidth}
        \centering
        \includegraphics[width=\textwidth]{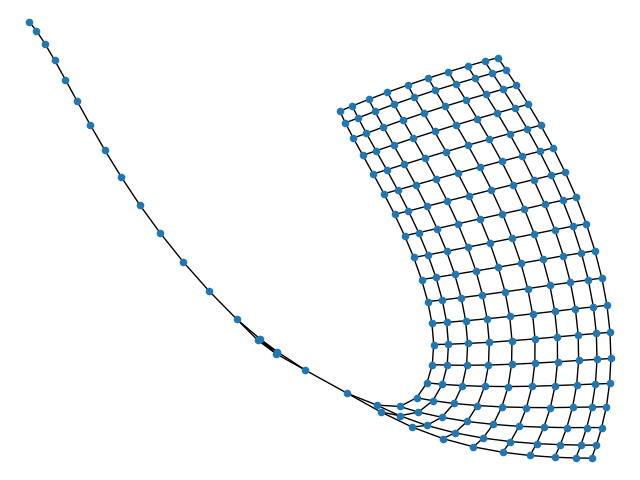}
    \end{subfigure}
    \hfill
     \begin{subfigure}[b]{0.13\textwidth}
         \centering
         \includegraphics[width=\textwidth]{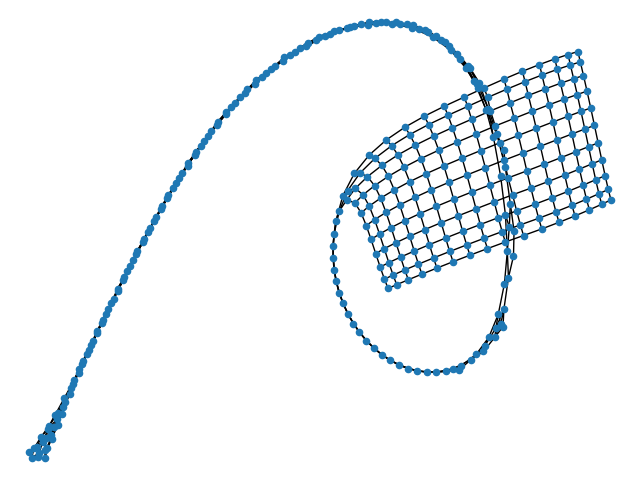}
     \end{subfigure}
          \hfill
     \begin{subfigure}[b]{0.13\textwidth}
         \centering
         \includegraphics[width=\textwidth]{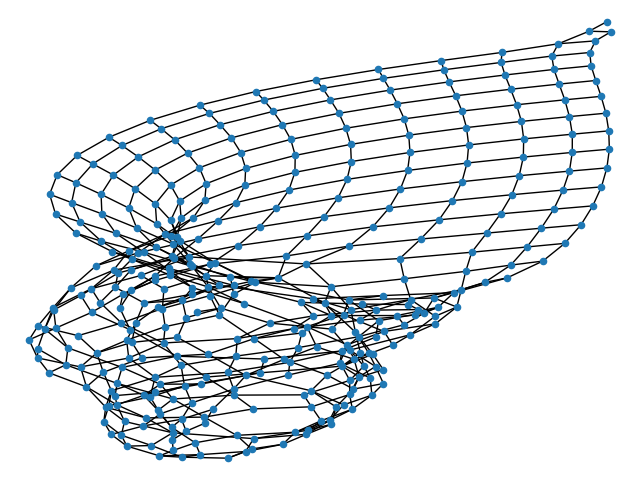}
     \end{subfigure}
     \vfill
     \rotatebox{90}{\qquad Grid}
         \begin{subfigure}[b]{.16\textwidth}
         \centering
         \includegraphics[width=\textwidth]{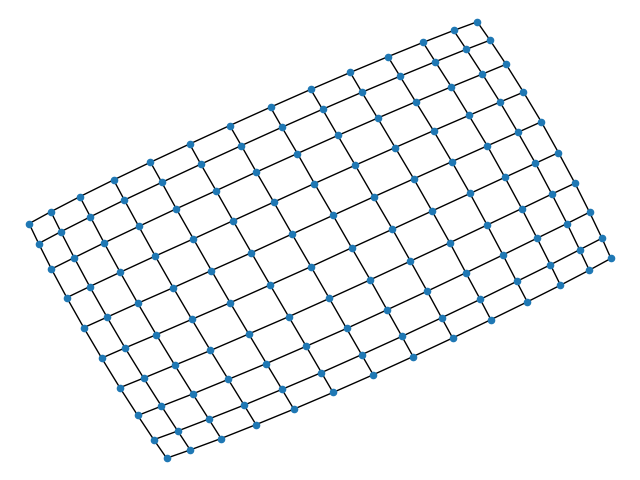}
         \captionsetup{format=plain, font=tiny, labelfont=bf}
         \caption{Test}
         
         \label{fig:lobstersamples}
     \end{subfigure}
          \hfill
     \begin{subfigure}[b]{0.13\textwidth}
     \captionsetup{format=plain, font=tiny, labelfont=bf}
         \centering
         \includegraphics[width=\textwidth]{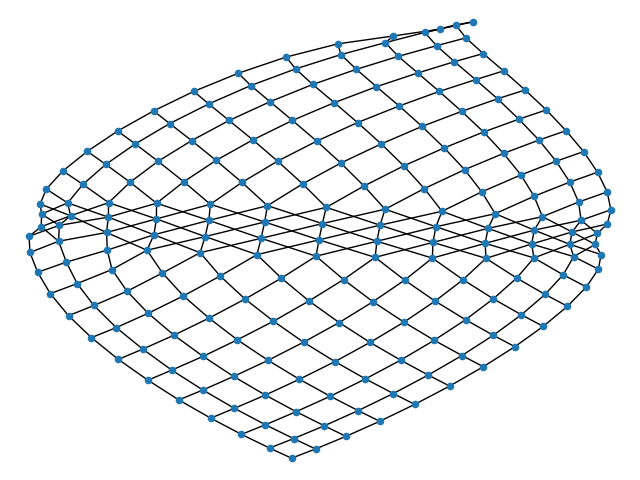}
         \caption{GraphVAE}
         \label{fig:molgan_lobster}
     \end{subfigure}
          \hfill
     \begin{subfigure}[b]{0.13\textwidth}
         \centering
         \captionsetup{format=plain, font=tiny, labelfont=bf}
         \includegraphics[width=\textwidth]{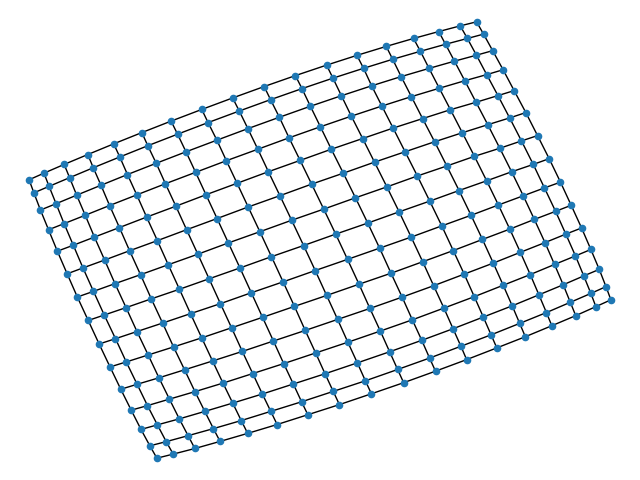}
         \caption{\textbf{GraphVAE-\ourModelAcronym}}
         \label{fig:graphite_lobster}
     \end{subfigure}
     \hfill
\begin{subfigure}[b]{0.13\textwidth}
    \centering
     \captionsetup{format=plain, font=tiny, labelfont=bf}
    \includegraphics[width=\textwidth]{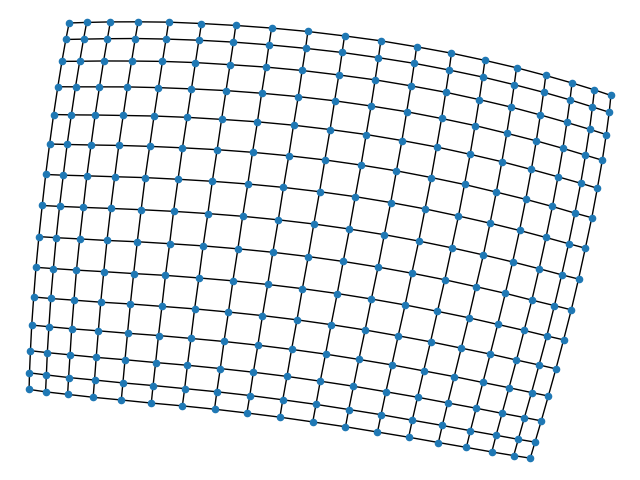}
    \caption{BiGG}
    \label{fig:lobsterkernel_big}
\end{subfigure} 
          \hfill
\begin{subfigure}[b]{0.13\textwidth}
    \centering
     \captionsetup{format=plain, font=tiny, labelfont=bf}
    \includegraphics[width=\textwidth]{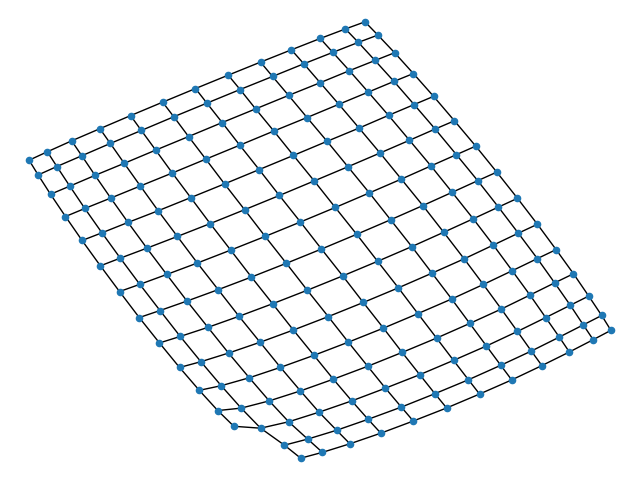}
    \caption{GRAN}
    \label{fig:lobsterkernel}
\end{subfigure} 
     \hfill
     \begin{subfigure}[b]{0.13\textwidth}
         \centering
         \captionsetup{format=plain, font=tiny, labelfont=bf}
         \includegraphics[width=\textwidth]{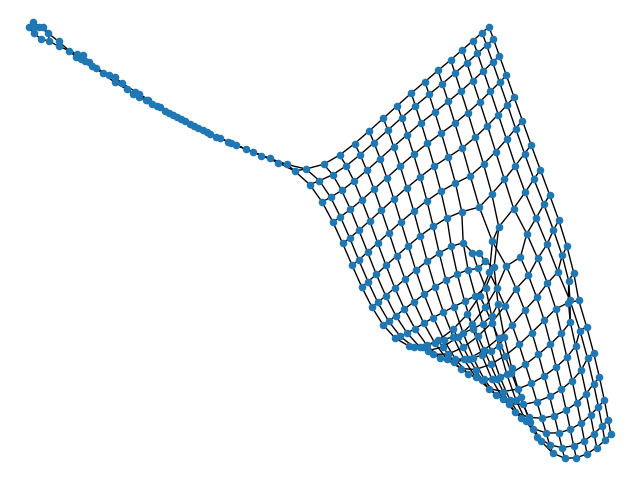}
         \caption{GraphRNN}
         \label{fig:lobsterFC}
     \end{subfigure}
              \hfill
     \begin{subfigure}[b]{0.13\textwidth}
         \centering
         \captionsetup{format=plain, font=tiny, labelfont=bf}
         \includegraphics[width=\textwidth]{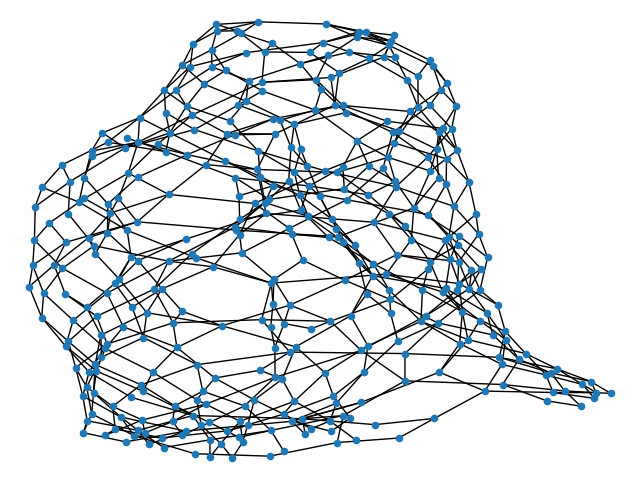}
         \caption{GraphRNN-S}
         \label{fig:lobster_DGLFRM}
     \end{subfigure}

    \caption{Visualization of generated graphs. 
    The left-most  column~(\ref{fig:lobstersamples}) shows  randomly selected graphs from the test set for each dataset, the other columns show graphs generated by each model from the prior. 
    The generated graphs shown are the two visually most similar samples in the generated set. GraphVAE-\ourModelAcronym~achieves the best visual match, much better than GraphVAE.}
        \label{fig:qualitative_comparision_grid}
\end{figure}

\textbf{Qualitative Evaluation.}
Following \cite{you2018graphrnn,DBLP:conf/nips/LiaoLSWHDUZ19} we compare the generated graphs by visual inspection. 
Figure~\ref{fig:qualitative_comparision_grid}  provides a visual comparison of randomly selected test graphs
and generated graphs. GraphVAE\ourModelAcronym~ and BiGG achieve the best visual match.
The Appendix, section \ref{section:Visualiztion}, provides more  examples. 

\textbf{Quantitative Evaluation.}
We follow the descriptor-based approach established in previous works (cf. Section~\ref{sec:related}). As recommended by~\citet{o2021evaluation}, we include  scores computed from a 50/50 split of the data set as an ideal score, i.e., a lower bound on the GGM scores.

(1) The GNN-based evaluation metrics \cite{thompson2022evaluation} MMD RBF and F1 PR compare generated and test graph embeddings computed by a reference GNN with randomly initialized weights. 
The reference embeddings are independent of any target graph statistics used in GraphVAE\ourModelAcronym. F1 PR stands for ``Improved Precision and Recall". Precision is the percentage of generated graph embeddings
that fall within the manifold of test graph embeddings, while recall is the percentage of test graph embeddings that fall
within the manifold of generated graph embeddings. This metric mainly captures the {\em diversity} of generated graphs~\cite{o2021evaluation}. 
MMD RBF compares generated graph embeddings and test graph embeddings using an RBF kernel. This metric captures the {\em realism} 
of the generated graphs. 

(2) A statistics-based evaluation compares generated and test graphs using {\em evaluation statistics},
namely node degree, clustering
coefficient, orbit counts, and the spectra of the graphs~\cite{DBLP:conf/nips/LiaoLSWHDUZ19,you2018graphrnn}. We add graph diameter, a statistic commonly studied in network science \cite{leskovec2010kronecker} related to graph connectivity. The target statistics used to train a GraphVAE\ourModelAcronym~model are distinct from the evaluation statistics used to measure graph quality (however, node degree and Degree histogram are closely related).
For each statistic, MMD is computed
using  total variation (TV) distance  \cite{DBLP:conf/nips/LiaoLSWHDUZ19}. 
  


\textbf{Impact on GraphVAE.} Table \ref{table:NN-BasedMertics-VGAE} shows a very large improvement from micro-macro modeling for both MMD RBF and F1 PR. The improvement in F1 PR ranges from 6\%-40\%, achieving a near perfect score. For MMD RBF the reduction ranges from 0.06 to 0.26 which is significant given the low ideal score.
%
Table~\ref{table:statMertics-VGAE} shows the results for evaluation graph statistics.  
Micro-macro modeling reduces MMD by 1-2 orders of magnitude on almost all datasets. In sum, MM modeling provides a large improvement in the realism and diversity of graphs generated by a GraphVAE architecture.

Lesion studies in the Appendix investigate the effects of each target statistic in isolation (Table~\ref{table:lesion}). No single statistic has the power of all three combined. We also observe that different target statistics have different importance for different datasets; see Table~\ref{table:variances}. 
%
\begin{table}[hbtp]
  \caption{
  GNN-based evaluation of micro-macro modeling for GraphVAEs. For each dataset we report the MMD RBF and F1 PR (see text) between the test set graphs (left-most column in Figure~\ref{fig:qualitative_comparision_grid}) and the generated graphs (see other columns in Figure~\ref{fig:qualitative_comparision_grid}). Values reported are the mean $\pm$ std. For MMD RBF smaller values are better, for F1 PR larger values are better.
  }
  \label{table:NN-BasedMertics-VGAE}
  \centering
  \resizebox{\textwidth}{!}{
  \begin{tabular}{lcccccccccc}
    \toprule
    \multirow{2}{3.5em}{\textbf{Method}} &  \multicolumn{2}{c}{\textbf{Triangle Grid}} &
    \multicolumn{2}{c}{\textbf{Lobster}} &
    \multicolumn{2}{c}{\textbf{Grid}} &  \multicolumn{2}{c}{\textbf{ogbg-molbbbp}} &
    \multicolumn{2}{c}{\textbf{Protein}} \\
    & \small{{MMD RBF}} & \small{F1 PR} & \small{{MMD RBF}} & \small{F1 PR} & \small{{MMD RBF}} & \small{F1 PR} & \small{MMD RBF} & \small{{F1 PR}} & \small{{MMD RBF}} & \small{{F1 PR}} \\
   \midrule
       50/50 split  & $0.03\pm0.00$ & $98.99\pm0.00$  & $0.04\pm0.00$ & $98.58\pm0.00$ & $0.009\pm0.00$ & $98.70\pm0.00$ & $0.002\pm0.00$ & $98.07\pm0.00$ & $0.04\pm0.00$ & $98.67\pm1.11$\\
    \midrule
   GraphVAE  & $0.23\pm0.01$ & $75.92\pm8.96$&$0.36\pm0.11$ & $78.48\pm24.13$ & $0.17\pm0.01$ & $75.52\pm2.53$   & $0.20\pm0.07$ & $54.53\pm6.15$ & $0.10\pm0.05$ & $84.11\pm9.56$\\
    GraphVAE\ourModelAcronym  &
    $\textbf{0.17}\pm\textbf{0.01}$ & $\textbf{83.58}\pm\textbf{5.50}$ & $\textbf{0.10}\pm\textbf{0.00}$ & $\textbf{100.00}\pm\textbf{0.00}$ & 
    $\textbf{0.13}\pm\textbf{0.01}$ & $\textbf{97.09}\pm\textbf{6.33}$ & $\textbf{0.02}\pm\textbf{0.01}$ & $\textbf{93.78}\pm\textbf{1.33}$ & $\textbf{\textbf{0.03}}\pm\textbf{0.01}$ & $\textbf{90.78}\pm\textbf{3.76}$\\
    \bottomrule
  \end{tabular}%
 } 
\end{table}
\begin{table}[hbtp]
         \caption{Statistics-based evaluation of micro-macro modeling for GraphVAEs.
         For a named evaluation graph statistic, each column reports the MMD between the test graphs  and the generated graphs.
         }
  \label{table:statMertics-VGAE}
    \begin{subtable}[h]{1\textwidth}
    \caption{{Synthetic Graphs} }
    \label{table:statMertics-VGAE-syn}
     \resizebox{\textwidth}{!}{
  \begin{tabular}{lccccccccccccccc}
    \toprule
    \multirow{2}{3.5em}{\textbf{Method}} &  \multicolumn{5}{c}{\textbf{Triangle Grid}} &
    \multicolumn{5}{c}{\textbf{Lobster}} & \multicolumn{5}{c}{\textbf{Grid}}
    \\
    & \small{Deg.} & \small{Clus.}& \small{Orbit} & \small{Spect} & \small{Diam.} & \small{Deg.} & \small{Clus.} &  \small{Orbit} & \small{Spect}& \small{Diam.} & \small{Deg.} & \small{Clus.} &  \small{Orbit} & \small{Spect}& \small{Diam.}  \\  
  \midrule
      50/50 split  & $3e^{-5}$  & $0.002$& $8e^{-5}$ & $0.004$ & $0.014$& $0.002$ & $0$  & $0.002$ & $0.005$& $0.032$ & $1e^{-5}$ & $0$  & $2e^{-5}$ & $0.004$ & $0.014$ \\
    \midrule
        GraphVAE & $0.082$  & $0.442$& $0.421$ & $0.020$ & $0.152$& $0.081$ & $0.739$  & $0.372$ & $0.056$  & $\textbf{0.129}$ & $0.062$ & $0.055$ & $0.515$ & $0.018$ & $0.143$ \\ 
    GraphVAE\ourModelAcronym   & $\textbf{0.001}$ & $\textbf{0.093}$& $\textbf{0.001}$& $\textbf{0.013}$& $\textbf{0.133}$ &  $\textbf{2}e^\textbf{{-4}}$ & $\textbf{0}$ & $\textbf{0.008}$ & $\textbf{0.017}$ & ${0.187}$ & $\textbf{5}e^\textbf{{-4}}$ & $\textbf{0}$ &  $\textbf{0.001}$ &  $\textbf{0.014}$ & $\textbf{0.065}$\\
    \bottomrule
  \end{tabular}%
  }  
    \end{subtable}
    \begin{subtable}[h]{1\textwidth}
    \centering
      \caption{Real Graphs} 
      \label{table:statMertics-VGAE-real}
    \resizebox{0.7\textwidth}{!}{
  \begin{tabular}{lcccccccccc}
    \toprule
    \multirow{2}{3.5em}{\textbf{Method}} &  \multicolumn{5}{c}{\textbf{Protein}} &
    \multicolumn{5}{c}{\textbf{ogbg-molbbbp}} 
%
    \\
    & \small{Deg.} & \small{Clus.} & \small{Orbit.} & \small{Spect.} & \small{Diam.} & \small{Deg.} & \small{Clus.} &  \small{Orbit.} & \small{Spect.} & \small{Diam.}     \\  
   \midrule
       50/50 split   & $4e^{-5}$ & $0.004$ & $5e^{-4}$ & $4e^{-4}$ & $0.003$& $2e^{-4}$ & $2e^{-5}$ & $9e^{-5}$ & $5e^{-4}$ & $0.002$ \\
    \midrule
    GraphVAE   & $0.022$ & $0.108$ &  $0.577$ & $0.016$ & $\textbf{0.080}$& $0.028$ & $0.442$ & $0.047$ & $0.015$ & $0.055$ \\
  GraphVAE\ourModelAcronym      & $\textbf{0.006}$ & $\textbf{0.059}$ & $\textbf{0.152}$ & $\textbf{0.007}$ & $0.091$ & $\textbf{0.001}$ & $\textbf{0.005}$ & $\textbf{8e}^\textbf{-4}$ & $\textbf{0.005}$ & $\textbf{0.018}$  \\
    \bottomrule
  \end{tabular}%
  }%
     \end{subtable}
\end{table}
\begin{table}[h]
  \caption{GNN-based comparison with benchmark GGMs. See table \ref{table:NN-BasedMertics-VGAE} caption. The best result is in bold and the second best is underlined.
  }
  \label{table:NN-BasedMertics-SOTA}
  \centering
  \resizebox{\textwidth}{!}{
  \begin{tabular}{lcccccccccc}
    \toprule
    \multirow{2}{3.5em}{\textbf{Method}} &  \multicolumn{2}{c}{\textbf{Triangle Grid}} &
    \multicolumn{2}{c}{\textbf{Lobster}} &
    \multicolumn{2}{c}{\textbf{Grid }} &  \multicolumn{2}{c}{\textbf{ogbg-molbbbp}} &
    \multicolumn{2}{c}{\textbf{Protein}}  \\
    & \small{{MMD RBF}} & \small{F1 PR} & \small{{MMD RBF}} & \small{F1 PR} & \small{{MMD RBF}} & \small{F1 PR} & \small{MMD RBF} & \small{{F1 PR}} & \small{{MMD RBF}} & \small{{F1 PR}}\\
   \midrule
       50/50 split  & $0.03\pm0.00$ & $98.58\pm0.00$ & $0.04\pm0.00$ & $98.58\pm0.00$ & $0.009\pm0.00$ & $98.70\pm0.00$   & $0.002\pm0.00$ & $98.07\pm0.00$ & $0.04\pm0.00$ & $98.67\pm1.11$ \\
    \midrule
    GraphVAE\ourModelAcronym  & $\textbf{0.17}\pm\textbf{0.01}$ & $\textbf{83.58}\pm\textbf{5.50}$  & $\textbf{0.10}\pm\textbf{0.00}$ & $\textbf{100.00}\pm\textbf{0.00}$ &
    $\textbf{0.13}\pm\textbf{0.01}$ & $\textbf{97.09}\pm\textbf{6.33}$ & $\textbf{0.02}\pm\textbf{0.01}$ & ${93.78}\pm{1.33}$ & $\textbf{0.03}\pm\textbf{0.01}$ & $90.78\pm3.76$\\
  \midrule 
  GraphRNN-S~\cite{you2018graphrnn} & $0.72\pm0.17$ & $33.68\pm19.44$ & $0.98\pm0.13$ & $58.72\pm7.55$  & $0.79\pm0.08$ & $71.18\pm2.36$ & $0.48\pm0.02$ & $81.41\pm0.71$ & $0.28\pm0.26$ & $72.36\pm27.63$\\
      GraphRNN~\cite{you2018graphrnn} &$0.64\pm0.11$ & $25.80\pm11.75$&  $0.87\pm0.04$ & $61.97\pm0.00$  &  $0.99\pm0.03$ &
      $13.22\pm0.05$  & $1.45\pm0.19$  & $\textbf{98.94}\pm\textbf{0.56}$ & ${0.32}\pm{0.14}$ & ${93.94}\pm{0.56}$\\
    GRAN~\cite{DBLP:conf/nips/LiaoLSWHDUZ19} &  $0.88\pm0.09$ & $23.71\pm9.72$ & $0.24\pm0.04$ & $50.53\pm12.12$  & $0.40\pm0.00$ & $78.73\pm0.02$& $0.39\pm0.07$ & $94.06\pm2.60$ & $\underline{0.07\pm0.00}$ & $\underline{98.05\pm0.76}$\\
        BiGG~\cite{dai2020scalable} & $\underline{0.41\pm0.13}$ & $\underline{62.08\pm0.14}$  & $\underline{0.12\pm0.00}$ & $\underline{99.74\pm0.76}$  &
        $\underline{0.35\pm0.00}$ & $\underline{{92.43}\pm{0.00}}$
        & $\underline{0.04\pm0.00}$ & $\underline{{96.16}\pm{0.31}}$ & $0.15\pm0.00$ & $\textbf{98.11}\pm\textbf{0.62}$\\
    \bottomrule
  \end{tabular}%
 } 
\end{table}

\textbf{GraphVAE\ourModelAcronym~vs. Benchmark GGMs.} For benchmarking we include GGMs 
that are known to generate realistic graphs. 
Tables \ref{table:NN-BasedMertics-SOTA} shows the GNN-based quality scores. Other than the most recent BiGG method, GraphVAE\ourModelAcronym~achieves a much better score.
MMD RBF scores, and 3 out of 5 F1 PR scores are also better in GraphVAE\ourModelAcronym~compared to BiGG.
Triangle Grid shows the biggest improvement, which illustrates the usefulness of matching triangle counts for this dataset.

Appendix Table~\ref{table:statMertics-SOTA} shows the benchmark results of statistics-based evaluation; we summarize them here.
On synthetic graphs, the GraphVAE\ourModelAcronym~ scores are superior to or  competitive with the BiGG and GRAN scores. On the real-world graphs, the GraphVAE\ourModelAcronym~ scores are  competitive with the BiGG and GRAN scores, and superior to the other benchmarks.
Given the already strong performance of the auto-regressive models, we conclude that GraphVAE\ourModelAcronym~generates high-quality graphs. 

 \begin{figure}[h]
     \centering
     \begin{subfigure}[b]{.495\textwidth}
         \centering
         \includegraphics[width=\textwidth,height=3cm]{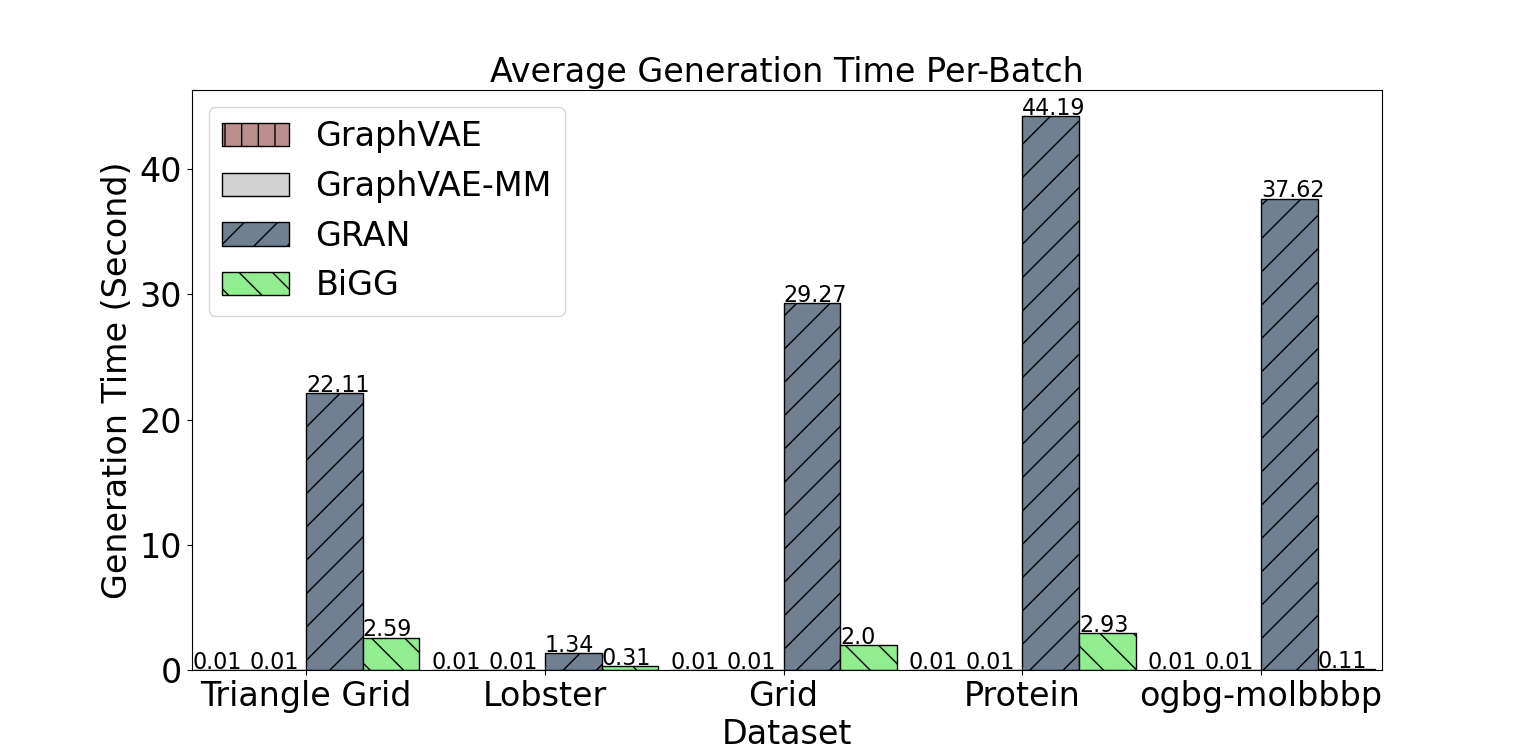}
     \end{subfigure}
     \hfill
     \begin{subfigure}[b]{0.495\textwidth}
         \centering
         \includegraphics[width=\textwidth,height=3cm]{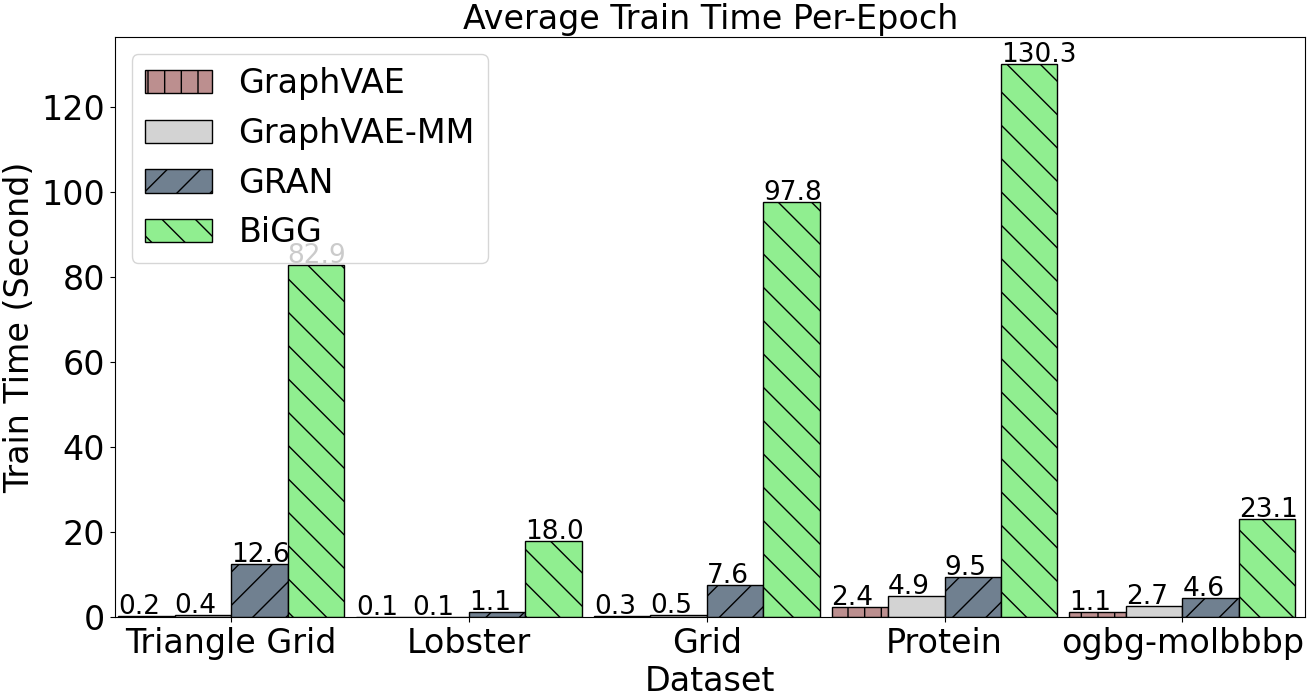}
     \end{subfigure}
     \hfill
     \caption{Comparison of benchmark GGMs with GraphVAEs on {\em generation time} (left) and  {\em training time} (right).  For visualizing the small generation time of GraphVAE and GraphVAE-\ourModelAcronym~, we round them up to 0.01, the first and second bar of each dataset respectively.}
        \label{fig:trainGenTime}
\end{figure}
\textbf{Generation Time.} 
The code for all models is run on the same system, detailed in  Appendix section \ref{sec:SystemArchitecture}.
Figure \ref{fig:trainGenTime} compares GraphVAEs to the fastest auto-regressive methods.
The auto-regressive methods require substantially more generation time than GraphVAEs.
While MM modeling slows down training, for the GraphVAE, the training time is still less than for the auto-regressive methods. The Appendix table \ref{table:time-SOTA} provides time measurements for all 
methods. The Appendix Table~\ref{table:SOTA-Complexity} presents worst-case complexity bounds  of the comparison methods.

\subsection{Further experiments on real-world graphs}
We conducted further experiments on the MUTAG, PTC \cite{xu2018powerful}, and QM9 \cite{DBLP:conf/icann/SimonovskyK18} datasets to evaluate micro-macro modeling on more real world graphs. The detailed results are in the Appendix section \ref{sec:ExtendedExp}; we summarize them here. For all datasets, GraphVAE\ourModelAcronym~offers much faster generation than the auto-regressive baselines. 
On MUTAG and PTC, the improvements from micro-macro modeling are even better than those we reported on the Protein and ogbg-molbbbp datasets. GraphVAE\ourModelAcronym~achieves substantive improvement in generation quality over all baselines, except for BiGG on PTC, which is competitive. 
 It was infeasible to train the auto-regressive methods on the  QM9 datasets (except for BiGG), so we report only the comparison of GraphVAE vs. GraphVAE\ourModelAcronym. Micro-macro modeling brings small improvement in graph quality on QM9, which is good given the strong score of GraphVAE 
on QM9.

\section{Limitations and discussion}
\label{limitation}
We discuss the limitations of our GraphVAE\ourModelAcronym~model and options for combining micro-macro modeling with other GGMs. 

\emph{Strengths and Weaknesses.}
GraphVAE\ourModelAcronym~inherits the strengths of  GraphVAE~\cite[Ch.9.1.2]{hamilton2020graph}: expressive power through graph embeddings, and fast generation due to all-at-once parallel edge generation. 
%
GraphVAE-\ourModelAcronym~also inherits the known limitations of GraphVAEs: 1) We need to know a maximum number of nodes before generation. (Smaller graphs can be generated using a mask.) 2) The decoder is an FCL that ouputs $\numnodes \times \numnodes$ numbers; we need to (implicitly) assume a node ordering to evaluate edge reconstruction probabilities based on the FCL output. The dependence on a node ordering is common to both GraphVAEs and auto-regressive GGMs, and efficient heuristics have been designed whose effectiveness has been confirmed in experiments, including those reported in this paper. We note that MM modeling makes GGM training less dependent on a node ordering because it uses permutation-invariant graph statistics.

\emph{Training Time Overhead.} Evaluating the GraphVAE\ourModelAcronym~\elbo during training incurs computational overhead compared to the GraphVAE \elbo due to computing graph statistics. We note that evaluating the edge reconstruction probability is already expensive and can dominate training time.
Table~\ref{table:SOTA-Complexity} in the Appendix presents worst-case complexity bounds for the training time of our comparison methods, which indicate that scaling GraphVAEs to very large graphs is a challenge.  
Despite the theoretical worst-case cost of evaluating graph statistics, Figure~\ref{fig:trainGenTime}(right) shows that the overhead was small in our experiments with medium-size graphs, due to parallelization. Also, 
approximating graph statistics 
is a promising avenue for significantly reducing the training time even for large graphs significantly~\cite{DBLP:conf/colt/JebaraK03,eden2017approximately,liao2019efficient,ribeiro2012estimation}.  
%


%
\emph{Micro-Macro Modeling for Other GGM Architectures.} In principle the MM objective can be applied with auto-regressive architectures as well: after a complete graph has been generated sequentially, the global loss can be backpropagated through the individual edge generation decisions.
%

In a graph GAN \cite{bojchevski2018netgan,NicolaMolGAN} the generator maps  a graph latent $\linstance$ to an adjacency matrix, or a random walk and the discriminator classifies them as real or synthetic. GAN models are not considered as SOTA GGMs \cite{dai2020scalable}.  Both random walk and adjacency matrices represent local micro-level information only \cite{cao2015grarep}. 
 A straightforward way to combine MM modeling with GANs is to augment the input to the discriminator with graph statistics computed for both real and generated graphs. 
%
%
\section{Conclusion and future work}
Our main idea is to model graph data jointly at two levels: a micro level based on local information (e.g. the existence of a link between two nodes) and a macro level based on aggregate graph statistics. We described a principled joint probabilistic model for both micro and macro levels, and derived an \elbo training objective for graph encoder-decoder models. Compared to previous micro level training objectives, the macro statistics regularize graph embeddings to match global graph statistics. To evaluate our model, we described a set of strong default graph statistics (node degree, number of triangles, transition probabilities). We applied the new training objective to the GraphVAE architecture, a widely used graph generative model based on latent graph representation. Micro-macro (MM) modeling greatly improved the quality of graphs generated by GraphVAE, to match or exceed that of benchmark models. MM modeling maintains the speed advantages of edge-parallel all-at-once graph generation. With an efficient computation of graph statistics, it provided fast training time as well compared to auto-regressive methods.  

Micro-macro modeling opens a number of fruitful avenues for future work. i) Investigating which graph statistics are important for generating which types of graphs. This connects with the rich area of graph kernels~\cite{nikolentzos2019graph} that are often based on graph statistics. ii) Investigating which graph statistics are important for particular domains.
iii) Developing a micro-macro model for other graph generative architectures, such as auto-regressive and GANs. 
iv) Evaluating the impact of micro-macro modeling on other graph-level tasks, such as graph classification.

In sum, modeling both global graph statistics and local information enhances the power of graph generation. Compared to using local information only, graph statistics can be used to regularize graph representations to efficiently generate realistic graphs.

\bibliography{main.bib}
\section*{Checklist}

\begin{enumerate}

\item For all authors...
\begin{enumerate}
  \item Do the main claims made in the abstract and introduction accurately reflect the paper's contributions and scope?
    \answerYes{}
  \item Did you describe the limitations of your work?
    \answerYes{See Section \ref{limitation} for details.}
  \item Did you discuss any potential negative societal impacts of your work?
    \answerYes{See Section \ref{hyper3} and \ref{sec:SocietalImpact} for details.}
  \item Have you read the ethics review guidelines and ensured that your paper conforms to them?
    \answerYes{}
\end{enumerate}

\item If you are including theoretical results...
\begin{enumerate}
  \item Did you state the full set of assumptions of all theoretical results?
    \answerYes{See Section~\ref{theory} for details.}
	\item Did you include complete proofs of all theoretical results?
    \answerYes{See Section~\ref{theory} and Appendix for details.}
\end{enumerate}

\item If you ran experiments...
\begin{enumerate}
  \item Did you include the code, data, and instructions needed to reproduce the main experimental results (either in the supplemental material or as a URL)?
    \answerYes{Code, datasets and  library requirements can be found at \url{https://github.com/kiarashza/GraphVAE-MM}}
  \item Did you specify all the training details (e.g., data splits, hyperparameters, how they were chosen)?
    \answerYes{See Sections~\ref{hyper3} and Appendix for all experiment settings. We also added all details, including models, generated samples, training logs and library requirements in our public repository.}
	\item Did you report error bars (e.g., with respect to the random seed after running experiments multiple times)?
    \answerYes{The focus of this paper is on the GNN-based metrics, tables \ref{table:NN-BasedMertics-VGAE} and \ref{table:NN-BasedMertics-SOTA}, containing error bar information. For all other compared metrics, the proposed approach typically results in 1-2 orders of magnitude of improvement. This evidence seems to be sufficient without error bars.} 
	\item Did you include the total amount of compute and the type of resources used (e.g., type of GPUs, internal cluster, or cloud provider)?
    \answerYes{See Section~\ref{sec:SystemArchitecture} for machine details.}
\end{enumerate}

\item If you are using existing assets (e.g., code, data, models) or curating/releasing new assets...
\begin{enumerate}
  \item If your work uses existing assets, did you cite the creators?
    \answerYes{Reported datasets and codes  are  open-source and publicly available. We cited them clearly in Tables and Context}
  \item Did you mention the license of the assets?
    \answerYes{All codes and datasets used in this study  are publicly available and clearly mentioned in the Section~\ref{hyper3} and \ref{publicRep}. Our implementation is publicly released under MIT license as mentioned in our repository.}
  \item Did you include any new assets either in the supplemental material or as a URL?
    \answerYes{We included our model implementation and  it is publicly available at \url{https://github.com/kiarashza/GraphVAE-MM}}
  \item Did you discuss whether and how consent was obtained from people whose data you're using/curating?
    \answerYes{As mentioned in  Section~\ref{publicRep}, we used the original papers' public repository; hence no consent was needed to curate this study. Also all used datasets are open source and publicly available.}
    \item Did you discuss whether the data you are using/curating contains personally identifiable information or offensive content?
    \answerYes{ As mentioned in  Section~\ref{hyper3}, None of the datasets used for this research study contain any personally identifiable information or offensive content.}
\end{enumerate}

\item If you used crowdsourcing or conducted research with human subjects...
\begin{enumerate}
  \item Did you include the full text of instructions given to participants and screenshots, if applicable?
    \answerNA{This point is not applicable for this research study.}
  \item Did you describe any potential participant risks, with links to Institutional Review Board (IRB) approvals, if applicable?
    \answerNA{This point is not applicable for this research study.}
  \item Did you include the estimated hourly wage paid to participants and the total amount spent on participant compensation?
    \answerNA{No such participants were used for this research study.}
\end{enumerate}

\end{enumerate}

\newpage

\section{Appendix} \label{Appendix} 

\subsection{Proof of proposition~\ref{pre:1}}
\label{sec:proof1}
 \begin{align} 
 &\mathcal{L}_{\parameters}(\training) = \localloss_{\parameters}(\training) + \tradeoff \globalloss_{\parameters}(\ofeature_{1},\ldots,\ofeature_{\featureNUM}) \notag
 \\&\le E_{\linstance \sim 
q_{\eparameters}(\linstance|\training,\xinstance,\ofeature_{1},\ldots,\ofeature_{\featureNUM})}
\big[-\ln p(\training|\prob{\training}_{\linstance} )  \big] + KL(q_{\eparameters}(\linstance|\training,\xinstance,\ofeature_{1},\ldots,\ofeature_{\featureNUM})||p(\linstance)) \label{eq:combine}
\\ &- \tradeoff \sum_{\kthFeature=1}^{\featureNUM} \frac{1}{|\ofeature_{\kthFeature}|}\big(E_{\linstance \sim 
q_{\eparameters}(\linstance|\training,\xinstance,\ofeature_{1},\ldots,\ofeature_{\featureNUM})}[ \ln{\mathcal{N}(\ofeature_{\kthFeature}|\featureFunction_{\kthFeature}(\prob{\training}_{\linstance}),\kweight_{\kthFeature}^{2} I)}]
\notag\\&+ KL(q_{\eparameters}(\linstance|\training,\xinstance,\ofeature_{1},\ldots,\ofeature_{\featureNUM})||p(\linstance)) \big)
\notag\\&=
E_{\linstance \sim 
q_{\eparameters}(\linstance|\training,\xinstance,\ofeature_{1},\ldots,\ofeature_{\featureNUM})}\big[-\ln p(\training|\prob{\training}_{\linstance} )  -\tradeoff\sum_{\kthFeature=1}^{\featureNUM}\frac{1}{|\ofeature_{\kthFeature}|}\ln{\mathcal{N}(\ofeature_{\kthFeature}|\featureFunction_{\kthFeature}(\prob{\training}_{\linstance}),\kweight_{\kthFeature}^{2} I)}\big] \label{eq:Linearity}\\&+ KL(q_{\eparameters}(\linstance|\training,\xinstance,\ofeature_{1},\ldots,\ofeature_{\featureNUM})||p(\linstance))
+\tradeoff \sum_{\kthFeature=1}^{\featureNUM} \frac{1}{|\ofeature_{\kthFeature}|}\big([ KL(q_{\eparameters}(\linstance|\training,\xinstance,\ofeature_{1},\ldots,\ofeature_{\featureNUM})||p(\linstance)) \big)\notag
\\&\le
E_{\linstance \sim 
q_{\eparameters}(\linstance|\training,\xinstance,\ofeature_{1},\ldots,\ofeature_{\featureNUM})}\big[-\ln p(\training|\prob{\training}_{\linstance} )  -\tradeoff\sum_{\kthFeature=1}^{\featureNUM}\frac{1}{|\ofeature_{\kthFeature}|}\ln{\mathcal{N}(\ofeature_{\kthFeature}|\featureFunction_{\kthFeature}(\prob{\training}_{\linstance}),\kweight_{\kthFeature}^{2} I)}\big]\label{eq:lowerbond}
\\&+
KL(q_{\eparameters}(\linstance|\training,\xinstance,\ofeature_{1},\ldots,\ofeature_{\featureNUM})||p(\linstance))
+\tradeoff\sum_{\kthFeature=1}^{\featureNUM} 
\big([ KL(q_{\eparameters}(\linstance|\training,\xinstance,\ofeature_{1},\ldots,\ofeature_{\featureNUM})||p(\linstance)) \big)\notag
\\&= E_{\linstance \sim 
q_{\eparameters}(\linstance|\training,\xinstance,\ofeature_{1},\ldots,\ofeature_{\featureNUM})}\big[-\ln p(\training|\prob{\training}_{\linstance} )  -\tradeoff\sum_{\kthFeature=1}^{\featureNUM}\frac{1}{|\ofeature_{\kthFeature}|}\ln{\mathcal{N}(\ofeature_{\kthFeature}|\featureFunction_{\kthFeature}(\prob{\training}_{\linstance}),\kweight_{\kthFeature}^{2} I)}\big] \label{eq:final}
\\&+(1+\tradeoff\featureNUM) KL(q_{\eparameters}(\linstance|\training,\xinstance,\ofeature_{1},\ldots,\ofeature_{\featureNUM})||p(\linstance))\notag
 \end{align}
 
 Inequality~\eqref{eq:combine} adds ELBOs for each individual loss term. 
 Equation~\eqref{eq:Linearity} uses the linearity of expectation. Inequality\eqref{eq:lowerbond} follows because  the fact that $|\ofeature_{\kthFeature}| \geq 1$. Equation~\eqref{eq:final} collects the KL expressions into a single term, and establishes the inequality~\eqref{eq:mm-objective}. 

\subsection{ Benchmark datasets }
\label{sec:datasets}
 Following previous studies \cite{you2018graphrnn,DBLP:conf/nips/LiaoLSWHDUZ19}, we utilize  synthetic and  real graph datasets as follows. 
 
\textbf{Triangle Grid.~} Includes 100 synthetic 2D graphs, regular tiling of the 2D plane with equilateral triangles, with $100\leq|\V|<400$~\cite{you2018graphrnn}.\\
\textbf{Lobster.~}  Includes 100 synthetic graphs with $10\leq|\V|\leq100$. Generated using the code from~\cite{you2018graphrnn}.\\
\textbf{(Square) Grid.~} Includes 100 synthetic 2D graphs, regular tiling of the 2D plane with equilateral squares, with $100\leq|\V|<400$~\cite{you2018graphrnn}.\\ 
\textbf{Protein.~~} Consists of  918 real-world protein graphs with $100\leq|\V|\leq500$~\cite{dobson2003distinguishing}.\\
\textbf{ogbg-molbbbp.~~}  Consists of  2039 real-world molecular graphs with $2\leq|\V|\leq132$~\cite{hu2020open}.

\subsection{Flowchart for evaluating generated graphs}

\begin{figure}[h]
\centering
\includegraphics[width=0.4\textwidth]{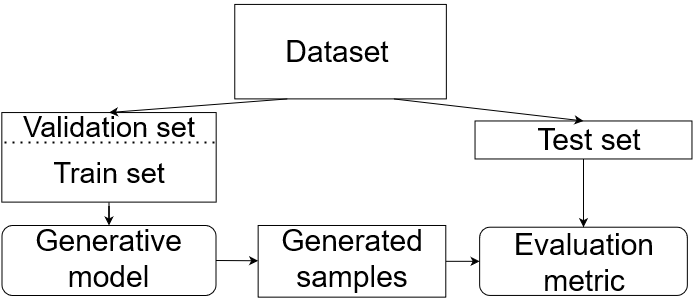}
\caption{ Each dataset is split into train, validation and test sets. A model is trained on the train set. The trained model generates new samples. The generated samples are then compared with the test set. \label{fig:test}}
\end{figure}

Figure \ref{fig:test} illustrates the  evaluation approach. To  split the datasets and for statistic-based evaluation metrics we use the code from~\cite{DBLP:conf/nips/LiaoLSWHDUZ19}. For GNN-based metrics we used the code from~\cite{thompson2022evaluation}. Following \citet{goyal2020graphgen}, for all the GGMs if a graph with disconnected components is generated, we take the maximum connected component.
\subsection{Setup and  hyper-parameters}
\paragraph{GraphVAE.} We examine the effect of micro-macro modeling on the GraphVAE architecture~\cite{DBLP:conf/icann/SimonovskyK18}.
 GraphVAE \textbf{encoder} \label{GraphVAE-architecture} utilizes a two-layer GCN (256, 1026 dimension)  followed by the graph-level output formulation (sum of nodes representation) finalized with FCL (1024 dimension) outputting the parameters of the variational posterior distribution. The model assumes isotropic and standard Gaussian distributions for the variational  posterior and prior, respectively.
The GraphVAE decoder is a three-layer fully connected neural network (1024,1024,1024 dimension) that directly maps the graph  representation to a probabilistic adjacency matrix. Layers utilize Layer Normalization \cite{ba2016layer}~and LeakyReLU activation function.
 Our AB methodology is to keep the architecture the same and train the model A) using GraphVAE \elbo i.e. $\localloss_{\parameters}(\training)$,  and B) the joint MM \elbo\eqref{eq:micromacro-elbo} that combines both local and global graph properties. GraphVAE and GraphVAE\ourModelAcronym~ are trained using the Adam optimizer~\cite{DBLP:journals/corr/KingmaW13} 
with a learning rate of $0.0003$ for $20,000$ epochs except for Lobster and ogbg-molbbbp, smaller datasets, which are trained for $10,000$ epochs. 
 For the hyperparameters $\tradeoff$ and $\beta$ see table \ref{table:lambdaWeights}. Hyperparameters are selected by validation set performance.
 
\paragraph{Baselines.}
For all Baselines, we used the implementation and hyperparameters setting provided by the original papers. 
 \begin{table}[h]
  \caption{ $\tradeoff$ and $\beta$ hyperparameters for each dataset  used in graph generation task.}
  \label{table:lambdaWeights}
  \centering
  \resizebox{\textwidth}{!}{
  \begin{tabular}{lcccccccc}
    \toprule
   Dataset & Triangle Grid & Lobster & Grid & ogbg-molbbbp & Protein & MUTAG & PTC & QM9\\
      \midrule
    $\tradeoff$ & $50$ & $40$ & $50$ & $40$ & $50$ & $4$& $2$& $80$ \\
   $\beta$ & $2e^{3}$ & $1.5e^{3}$ & $2e^{3}$ & $1.5e^{3}$ & $1e^{3}$   & $60$& $60$& $200$\\
    \bottomrule
  \end{tabular}%
 } 
\end{table}

\subsection{Qualitative evaluation of GGMs in detail} 
\label{section:Visualiztion}
\begin{figure}
\begin{tabular}{m{0.09\textwidth} m{0.19\textwidth} m{0.19\textwidth} m{0.19\textwidth} m{0.19\textwidth}}
\centering
    \begin{tabular}{l}
    \centering \scriptsize{Test}
  \end{tabular}  &   \includegraphics[width=30mm]{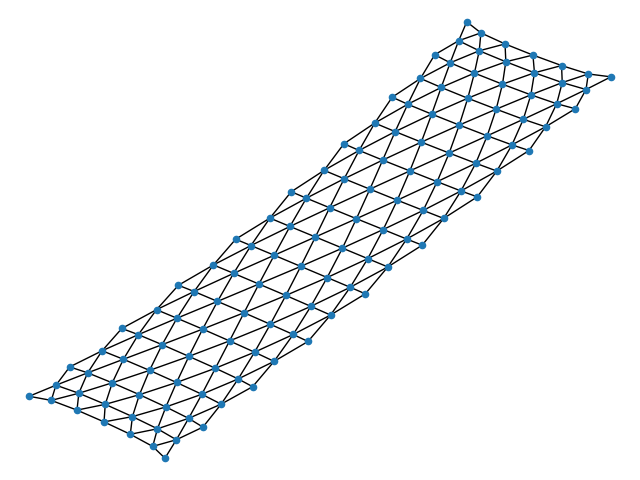} &   \includegraphics[width=30mm]{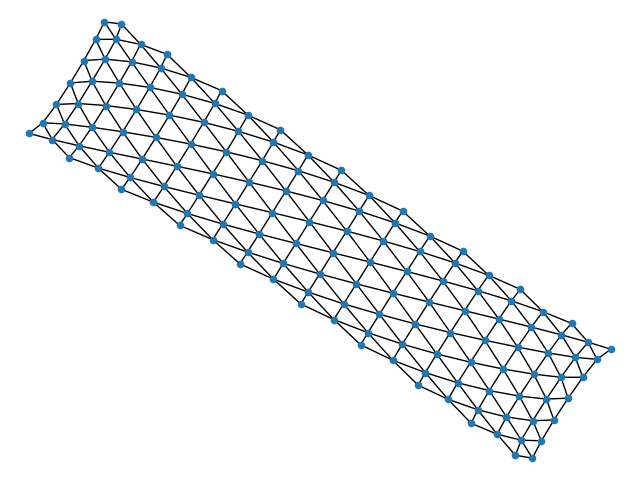}  &   \includegraphics[width=30mm]{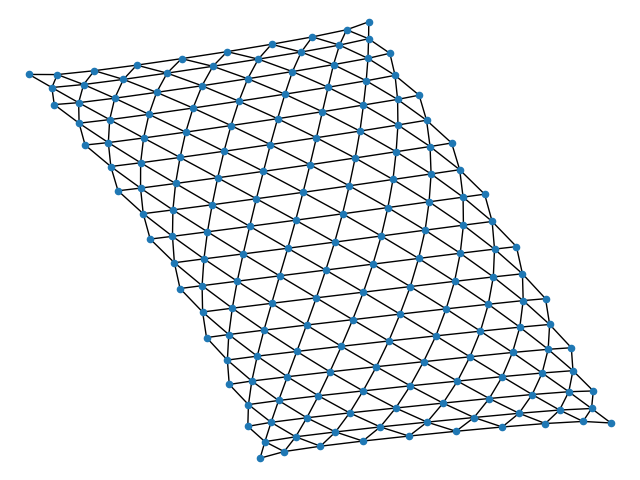} &   \includegraphics[width=30mm]{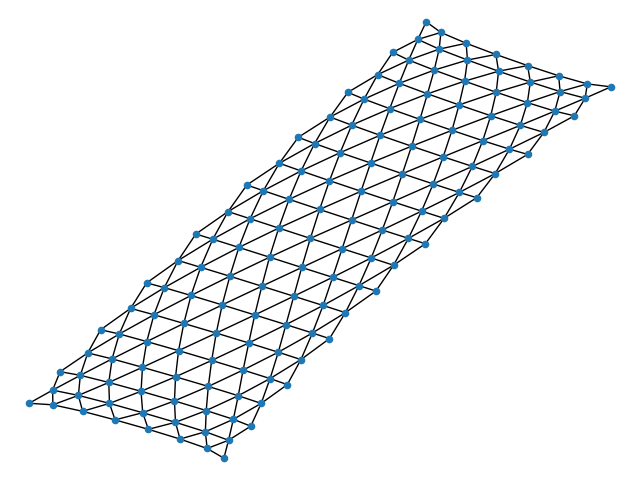}  \\ \hline
\centering
    \begin{tabular}{l}
    \centering \scriptsize{GraphVAE}
  \end{tabular}  &   \includegraphics[width=30mm]{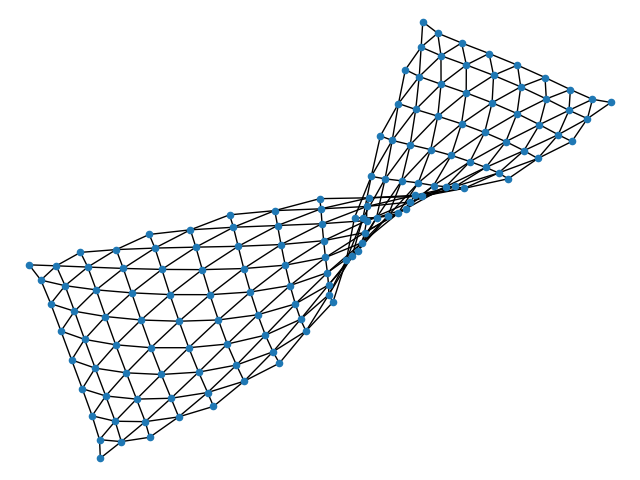} &   \includegraphics[width=30mm]{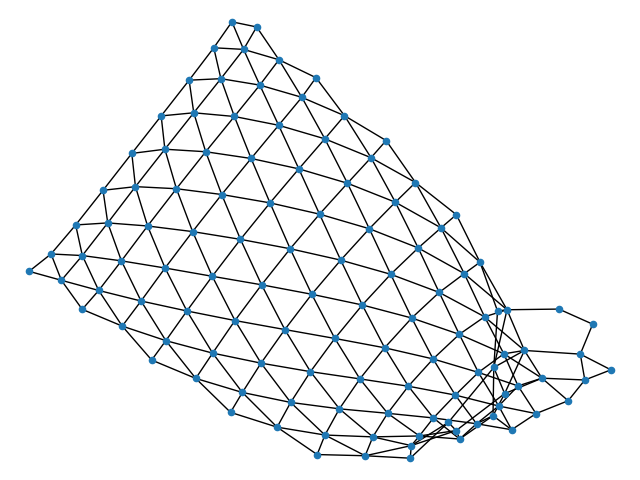}  &   \includegraphics[width=30mm]{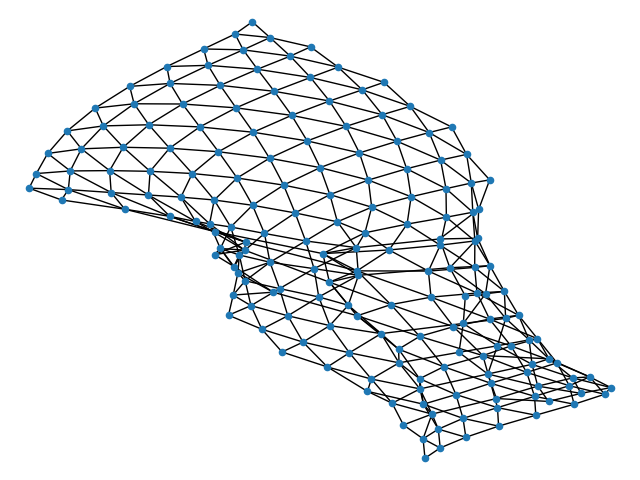} &   \includegraphics[width=30mm]{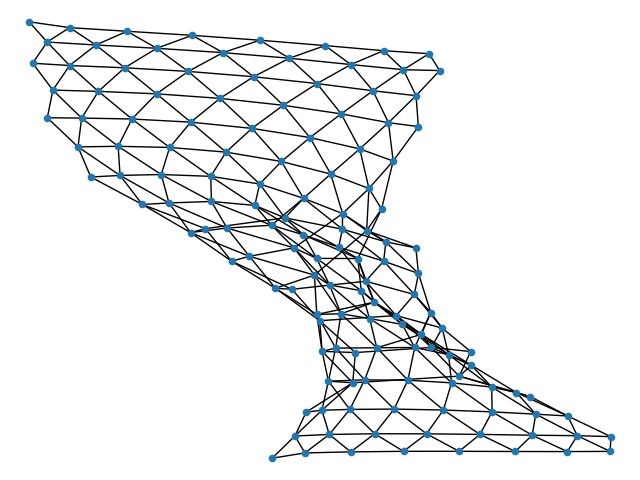}  \\
  \centering
    \begin{tabular}{l}
    \centering \scriptsize{\textbf{\tiny{GraphVAE\ourModelAcronym}}}
  \end{tabular}  &   \includegraphics[width=30mm]{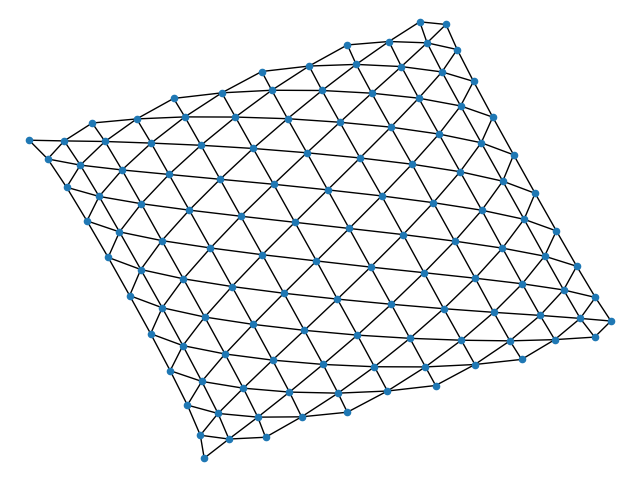} &   \includegraphics[width=30mm]{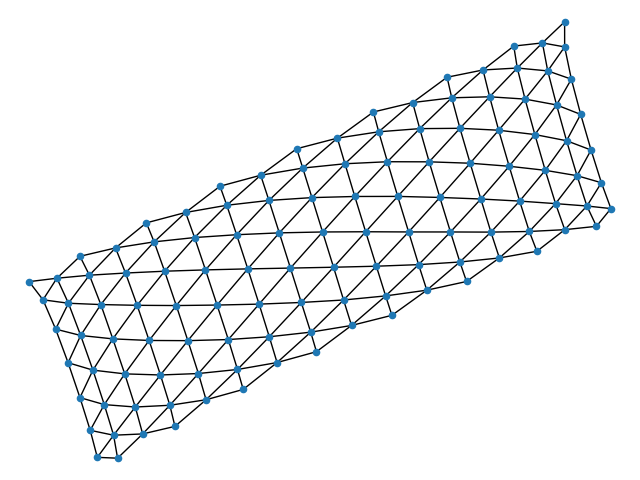}  &   \includegraphics[width=30mm]{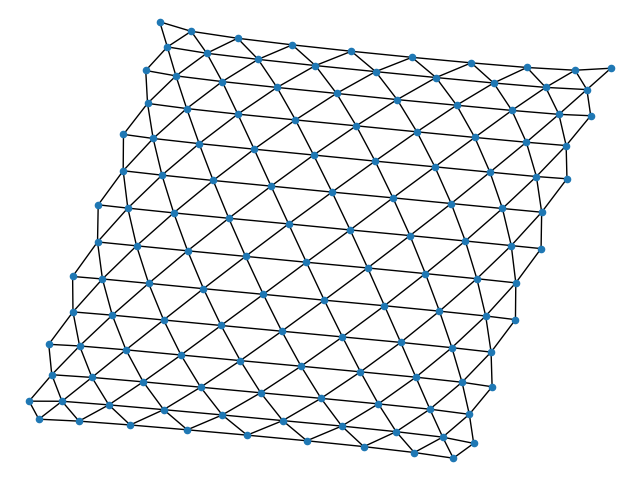} &   \includegraphics[width=30mm]{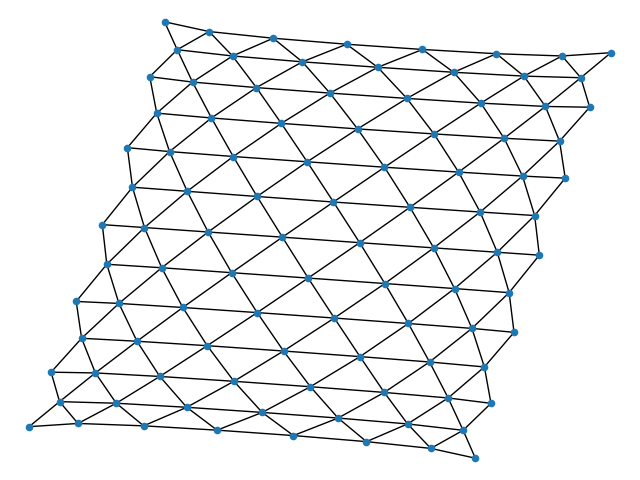}  \\ \hline
    \centering
    \begin{tabular}{l}
    \centering \scriptsize{BIGG}
  \end{tabular}  &   \includegraphics[width=30mm]{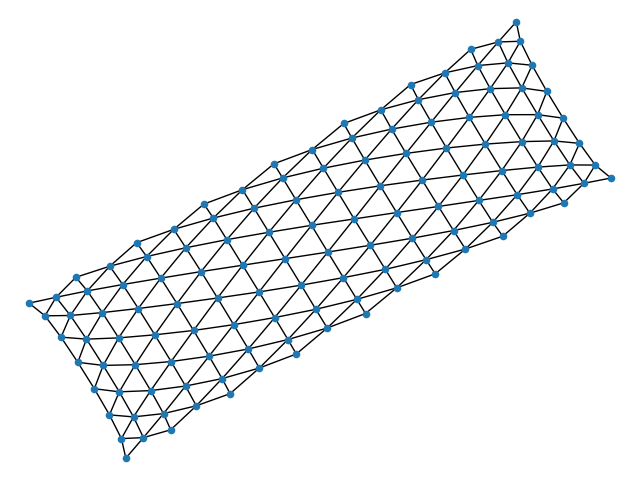} &   \includegraphics[width=30mm]{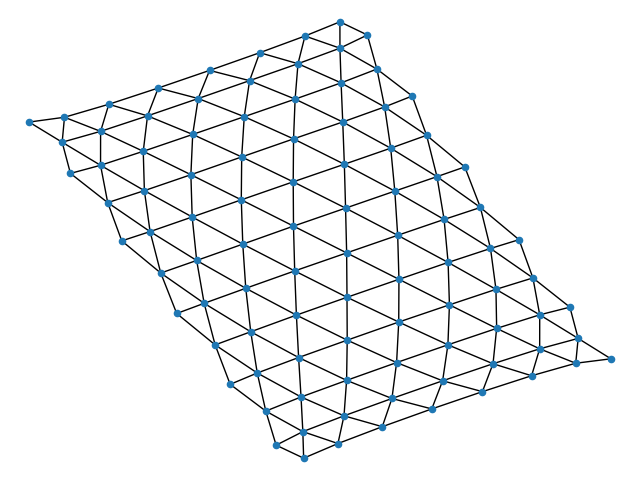}  &   \includegraphics[width=30mm]{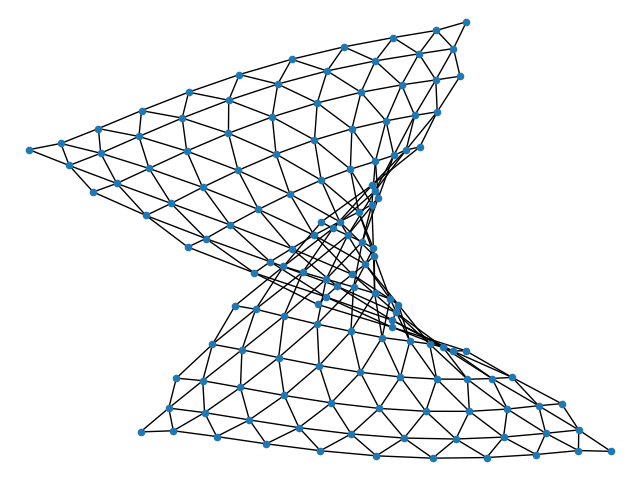} &   \includegraphics[width=30mm]{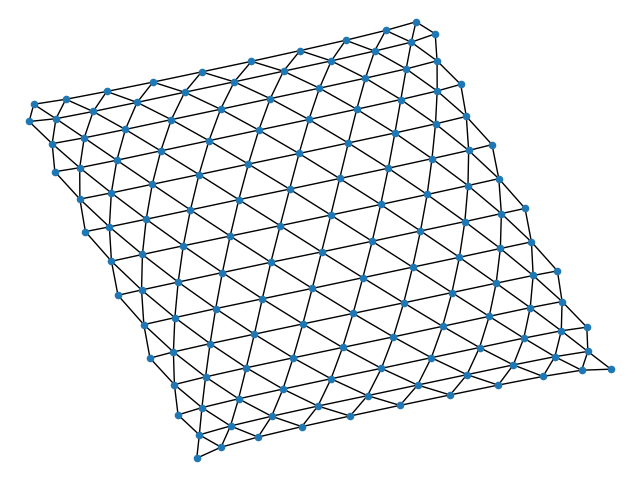}  \\
    \centering
    \begin{tabular}{l}
    \centering \pbox{15cm}{\scriptsize{GRAN}}
  \end{tabular}  &   \includegraphics[width=30mm]{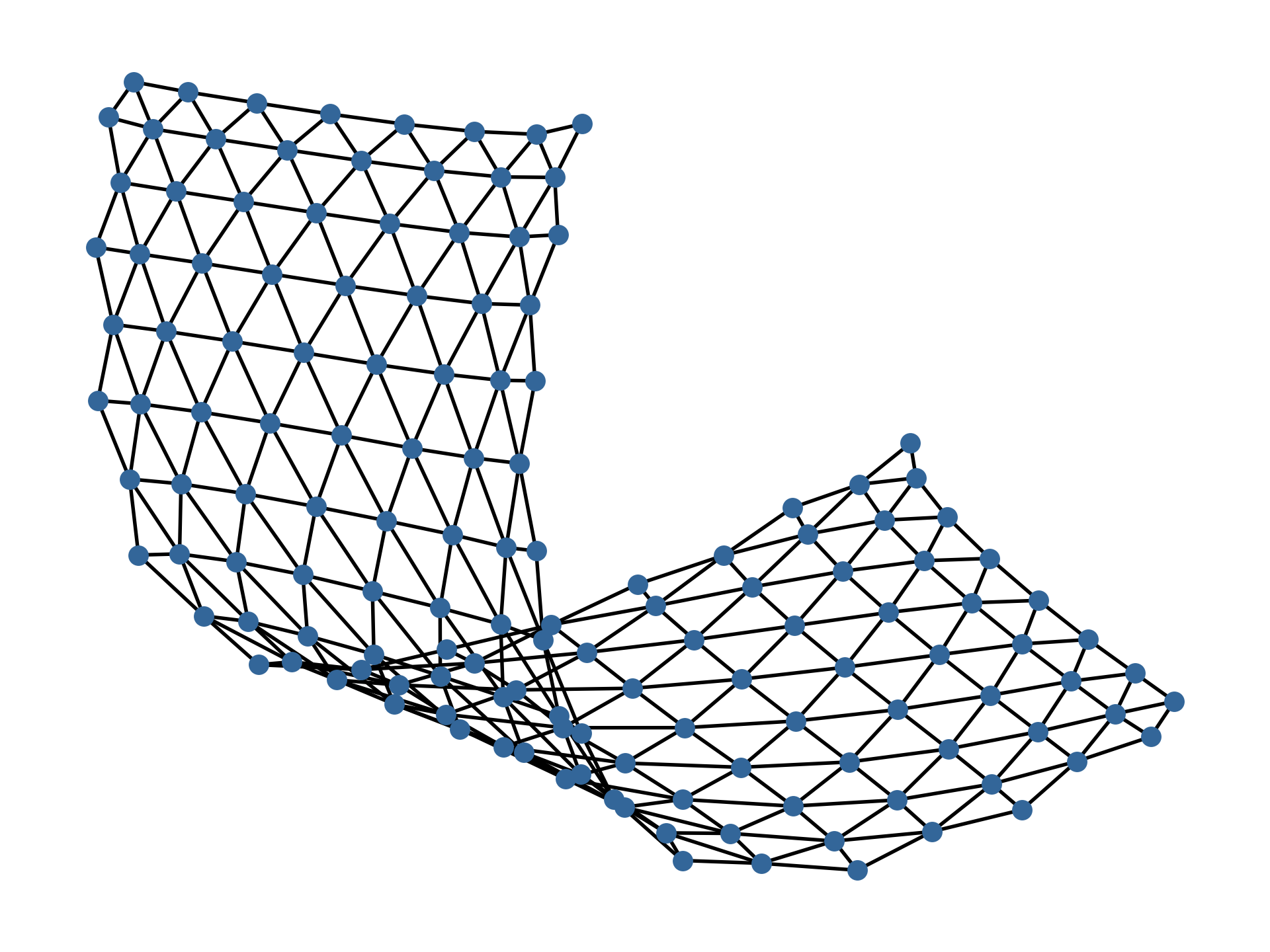} &   \includegraphics[width=30mm]{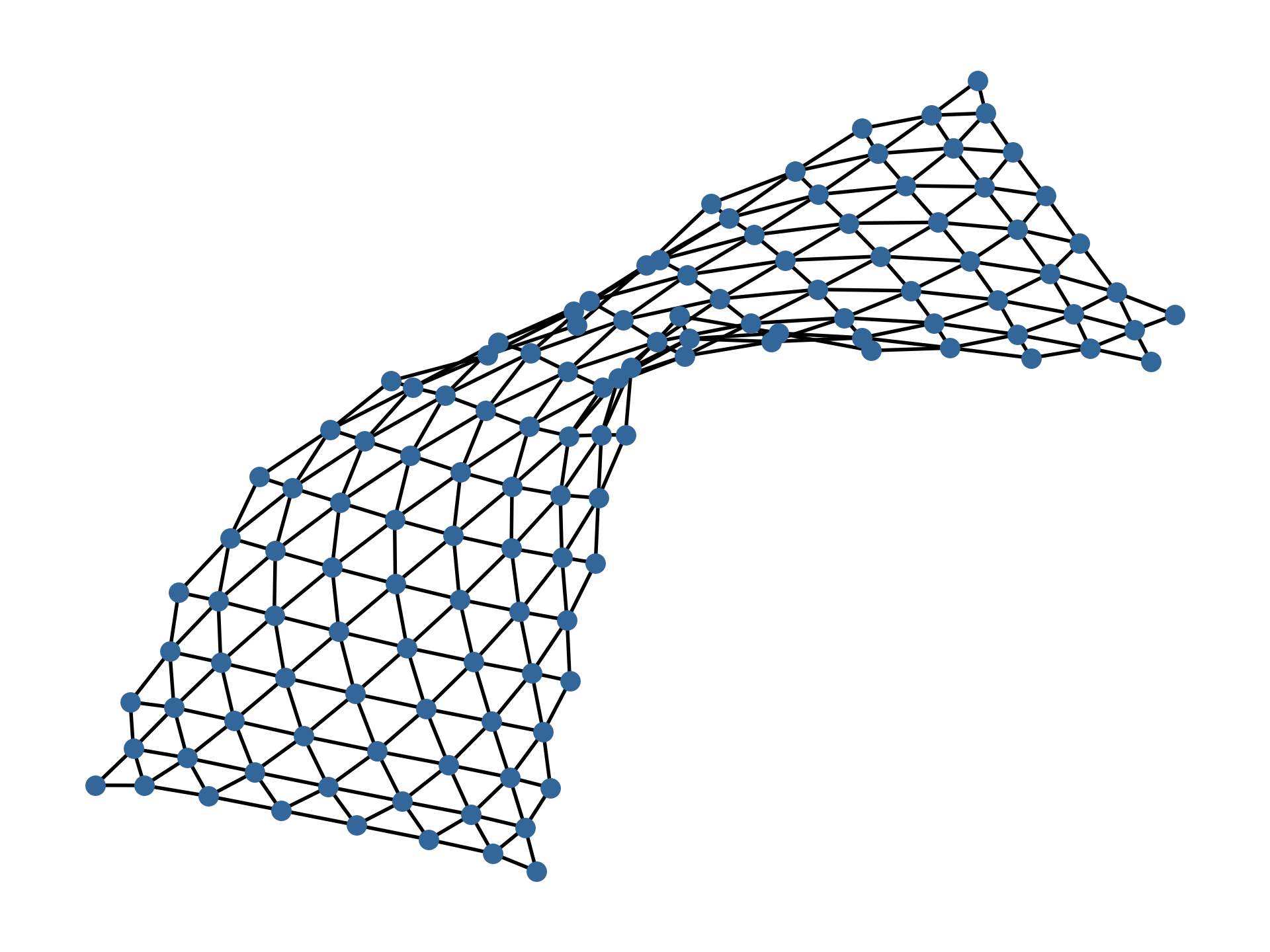}  &   \includegraphics[width=30mm]{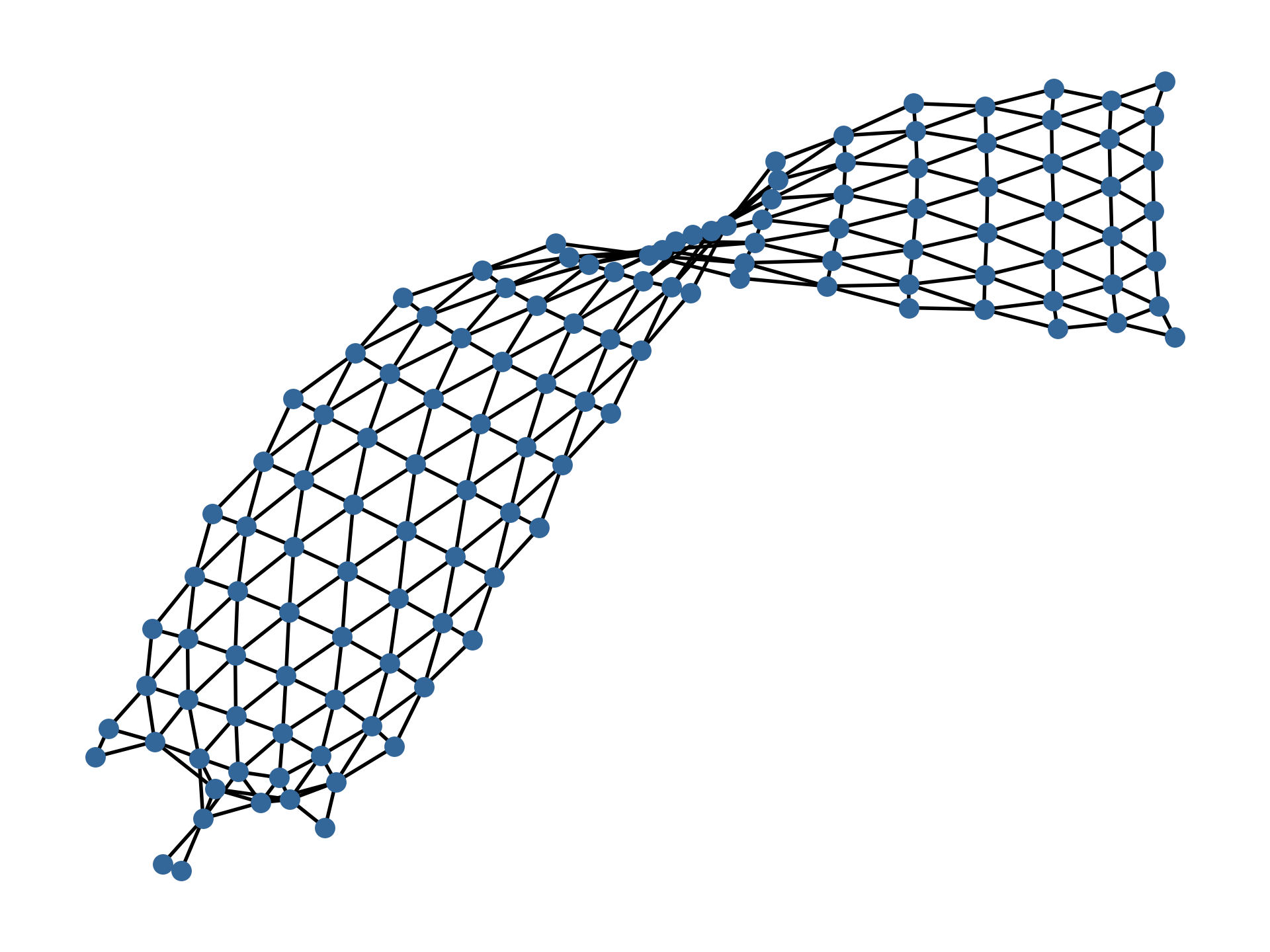} &   \includegraphics[width=30mm]{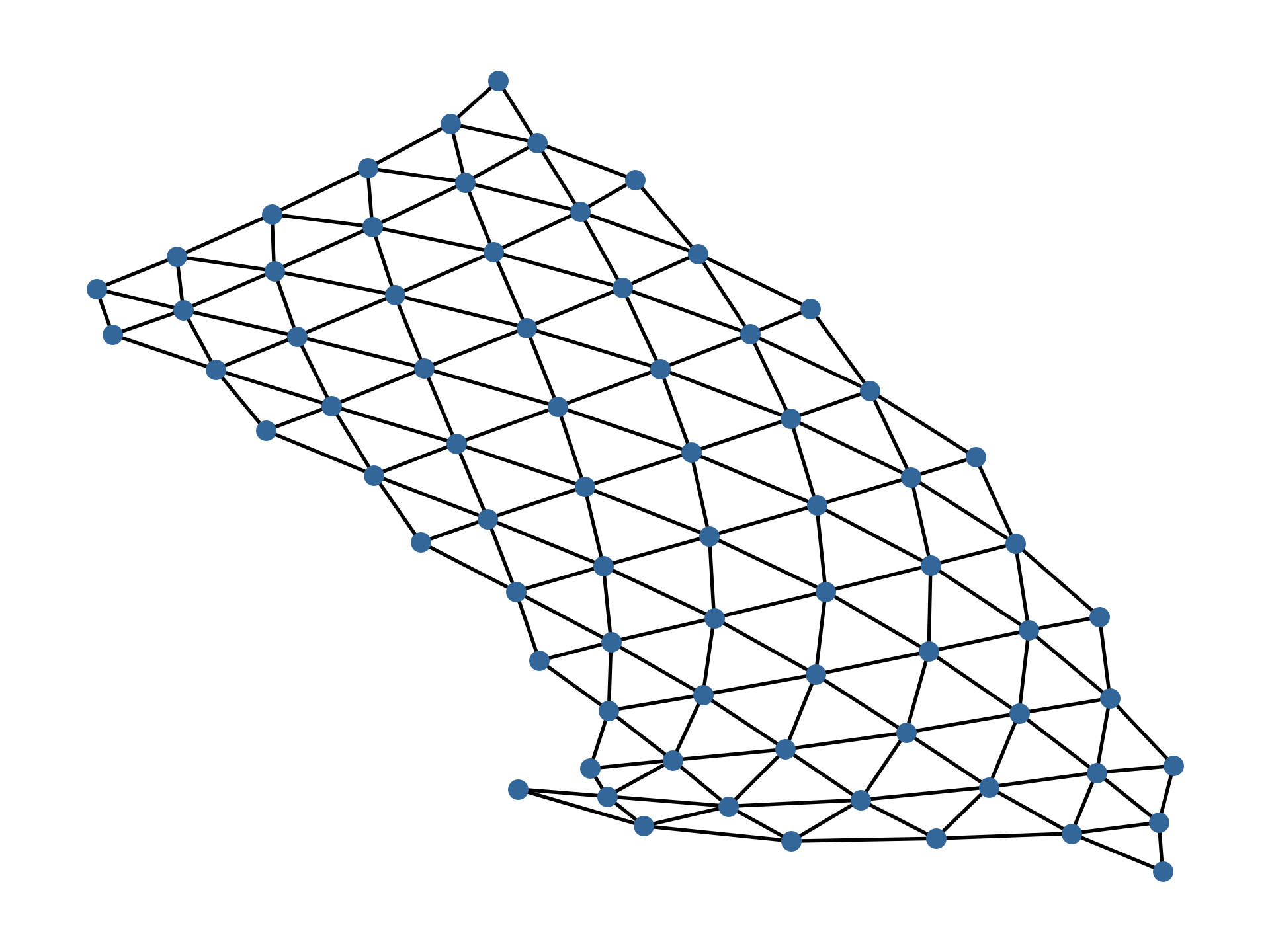}  \\
  \centering
    \begin{tabular}{l}
    \centering \scriptsize{GraphRNN}
  \end{tabular}  &   \includegraphics[width=30mm]{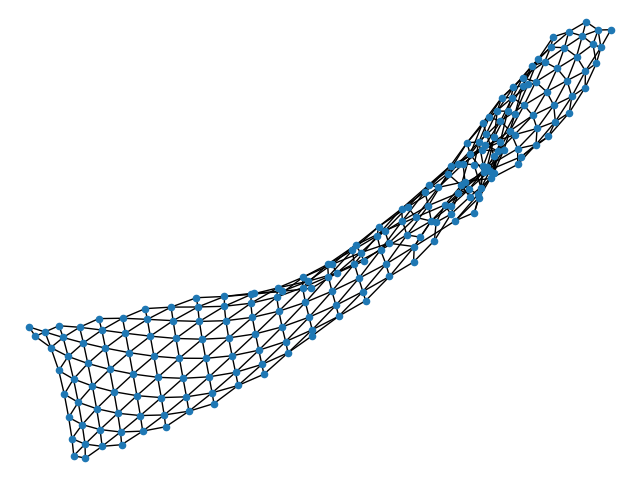} &   \includegraphics[width=30mm]{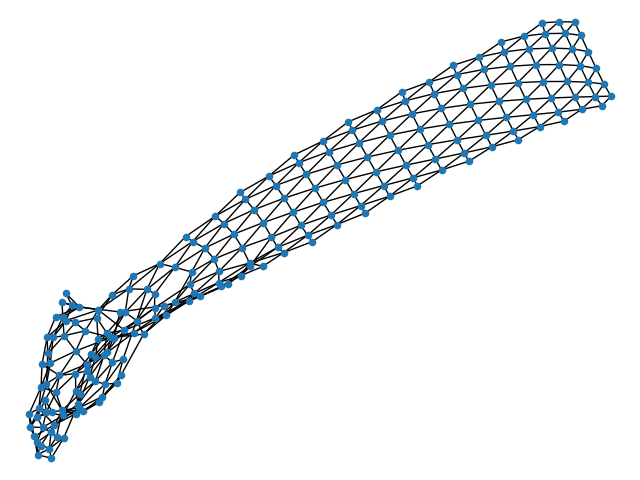}  &   \includegraphics[width=30mm]{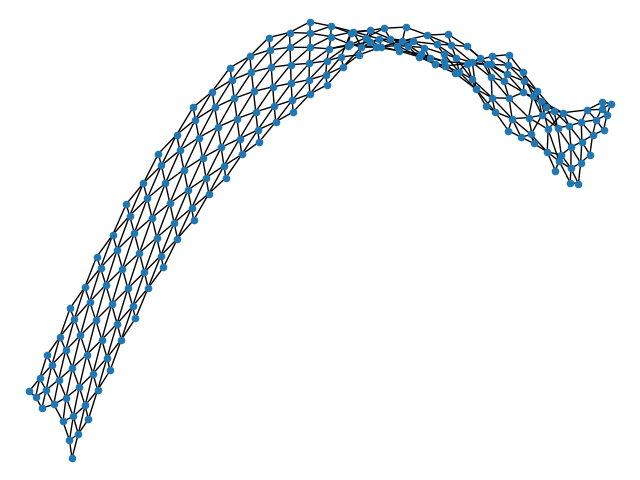} &   \includegraphics[width=30mm]{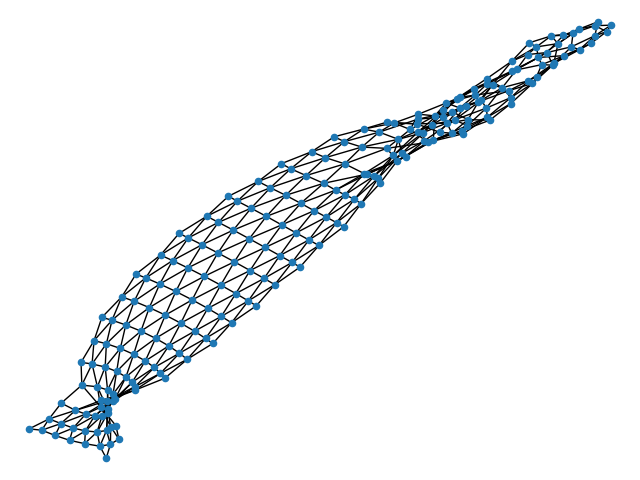}  \\
    \centering
    \begin{tabular}{l}
    \centering \pbox{15cm}{\scriptsize{GraphRNN-S}}
  \end{tabular}  &   \includegraphics[width=30mm]{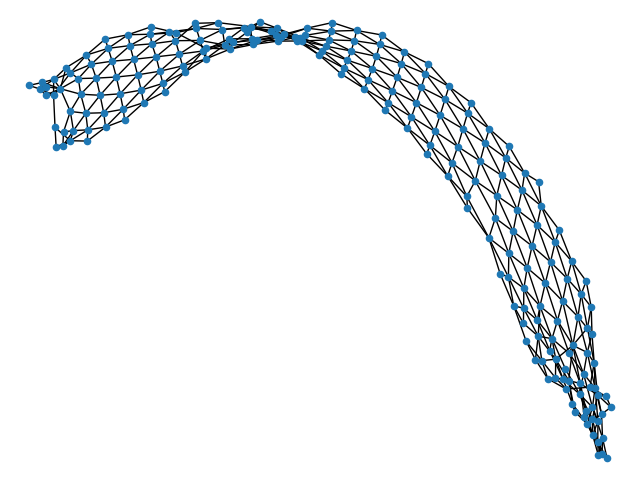} &   \includegraphics[width=30mm]{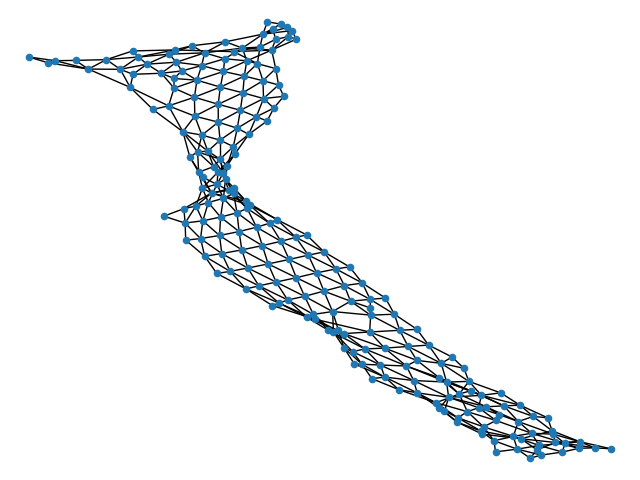}  &   \includegraphics[width=30mm]{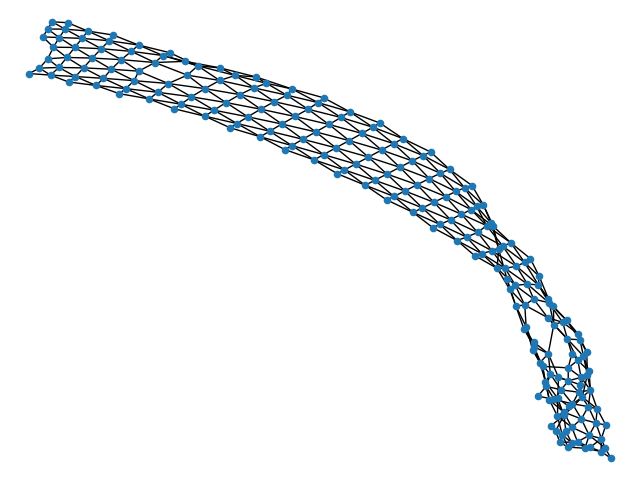} &   \includegraphics[width=30mm]{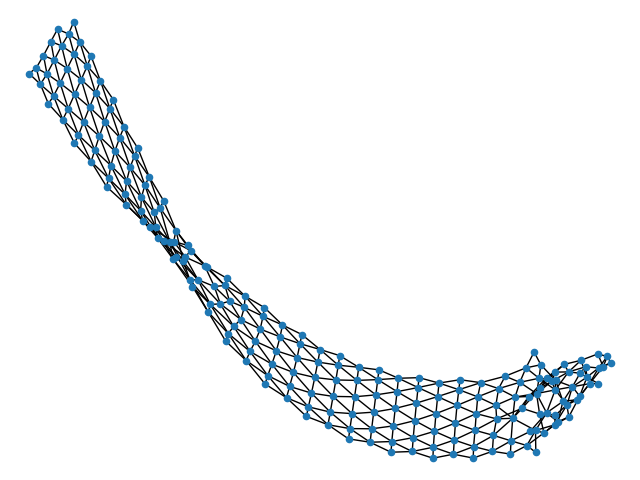}  \\
\end{tabular}
\caption{Visualization of generated \textbf{Triangle Grid} graphs by  benchmark GGMs and the micro-macro modeling effect. The first block shows four randomly selected graphs from the test set. The first and second rows in the second block show samples generated by GraphVAE and GraphVAE\ourModelAcronym~models respectively. The bottom block shows graphs generated  by benchmark GGMs. Graphs generated with micro-macro modeling, GraphVAE\ourModelAcronym, match the target graph the best and make a noticeable improvement in comparison to GraphVAE. For each model we visually select and visualize the most similar generated samples to the test set.
}
\label{fig:TriGridVisualization2}
\end{figure}

\begin{figure}
\begin{tabular}{m{0.09\textwidth} m{0.19\textwidth} m{0.19\textwidth} m{0.19\textwidth} m{0.19\textwidth}}
\centering
    \begin{tabular}{l}
    \centering \scriptsize{Test}
  \end{tabular}  &   \includegraphics[width=30mm]{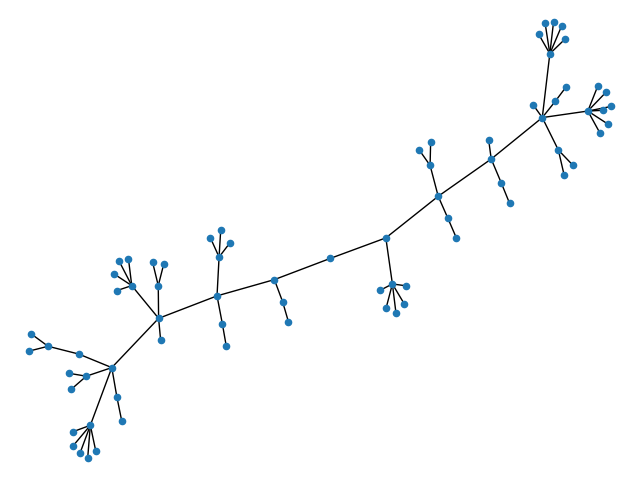} &   \includegraphics[width=30mm]{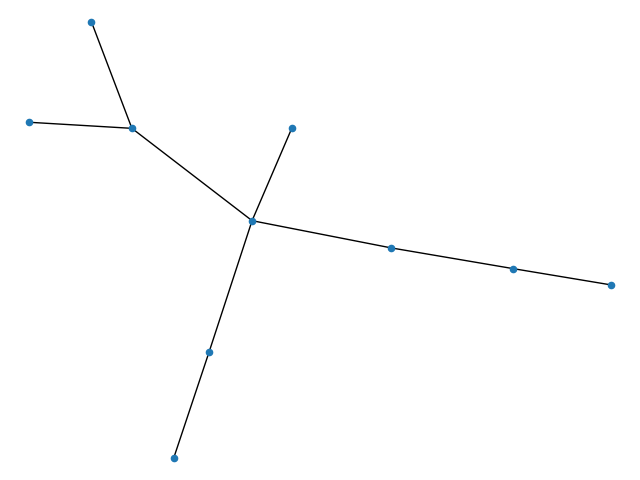}  &   \includegraphics[width=30mm]{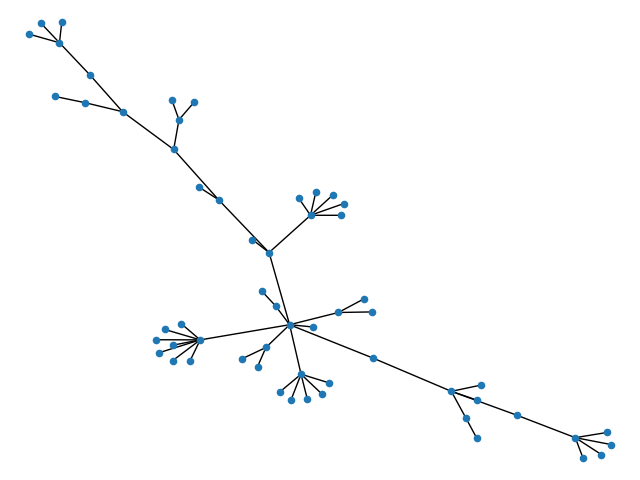} &   \includegraphics[width=30mm]{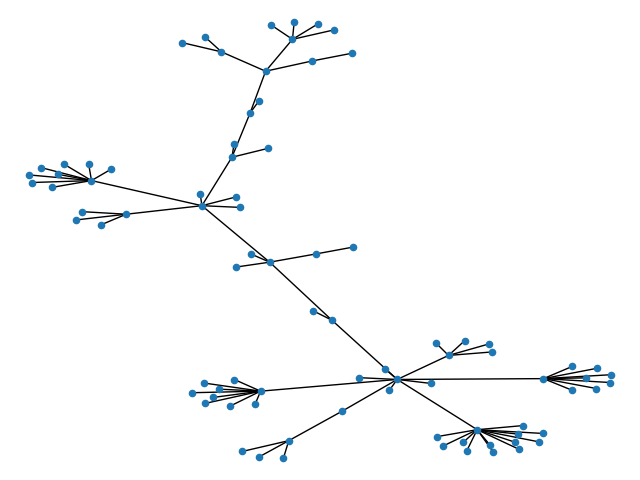}  \\ \hline
\centering
    \begin{tabular}{l}
    \centering \scriptsize{GraphVAE}
  \end{tabular}  &   \includegraphics[width=30mm]{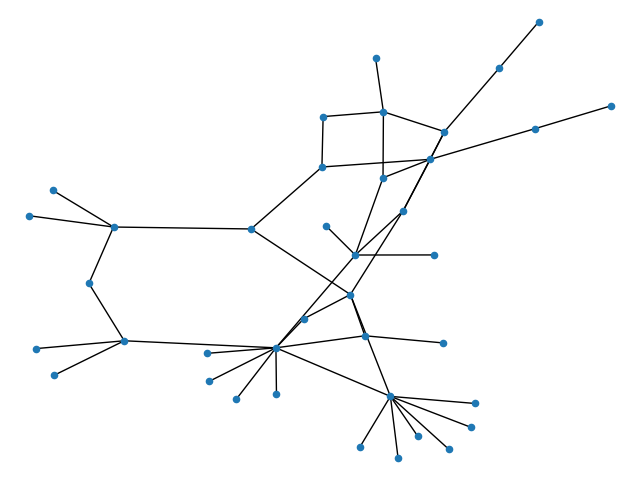} &   \includegraphics[width=30mm]{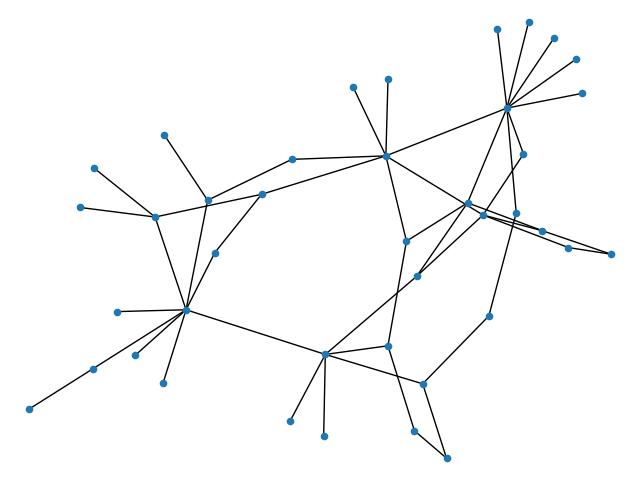}  &   \includegraphics[width=30mm]{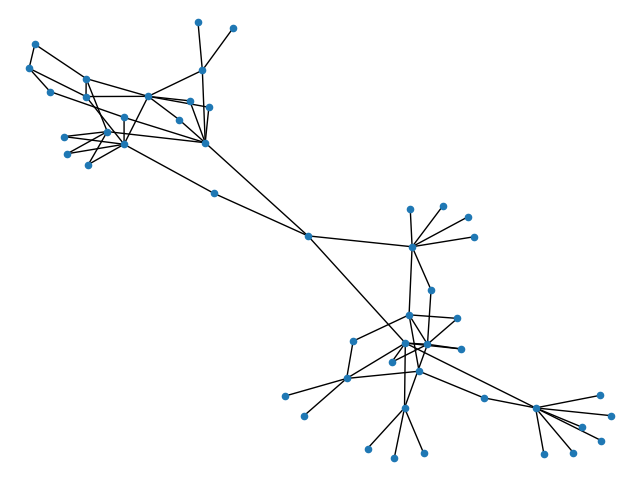} &   \includegraphics[width=30mm]{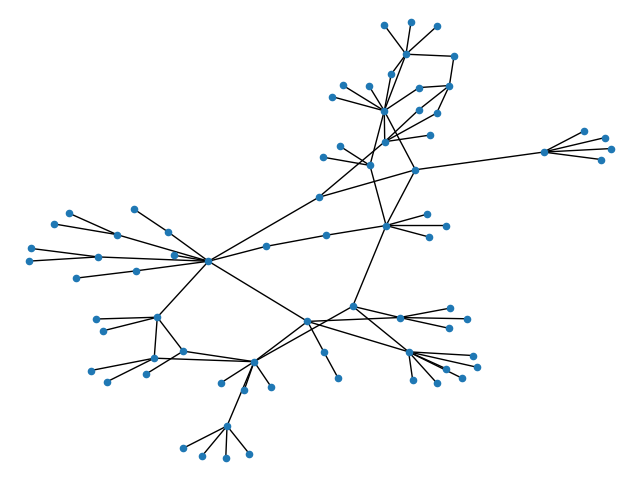}  \\
  \centering
    \begin{tabular}{l}
    \centering \scriptsize{\textbf{\tiny{GraphVAE\ourModelAcronym}}}
  \end{tabular}  &   \includegraphics[width=30mm]{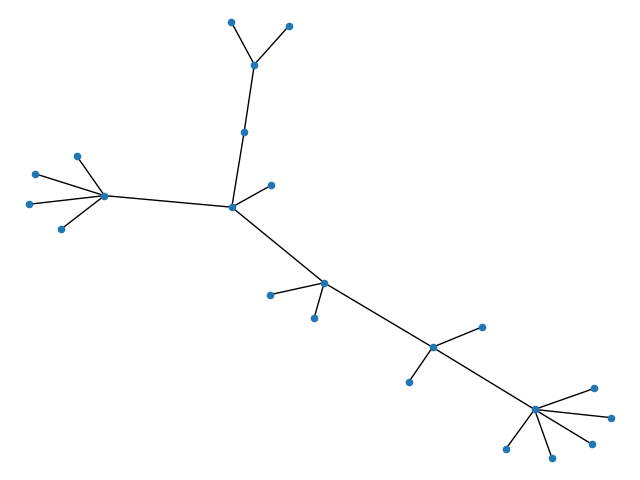} &   \includegraphics[width=30mm]{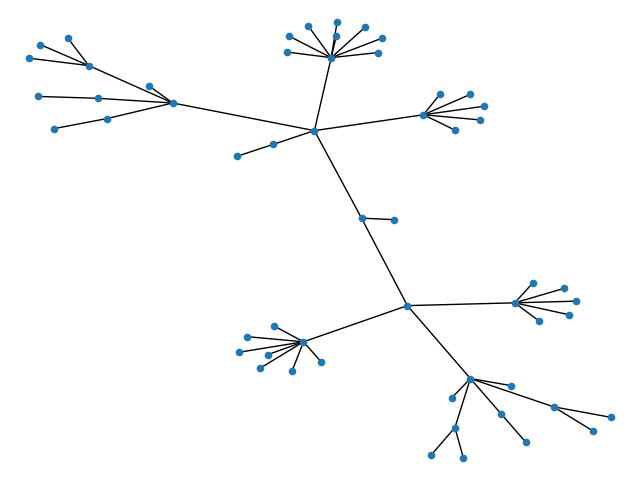}  &   \includegraphics[width=30mm]{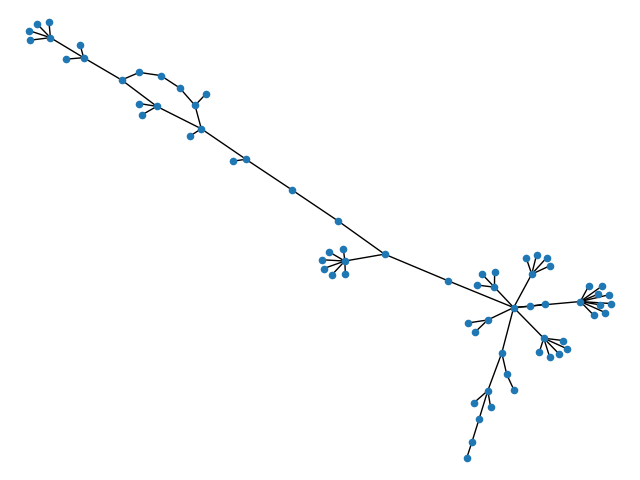} &   \includegraphics[width=30mm]{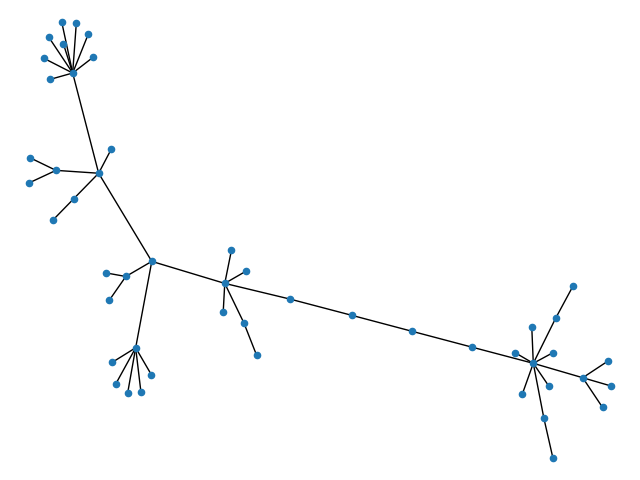}  \\ \hline
    \centering
    \begin{tabular}{l}
    \centering \scriptsize{BIGG}
  \end{tabular}  &   \includegraphics[width=30mm]{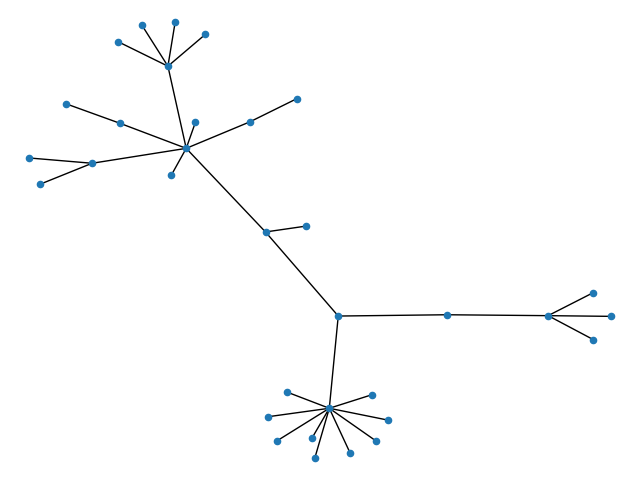} &   \includegraphics[width=30mm]{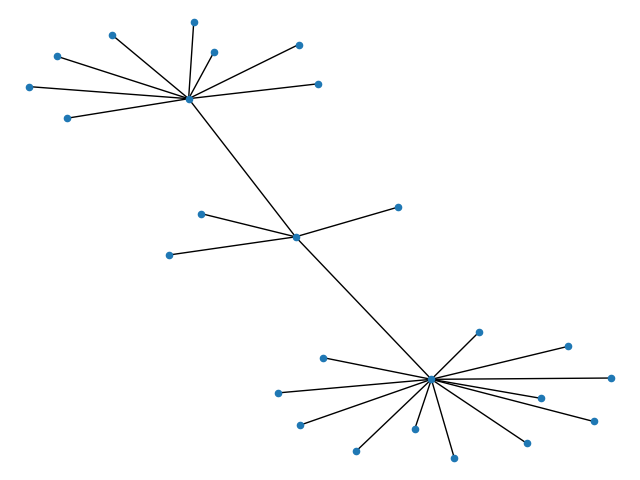}  &   \includegraphics[width=30mm]{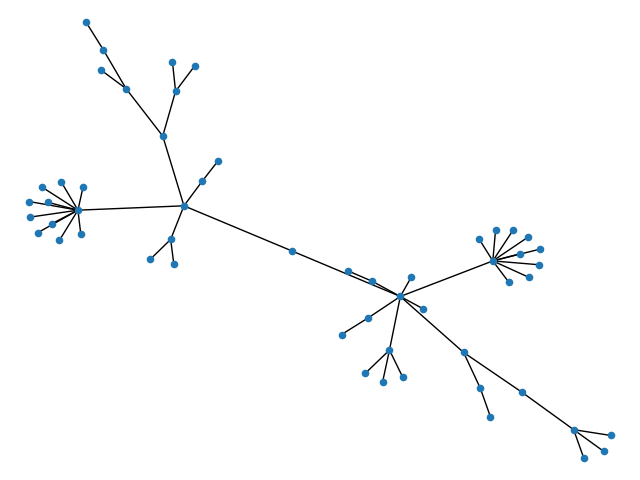} &   \includegraphics[width=30mm]{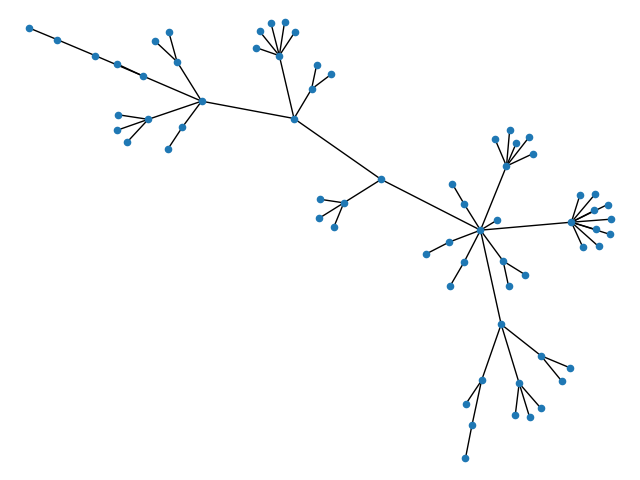}  \\
    \centering
    \begin{tabular}{l}
    \centering \pbox{15cm}{\scriptsize{GRAN}}
  \end{tabular}  &   \includegraphics[width=30mm]{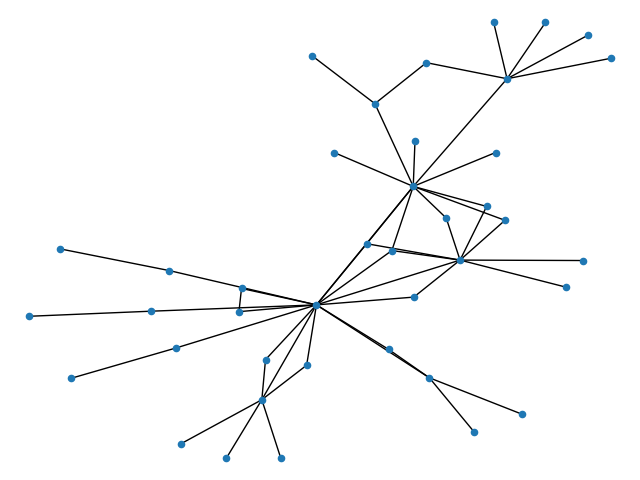} &   \includegraphics[width=30mm]{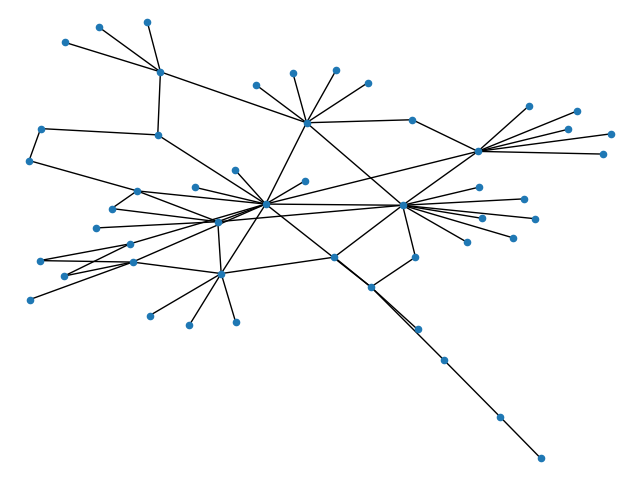}  &   \includegraphics[width=30mm]{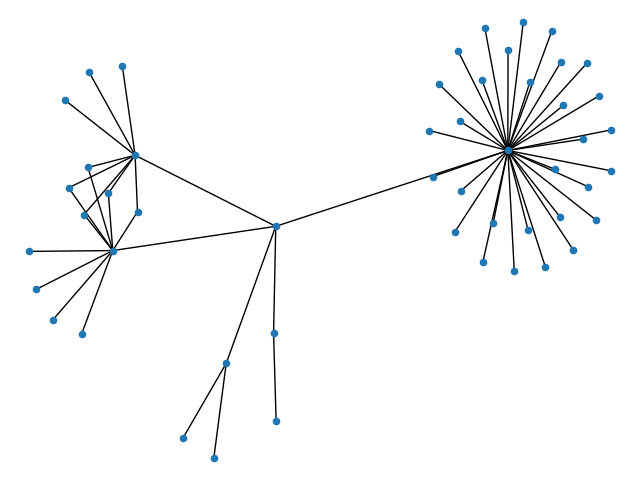} &   \includegraphics[width=30mm]{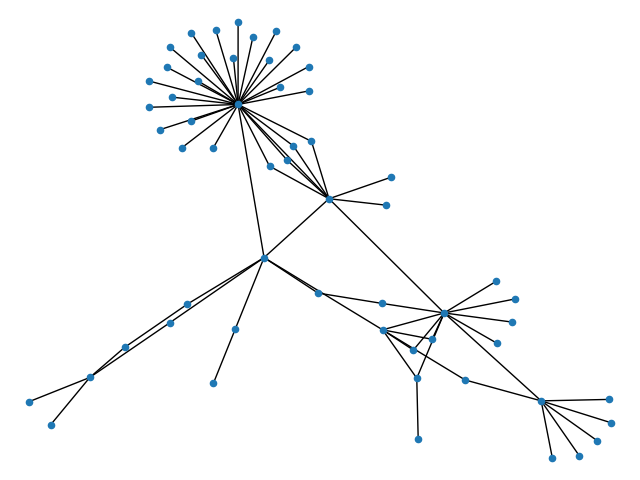}  \\
  \centering
    \begin{tabular}{l}
    \centering \scriptsize{GraphRNN}
  \end{tabular}  &   \includegraphics[width=30mm]{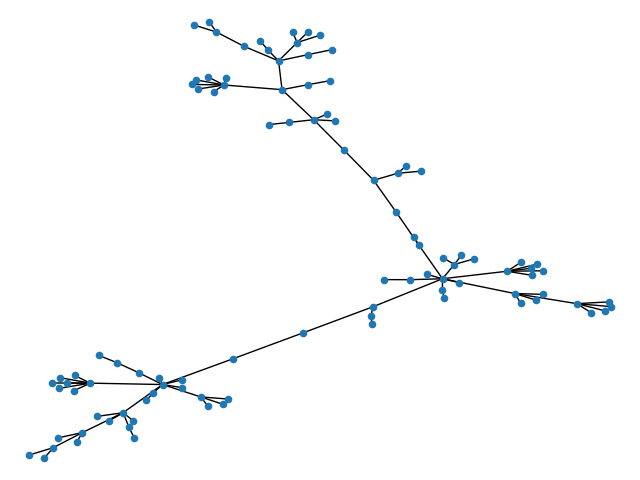} &   \includegraphics[width=30mm]{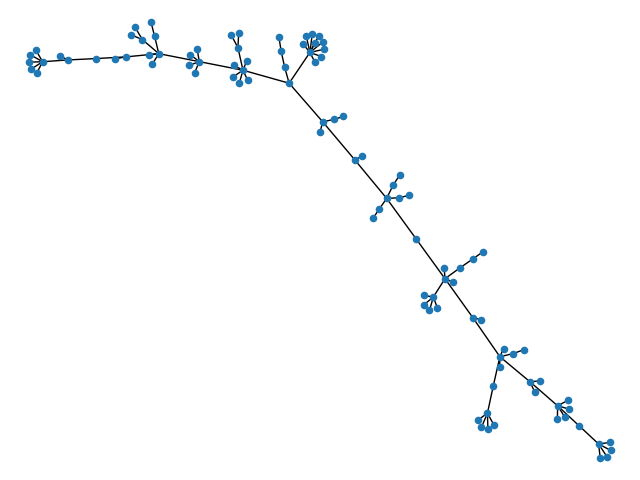}  &   \includegraphics[width=30mm]{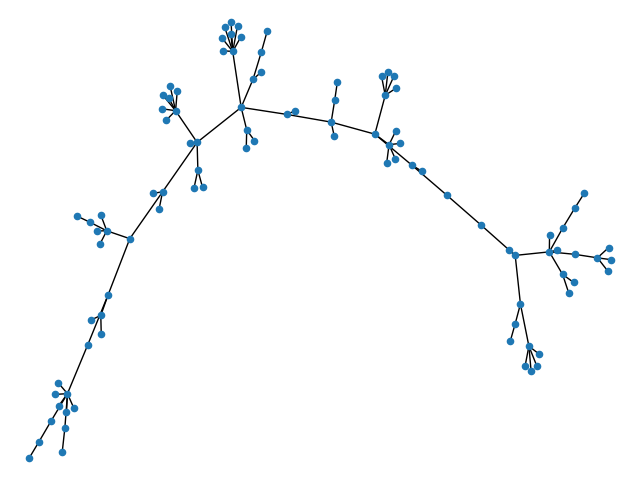} &   \includegraphics[width=30mm]{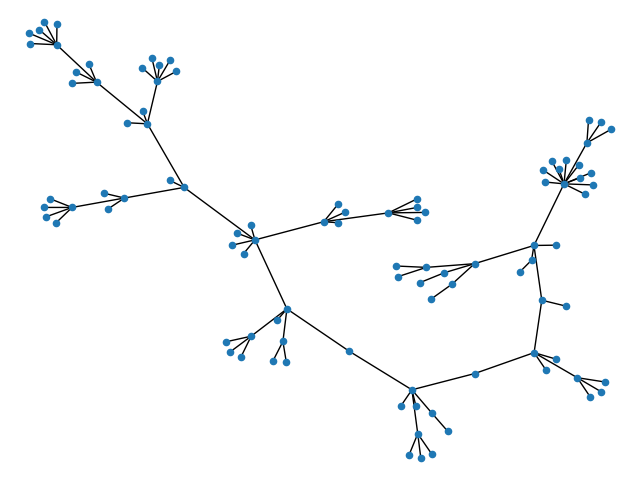}  \\
    \centering
    \begin{tabular}{l}
    \centering \pbox{15cm}{\scriptsize{GraphRNN-S}}
  \end{tabular}  &   \includegraphics[width=30mm]{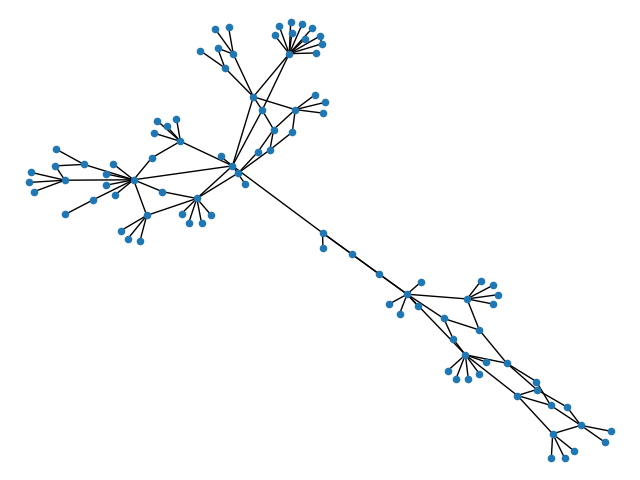} &   \includegraphics[width=30mm]{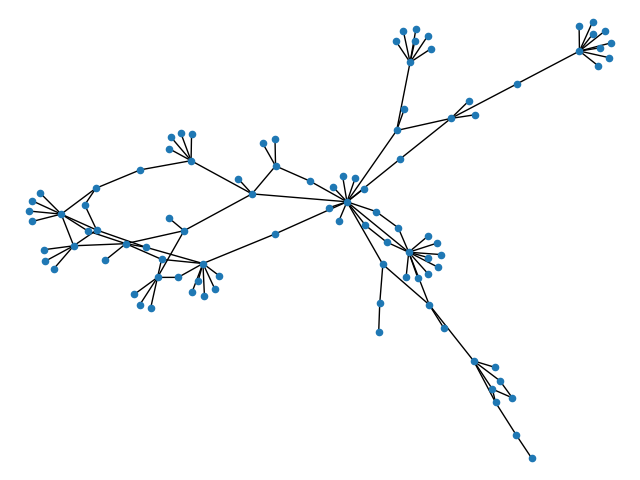}  &   \includegraphics[width=30mm]{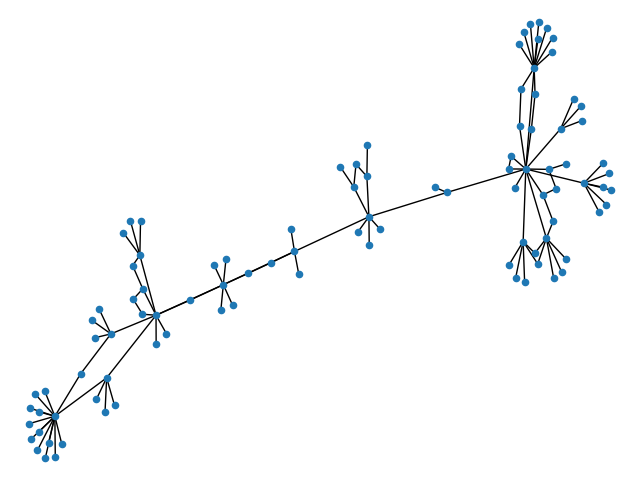} &   \includegraphics[width=30mm]{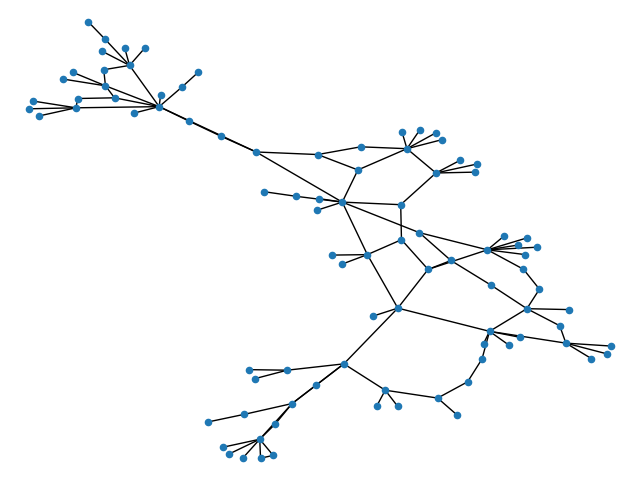}  \\
\end{tabular}
\caption{Visualization of generated \textbf{Lobster} graphs by  benchmark GGMs and the micro-macro modeling effect. The first block shows four randomly selected graphs from the test set. The first and second rows in the second block show samples generated by GraphVAE and GraphVAE\ourModelAcronym~models respectively. The bottom block shows graphs generated  by benchmark GGMs. Graphs generated with micro-macro modeling, GraphVAE\ourModelAcronym, match the target graph the best and make a noticeable improvement in comparison to GraphVAE. For each model we visually select and visualize the most similar generated samples to the test set.
}
\label{fig:lobster-Visualization2}
\end{figure}

\begin{figure}
\begin{tabular}{m{0.09\textwidth} m{0.19\textwidth} m{0.19\textwidth} m{0.19\textwidth} m{0.19\textwidth}}
\centering
    \begin{tabular}{l}
    \centering \scriptsize{Test}
  \end{tabular}  &   \includegraphics[width=30mm]{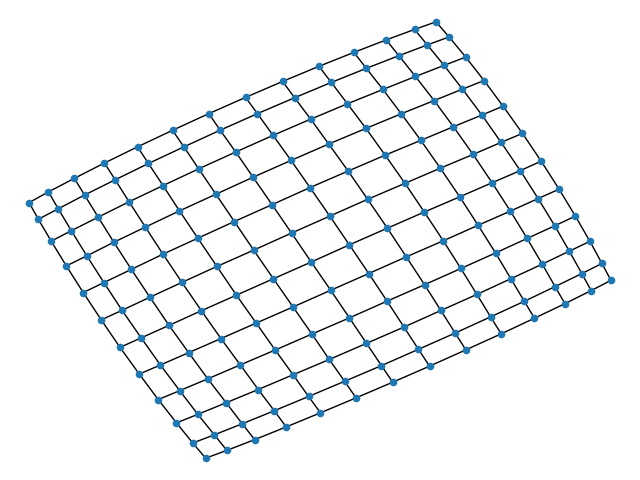} &   \includegraphics[width=30mm]{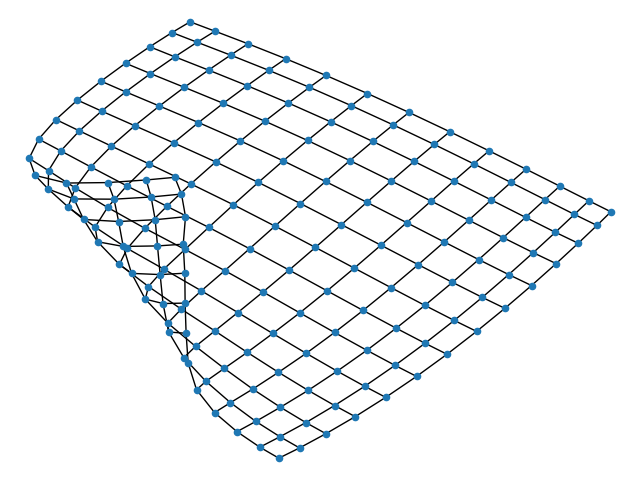}  &   \includegraphics[width=30mm]{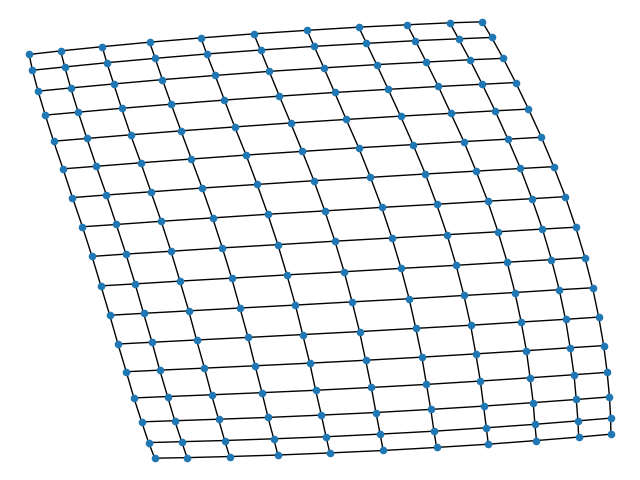} &   \includegraphics[width=30mm]{Graphs/grid/test/test13.png}  \\ \hline
\centering
    \begin{tabular}{l}
    \centering \scriptsize{GraphVAE}
  \end{tabular}  &   \includegraphics[width=30mm]{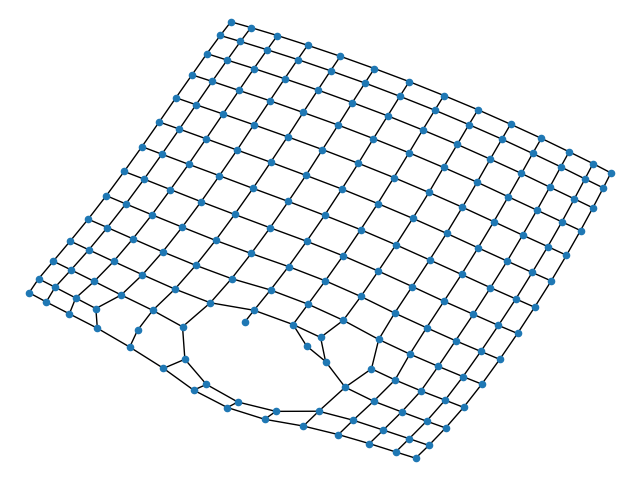} &   \includegraphics[width=30mm]{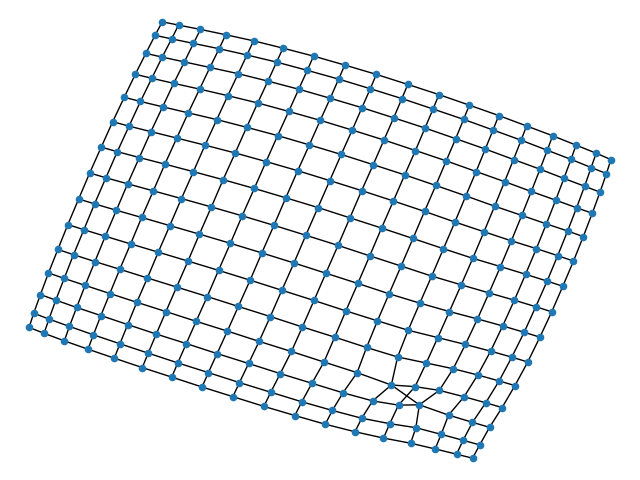}  &   \includegraphics[width=30mm]{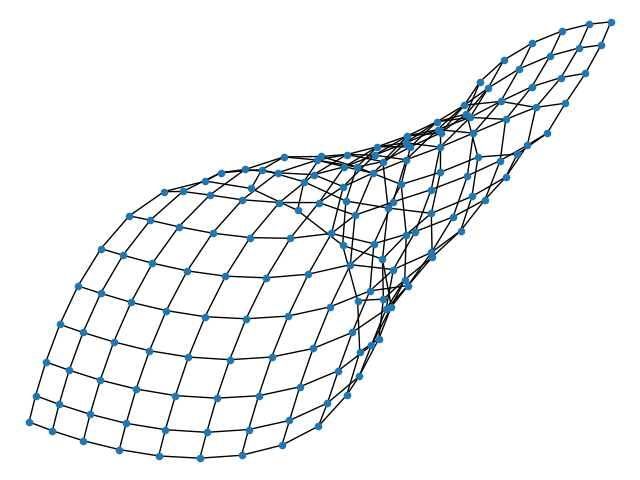} &   \includegraphics[width=30mm]{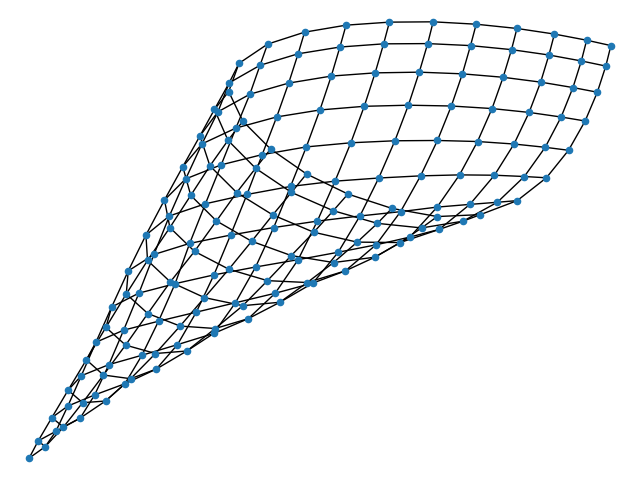}  \\
  \centering
    \begin{tabular}{l}
    \centering \scriptsize{\textbf{\tiny{GraphVAE\ourModelAcronym}}}
  \end{tabular}  &   \includegraphics[width=30mm]{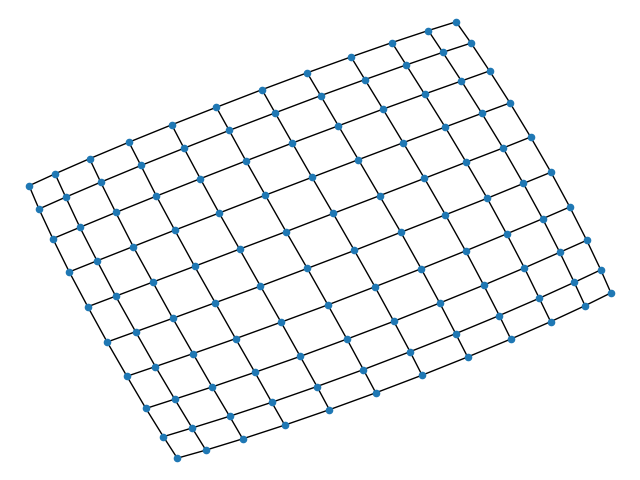} &   \includegraphics[width=30mm]{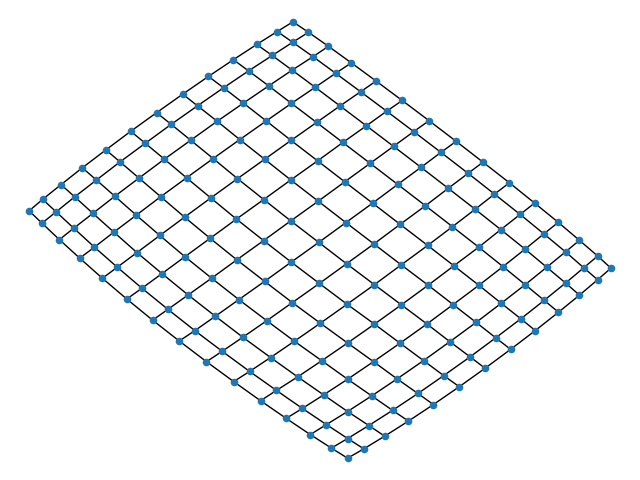}  &   \includegraphics[width=30mm]{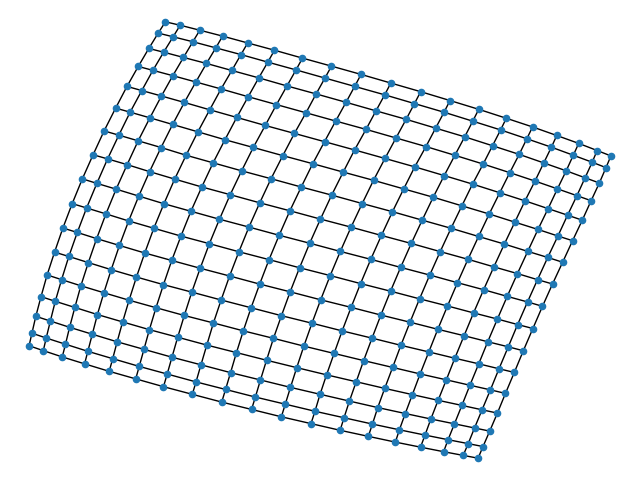} &   \includegraphics[width=30mm]{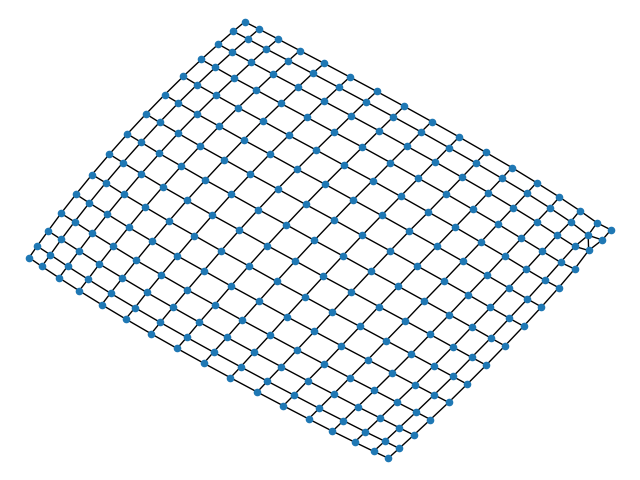}  \\ \hline
    \centering
    \begin{tabular}{l}
    \centering \scriptsize{BIGG}
  \end{tabular}  &   \includegraphics[width=30mm]{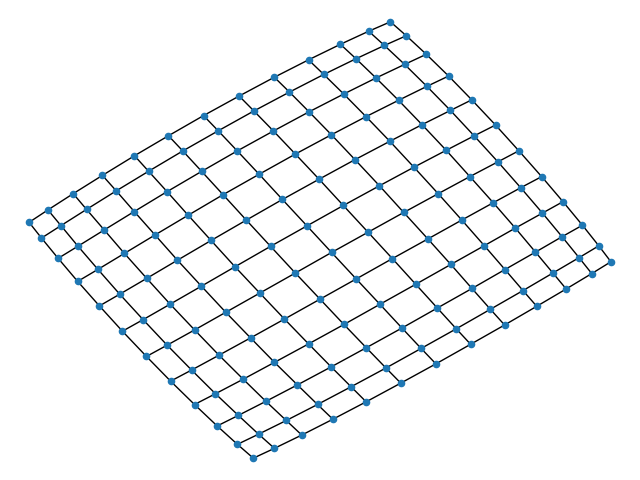} &   \includegraphics[width=30mm]{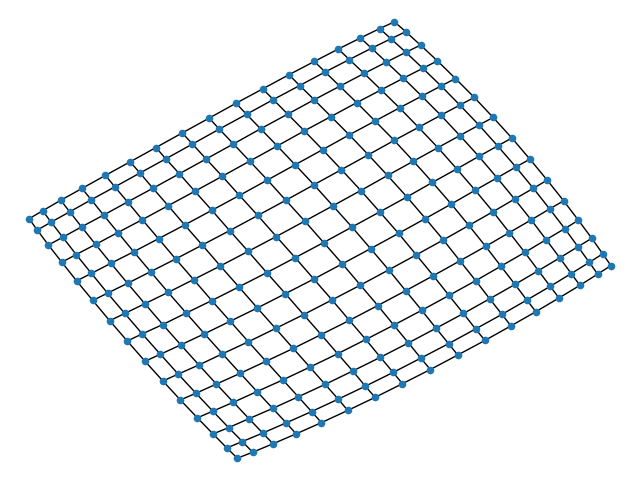}  &   \includegraphics[width=30mm]{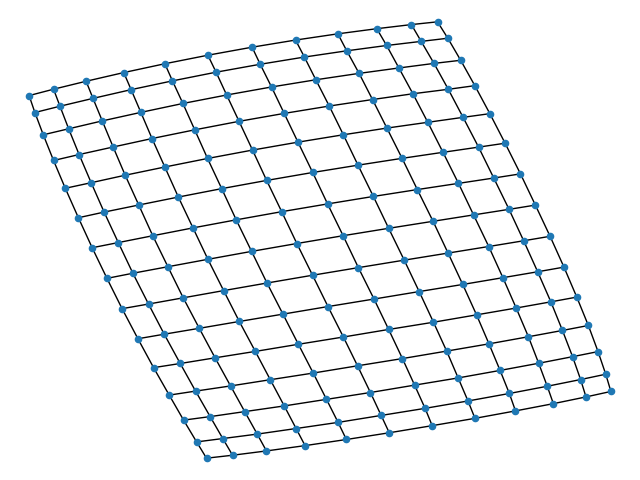} &   \includegraphics[width=30mm]{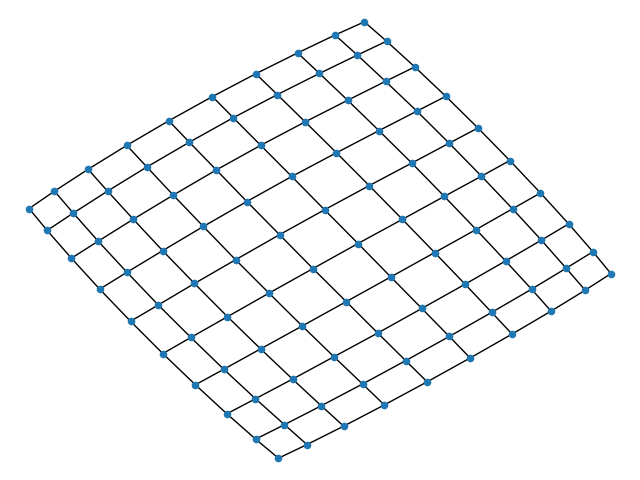}  \\
    \centering
    \begin{tabular}{l}
    \centering \pbox{15cm}{\scriptsize{GRAN}}
  \end{tabular}  &   \includegraphics[width=30mm]{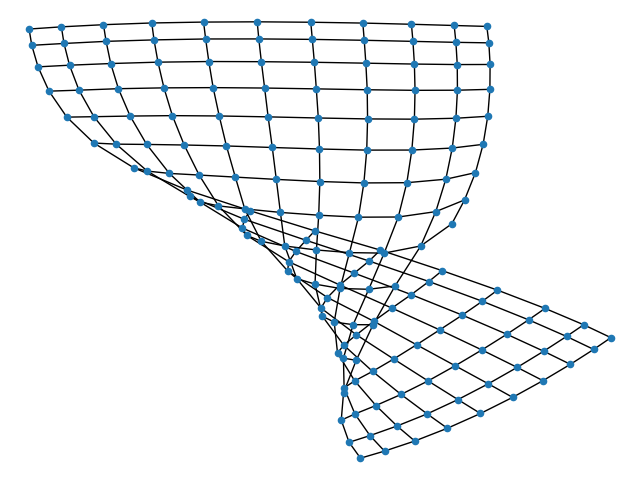} &   \includegraphics[width=30mm]{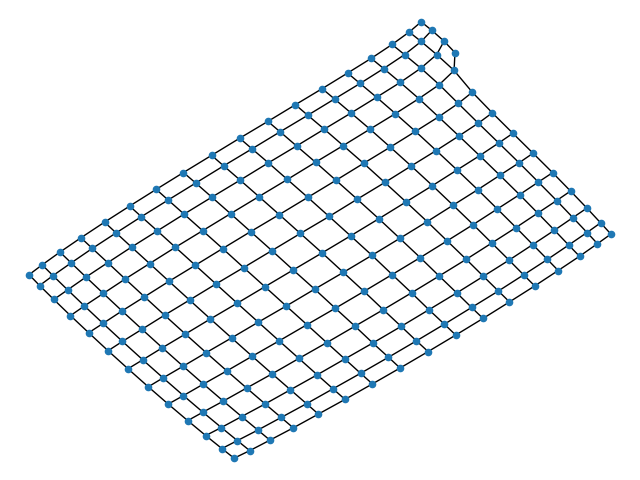}  &   \includegraphics[width=30mm]{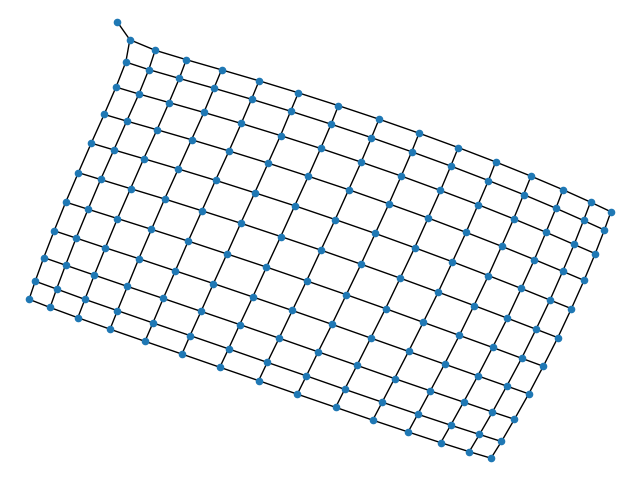} &   \includegraphics[width=30mm]{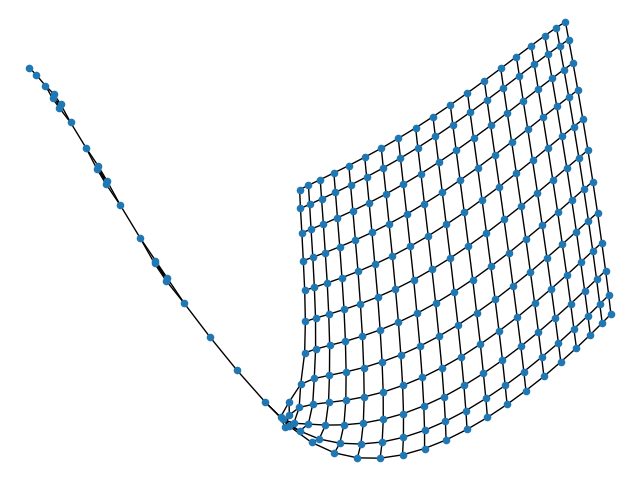}  \\
  \centering
    \begin{tabular}{l}
    \centering \scriptsize{GraphRNN}
  \end{tabular}  &   \includegraphics[width=30mm]{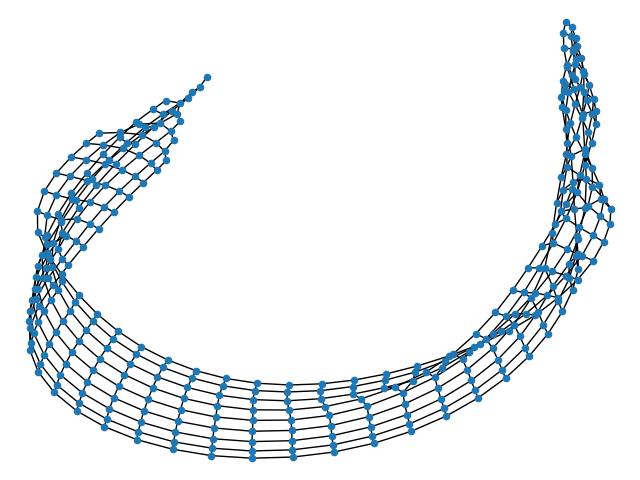} &   \includegraphics[width=30mm]{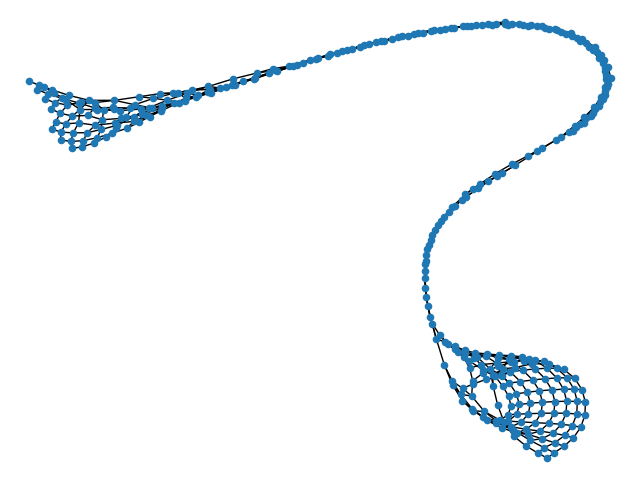}  &   \includegraphics[width=30mm]{Graphs/grid/graphrnn/GRIDRNN-rnngrid_Best_00163.png} &   \includegraphics[width=30mm]{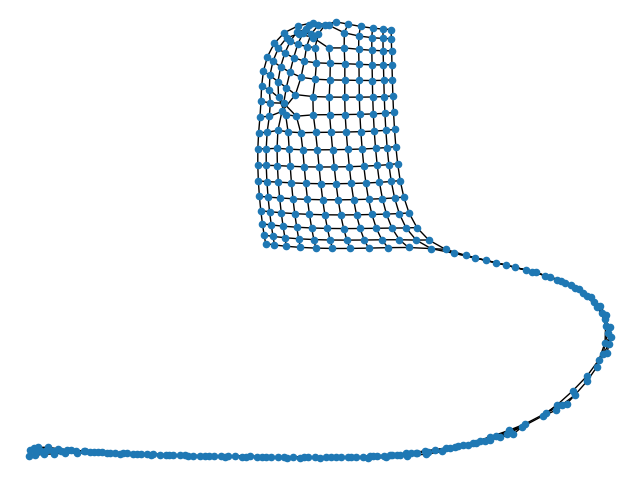}  \\
    \centering
    \begin{tabular}{l}
    \centering \pbox{15cm}{\scriptsize{GraphRNN-S}}
  \end{tabular}  &   \includegraphics[width=30mm]{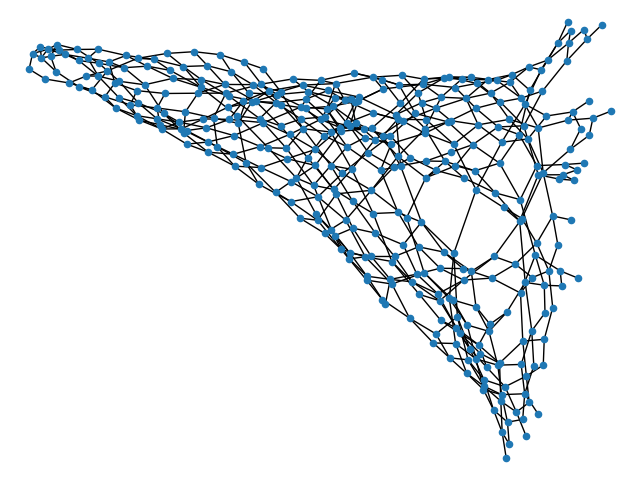} &   \includegraphics[width=30mm]{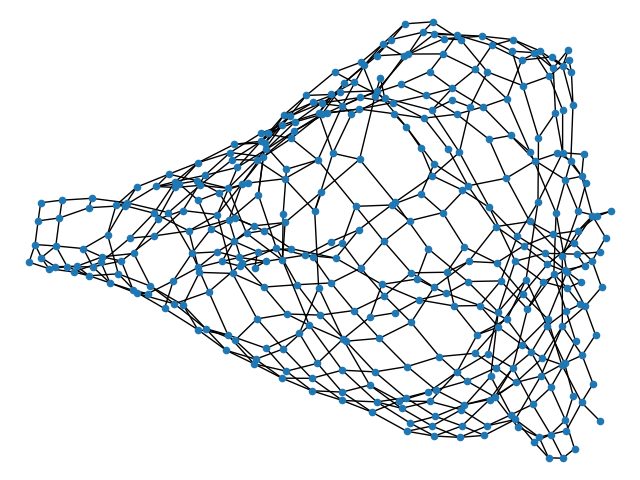}  &   \includegraphics[width=30mm]{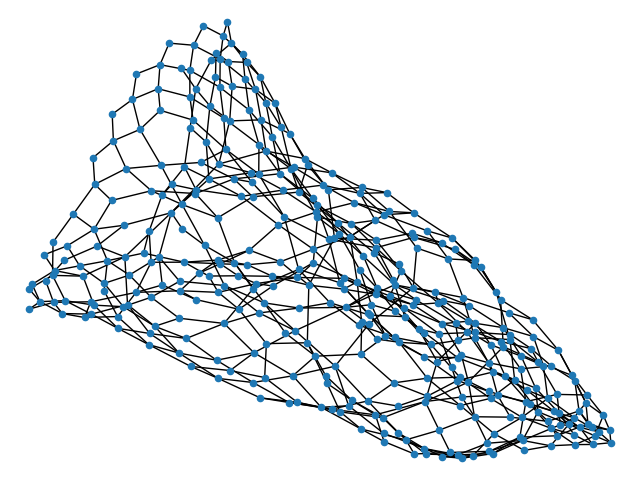} &   \includegraphics[width=30mm]{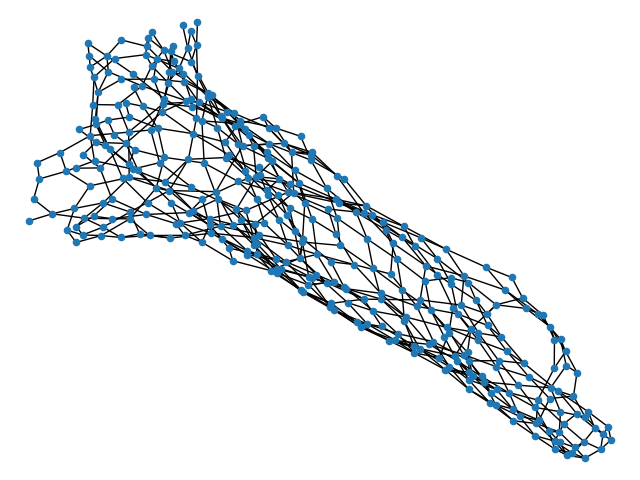}  \\
\end{tabular}
\caption{Visualization of generated \textbf{Grid} graphs by  benchmark GGMs and the micro-macro modeling effect. The first block shows four randomly selected graphs from the test set. The first and second rows in the second block show samples generated by GraphVAE and GraphVAE\ourModelAcronym~models respectively. The bottom block shows graphs generated  by benchmark GGMs. Graphs generated with micro-macro modeling, GraphVAE\ourModelAcronym, match the target graph the best and make a noticeable improvement in comparison to GraphVAE. For each model we visually select and visualize the most similar generated samples to the test set.
}
\label{fig:gridVisualization}
\end{figure}

\begin{figure}
\begin{tabular}{m{0.09\textwidth} m{0.19\textwidth} m{0.19\textwidth} m{0.19\textwidth} m{0.19\textwidth}}
\centering
    \begin{tabular}{l}
    \centering \scriptsize{Test}
  \end{tabular}  &   \includegraphics[width=30mm]{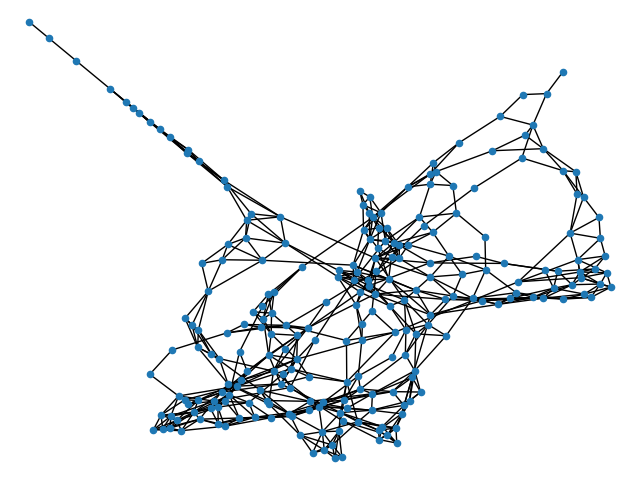} &   \includegraphics[width=30mm]{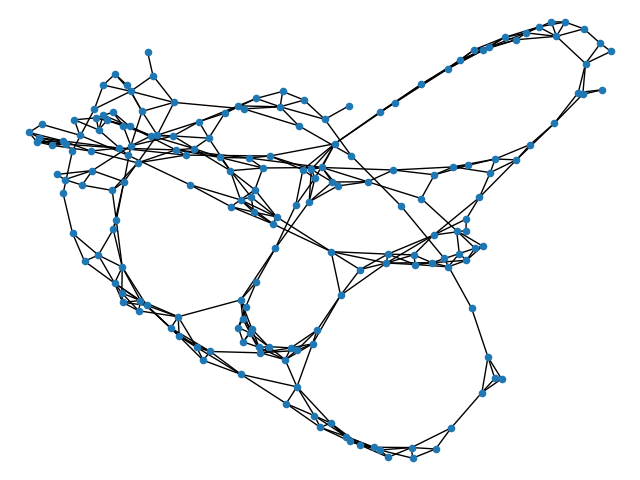}  &   \includegraphics[width=30mm]{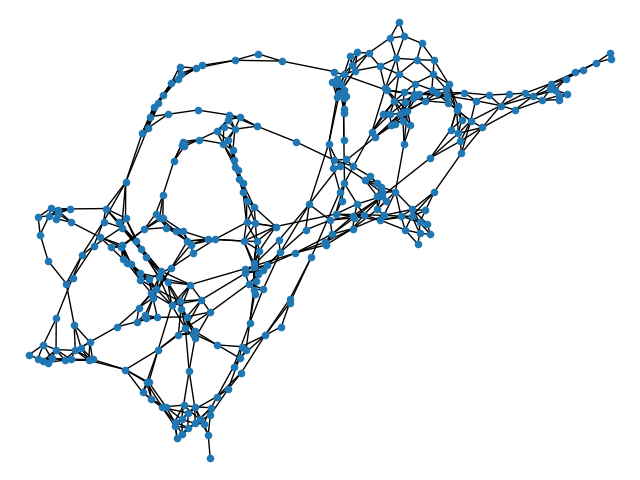} &   \includegraphics[width=30mm]{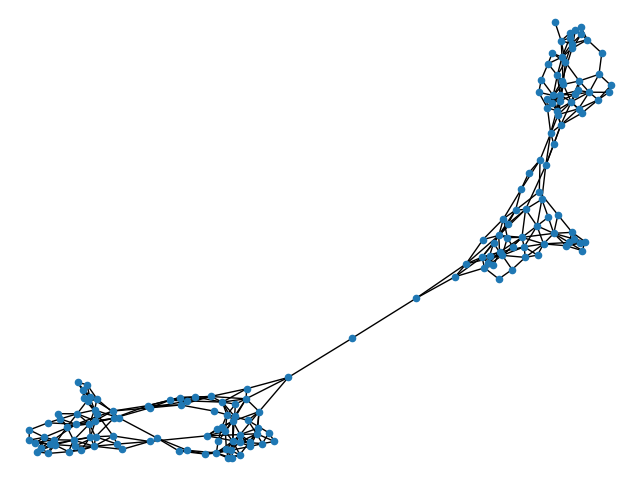}  \\ \hline
\centering
    \begin{tabular}{l}
    \centering \scriptsize{GraphVAE}
  \end{tabular}  &   \includegraphics[width=30mm]{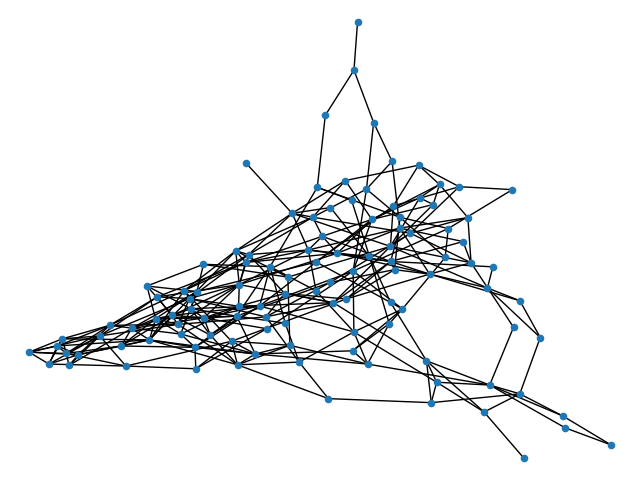} &   \includegraphics[width=30mm]{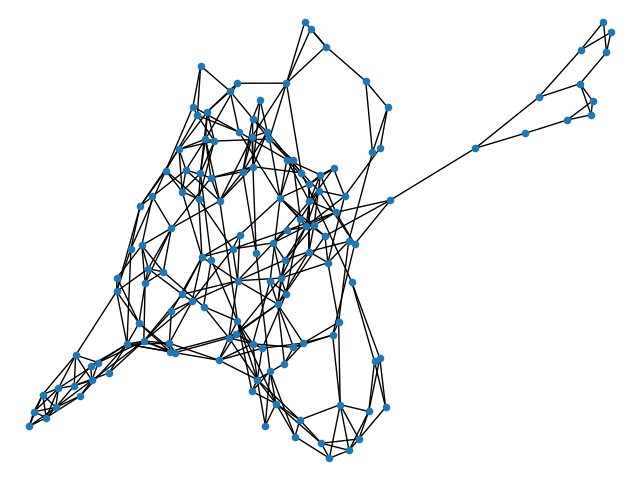}  &   \includegraphics[width=30mm]{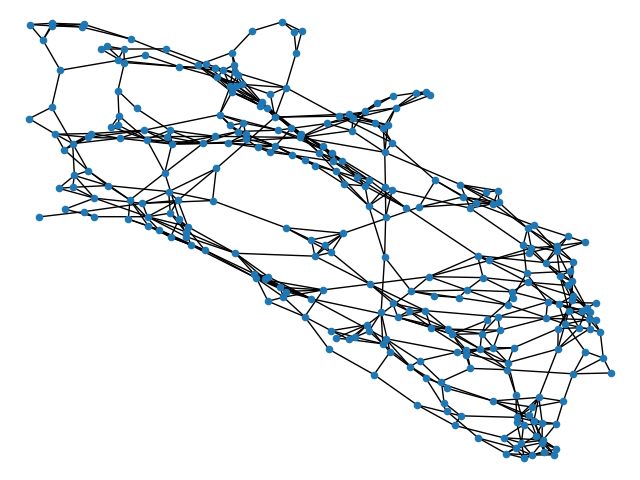} &   \includegraphics[width=30mm]{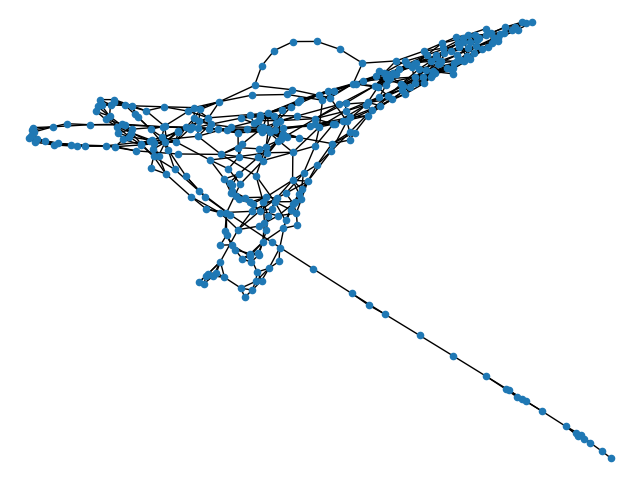}  \\
  \centering
    \begin{tabular}{l}
    \centering \scriptsize{\textbf{\tiny{GraphVAE\ourModelAcronym}}}
  \end{tabular}  &   \includegraphics[width=30mm]{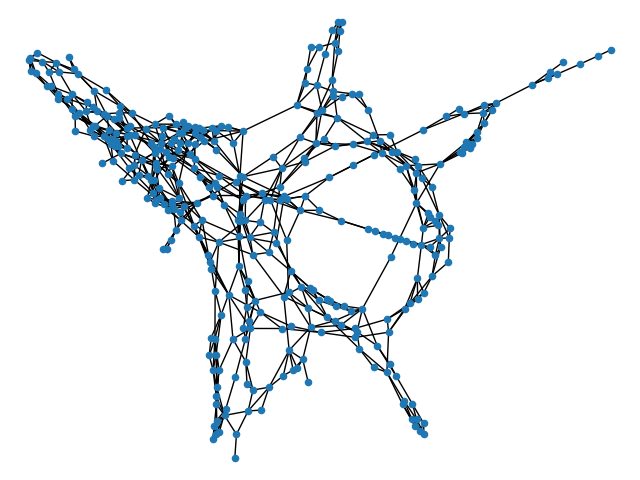} &   \includegraphics[width=30mm]{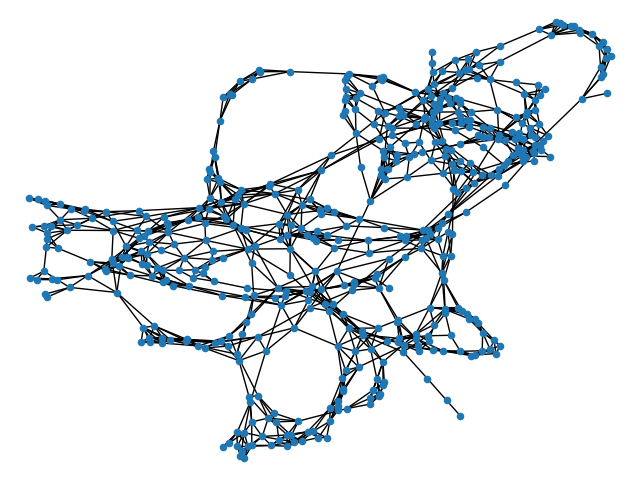}  &   \includegraphics[width=30mm]{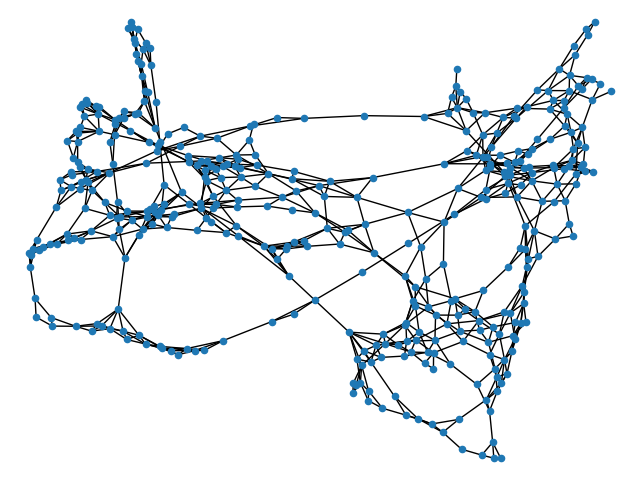} &   \includegraphics[width=30mm]{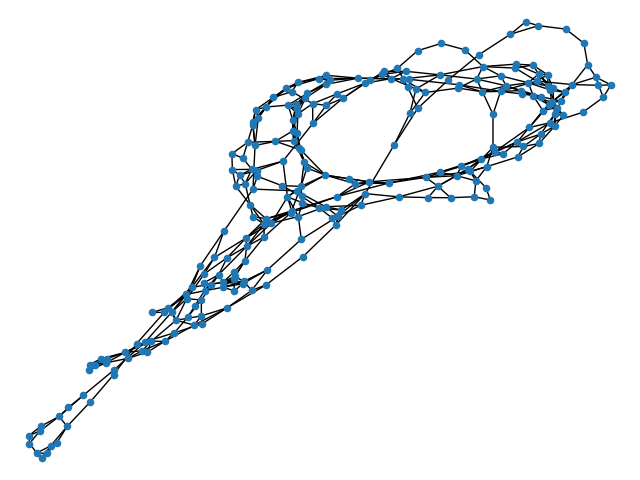}  \\ \hline
    \centering
    \begin{tabular}{l}
    \centering \scriptsize{BIGG}
  \end{tabular}  &   \includegraphics[width=30mm]{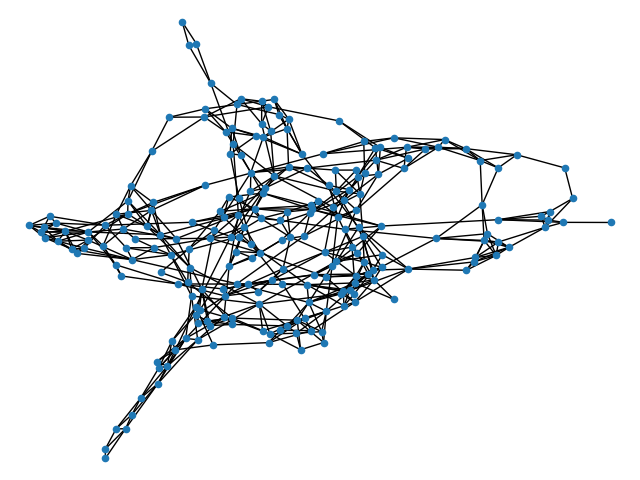} &   \includegraphics[width=30mm]{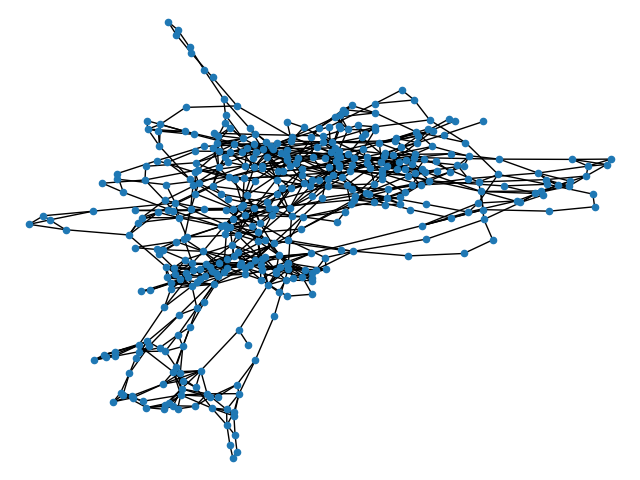}  &   \includegraphics[width=30mm]{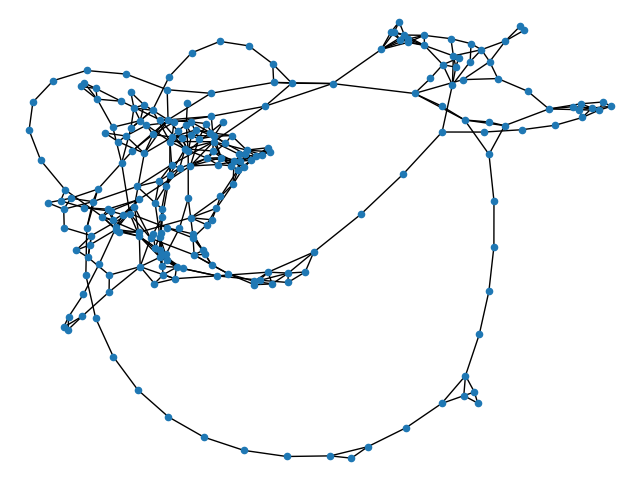} &   \includegraphics[width=30mm]{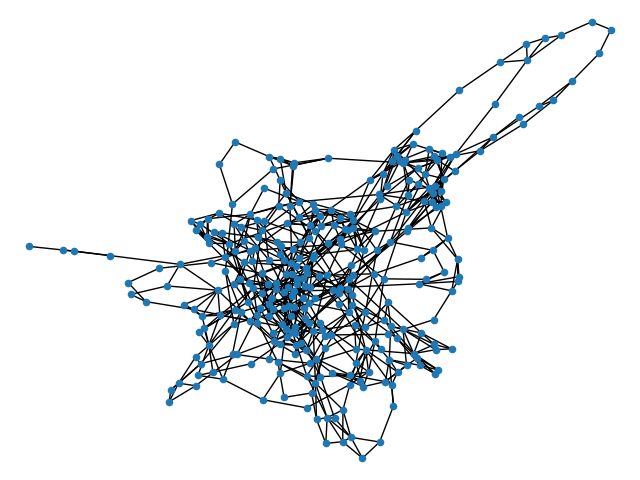}  \\
    \centering
    \begin{tabular}{l}
    \centering \pbox{15cm}{\scriptsize{GRAN}}
  \end{tabular}  &   \includegraphics[width=30mm]{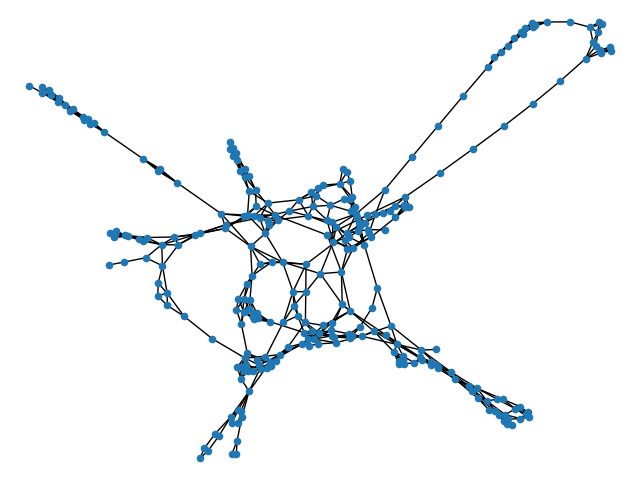} &   \includegraphics[width=30mm]{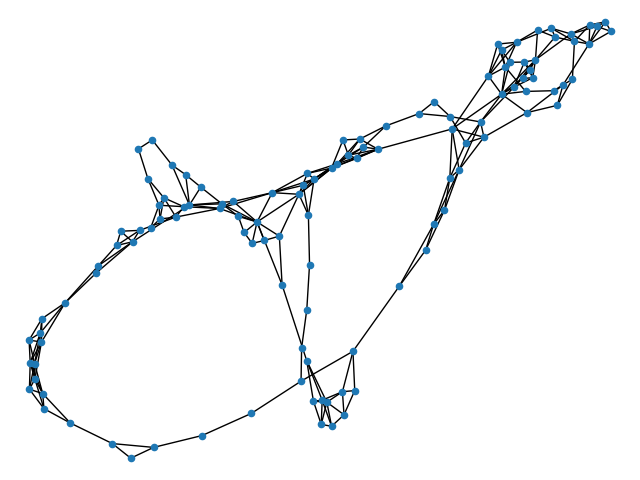}  &   \includegraphics[width=30mm]{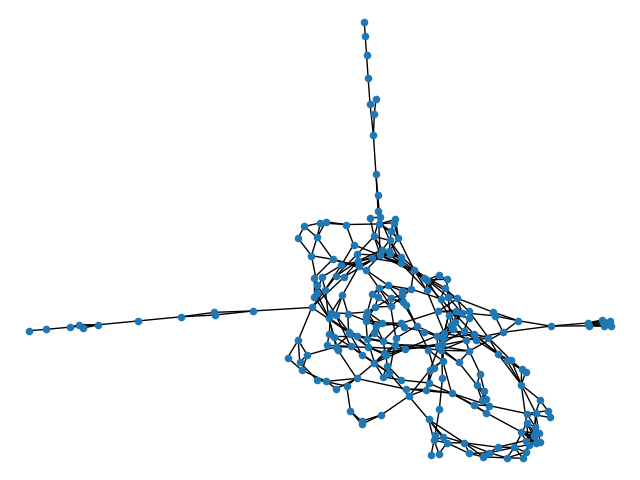} &   \includegraphics[width=30mm]{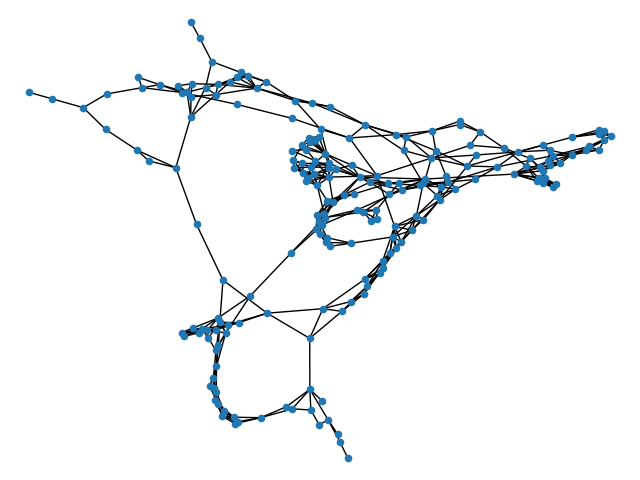}  \\
  \centering
    \begin{tabular}{l}
    \centering \scriptsize{GraphRNN}
  \end{tabular}  &   \includegraphics[width=30mm]{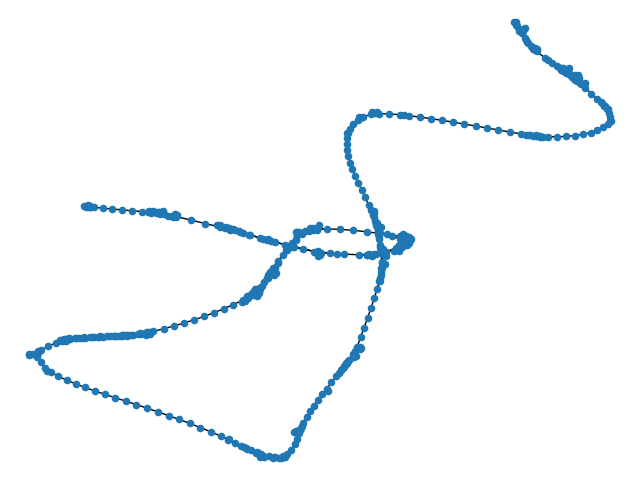} &   \includegraphics[width=30mm]{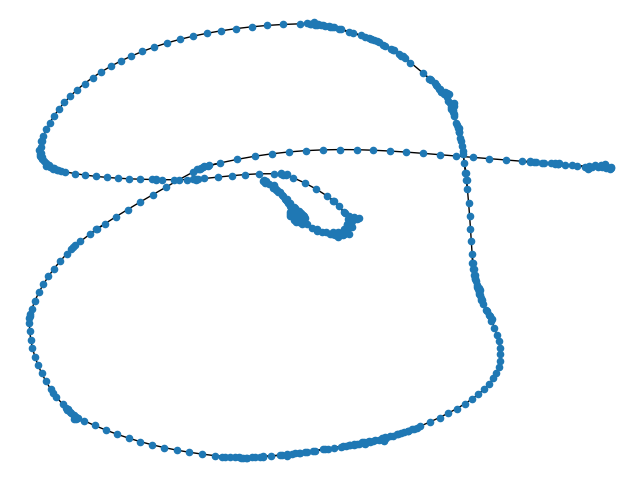}  &   \includegraphics[width=30mm]{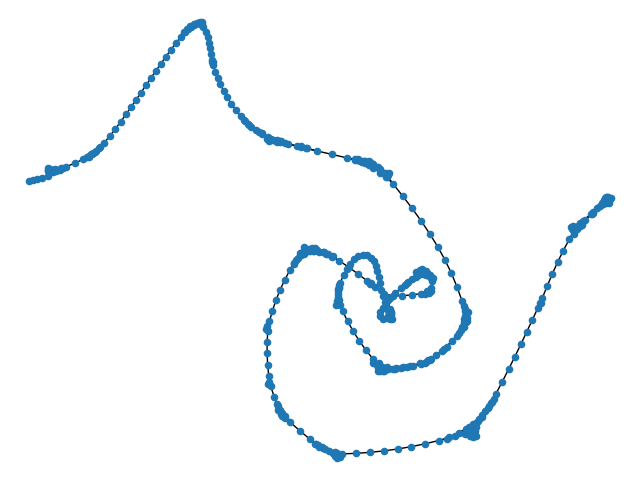} &   \includegraphics[width=30mm]{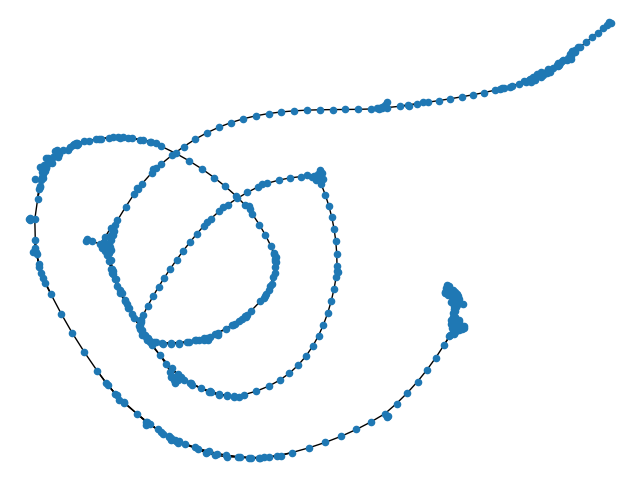}  \\
    \centering
    \begin{tabular}{l}
    \centering \pbox{15cm}{\scriptsize{GraphRNN-S}}
  \end{tabular}  &   \includegraphics[width=30mm]{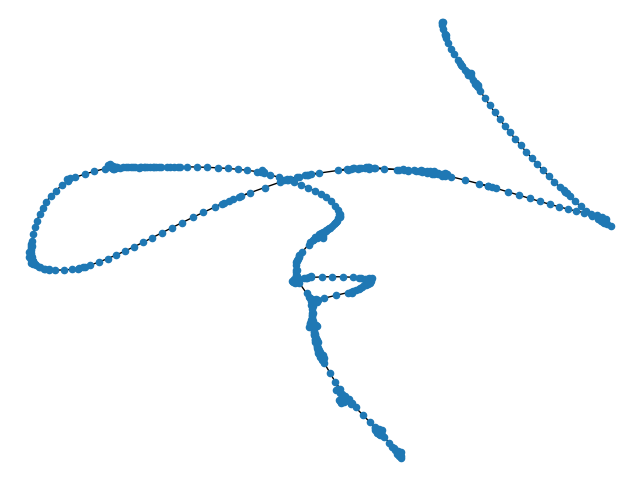} &   \includegraphics[width=30mm]{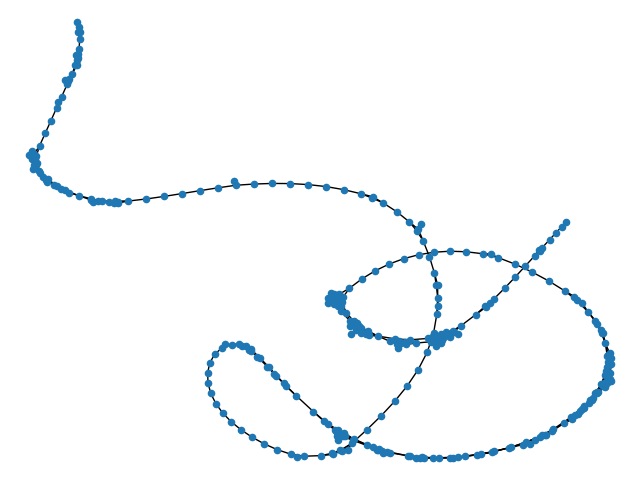}  &   \includegraphics[width=30mm]{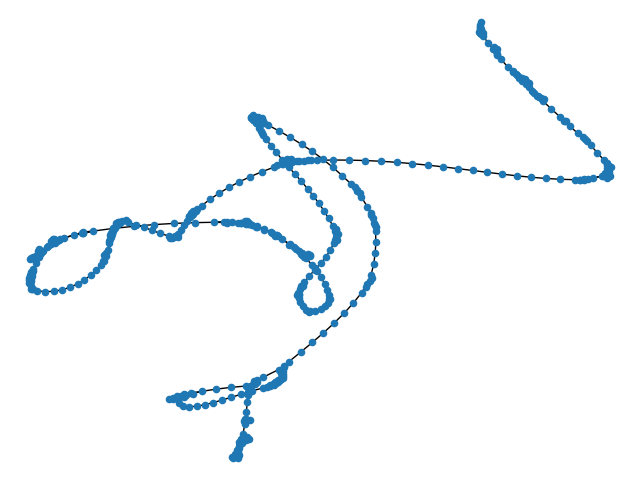} &   \includegraphics[width=30mm]{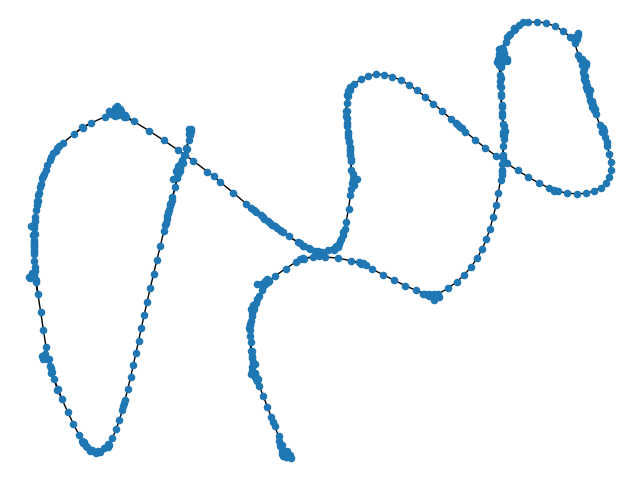}  \\
\end{tabular}
\caption{Visualization of generated \textbf{Protein} graphs by  benchmark GGMs and the micro-macro modeling effect. The first block shows four randomly selected graphs from the test set. The first and second rows in the second block show samples generated by GraphVAE and GraphVAE\ourModelAcronym~models respectively. The bottom block shows graphs generated  by benchmark GGMs. Graphs generated with micro-macro modeling, GraphVAE\ourModelAcronym, make a noticeable improvement in comparison to GraphVAE. For each model we visually select and visualize the most similar generated samples to the test set.
}
\label{fig:DDVisualization2}
\end{figure}

\begin{figure}
\begin{tabular}{m{0.09\textwidth} m{0.19\textwidth} m{0.19\textwidth} m{0.19\textwidth} m{0.19\textwidth}}
\centering
    \begin{tabular}{l}
    \centering \scriptsize{Test}
  \end{tabular}  &   \includegraphics[width=30mm]{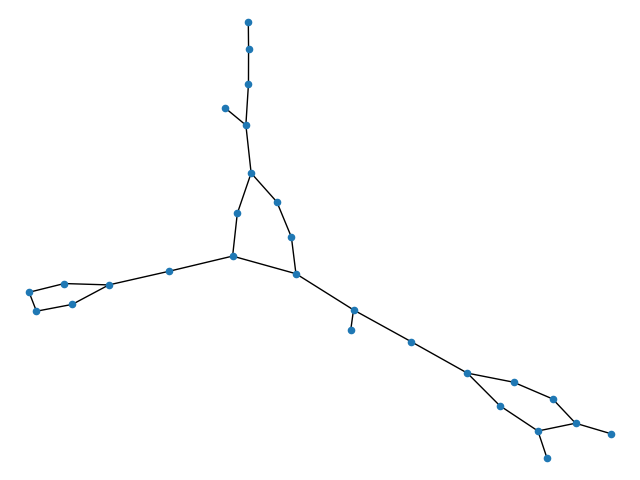} &   \includegraphics[width=30mm]{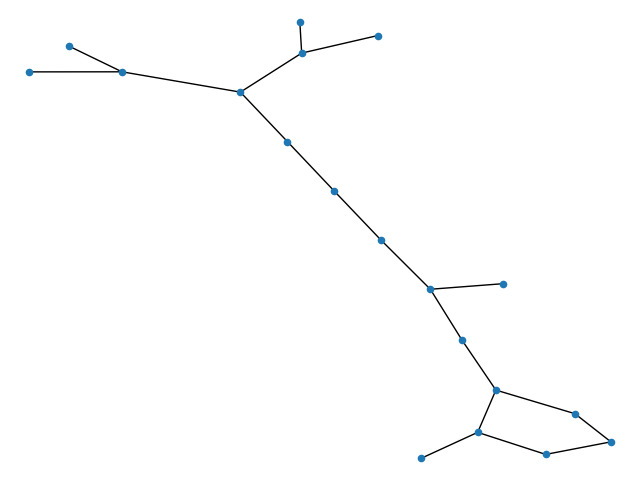}  &   \includegraphics[width=30mm]{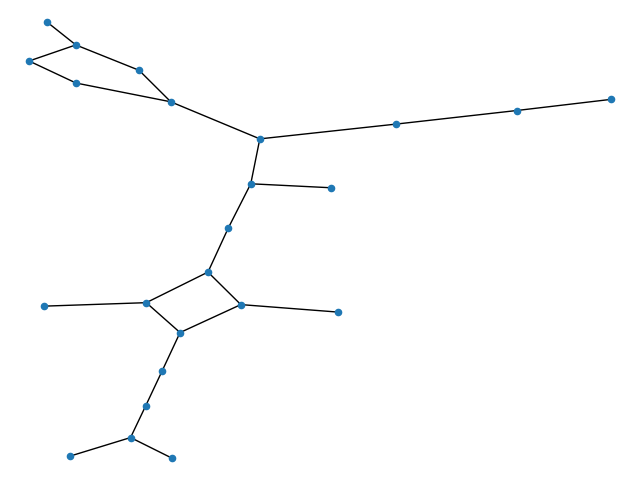} &   \includegraphics[width=30mm]{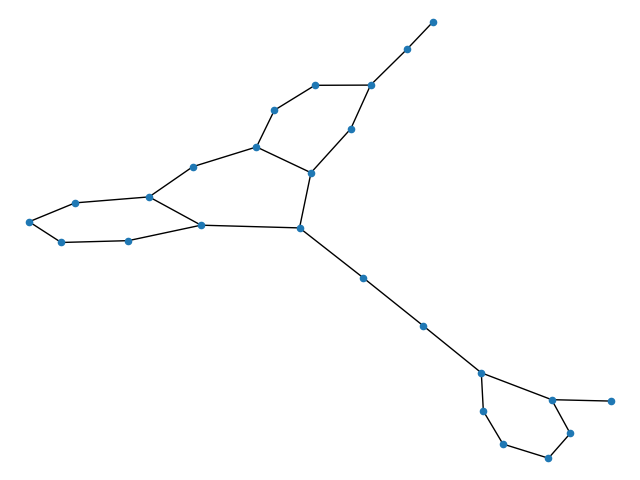}  \\ \hline
\centering
    \begin{tabular}{l}
    \centering \scriptsize{GraphVAE}
  \end{tabular}  &   \includegraphics[width=30mm]{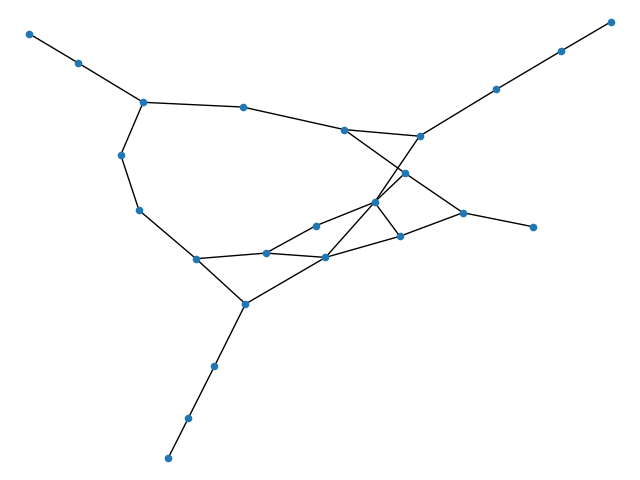} &   \includegraphics[width=30mm]{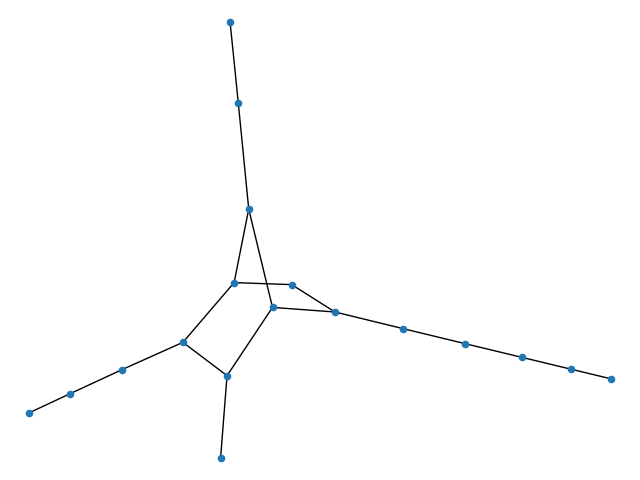}  &   \includegraphics[width=30mm]{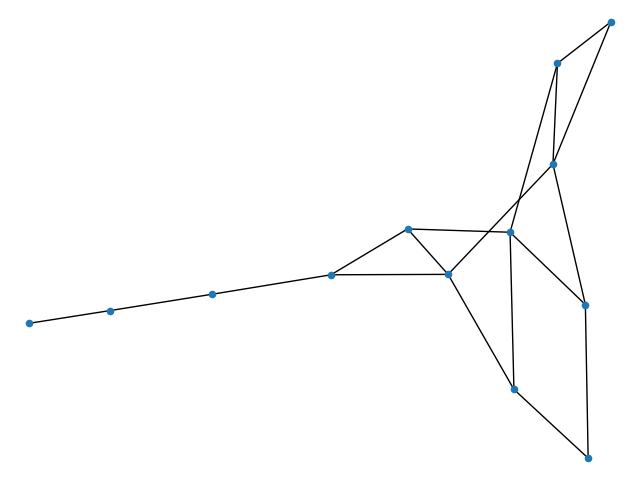} &   \includegraphics[width=30mm]{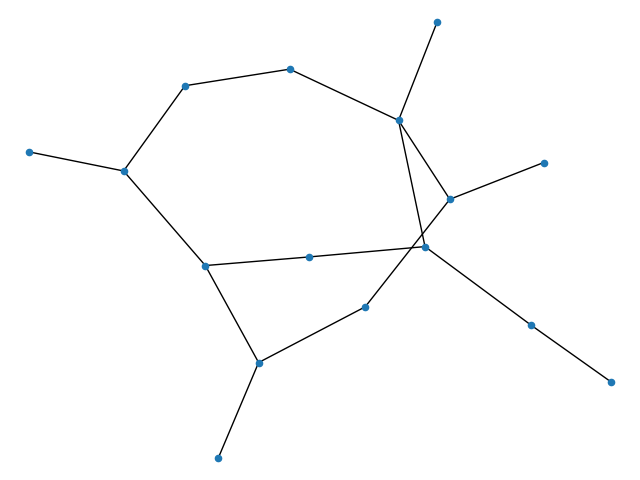}  \\
  \centering
    \begin{tabular}{l}
    \centering \scriptsize{\textbf{\tiny{GraphVAE\ourModelAcronym}}}
  \end{tabular}  &   \includegraphics[width=30mm]{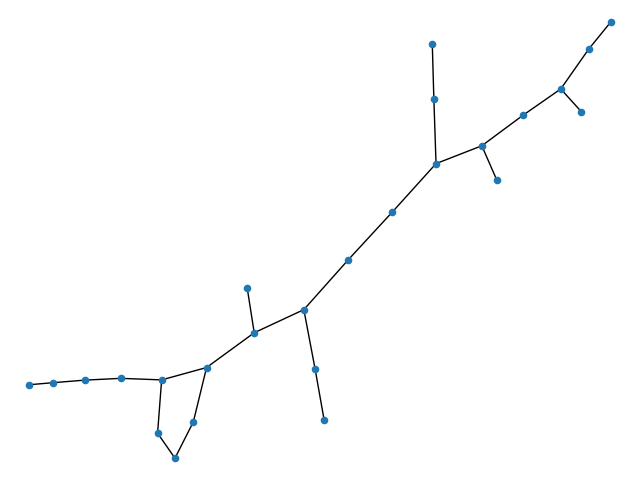} &   \includegraphics[width=30mm]{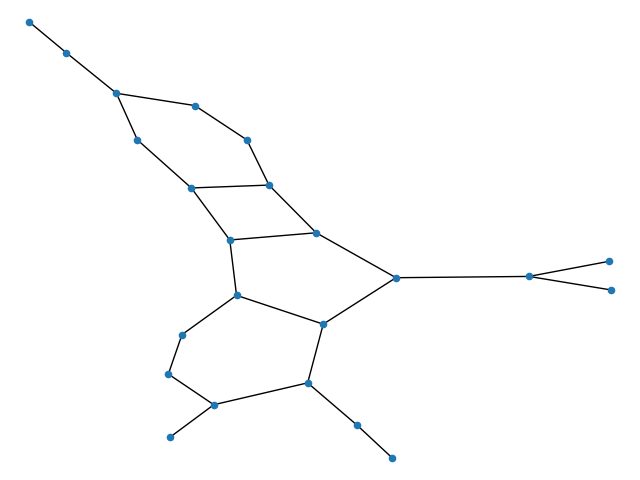}  &   \includegraphics[width=30mm]{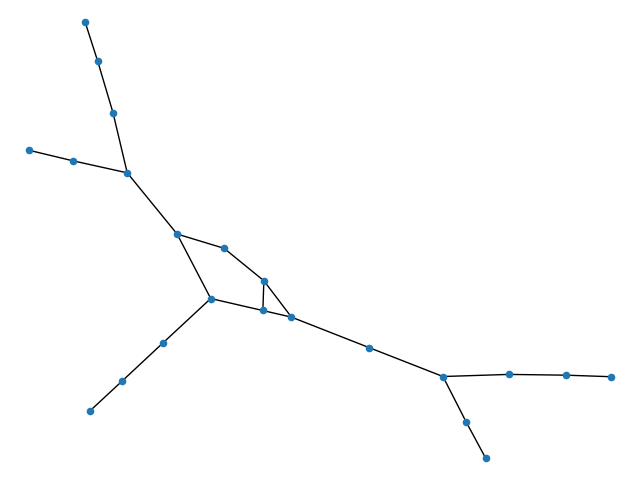} &   \includegraphics[width=30mm]{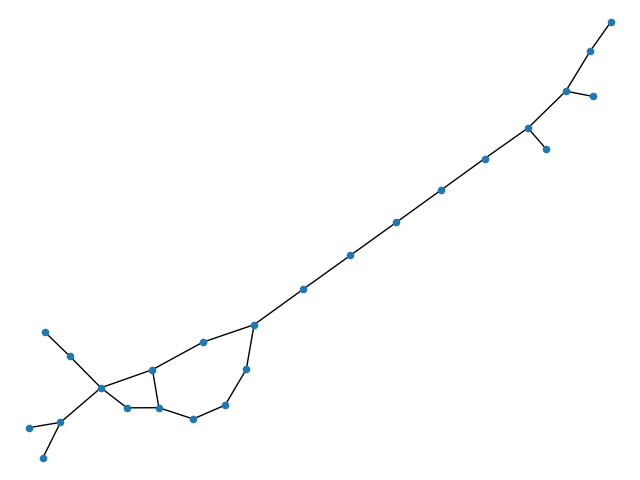}  \\ \hline
    \centering
    \begin{tabular}{l}
    \centering \scriptsize{BIGG}
  \end{tabular}  &   \includegraphics[width=30mm]{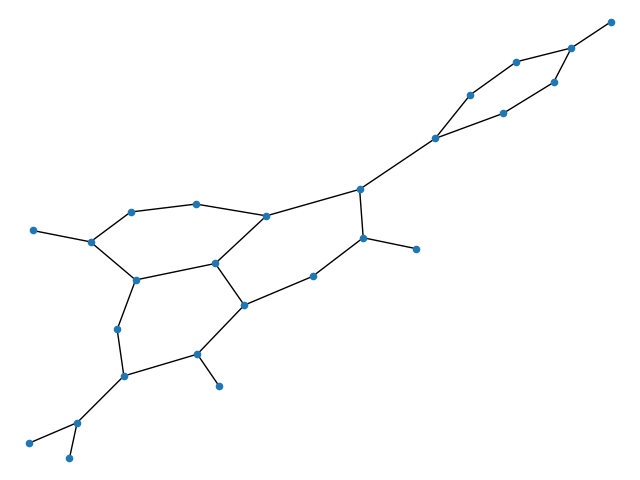} &   \includegraphics[width=30mm]{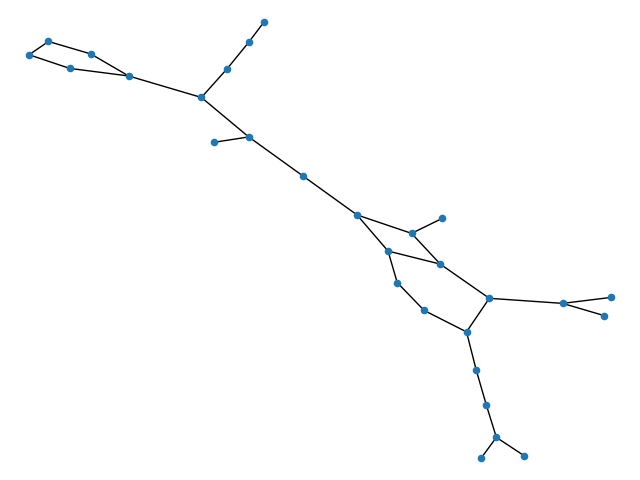}  &   \includegraphics[width=30mm]{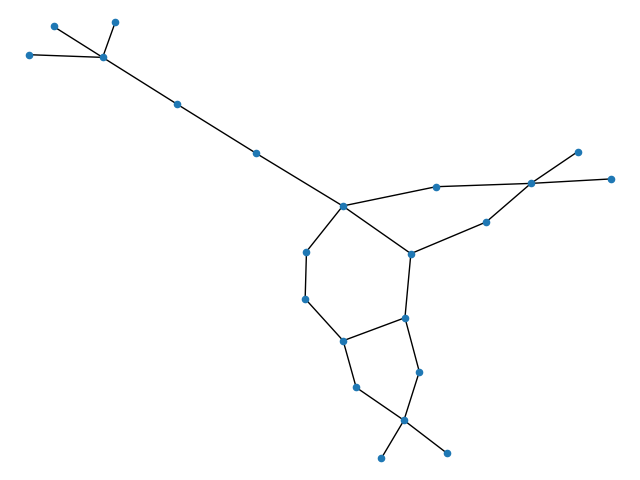} &   \includegraphics[width=30mm]{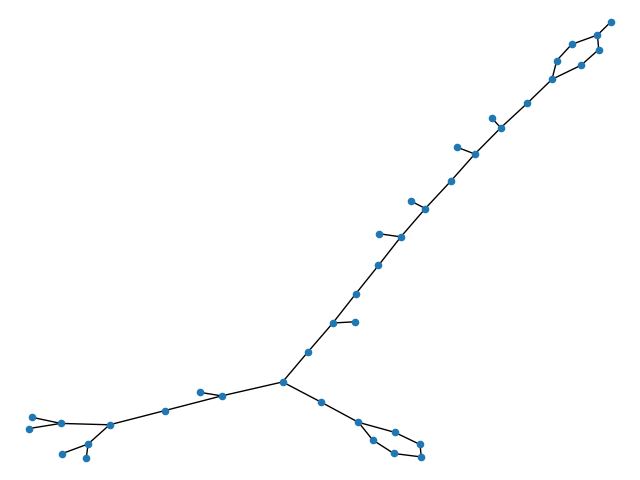}  \\
    \centering
    \begin{tabular}{l}
    \centering \pbox{15cm}{\scriptsize{GRAN}}
  \end{tabular}  &   \includegraphics[width=30mm]{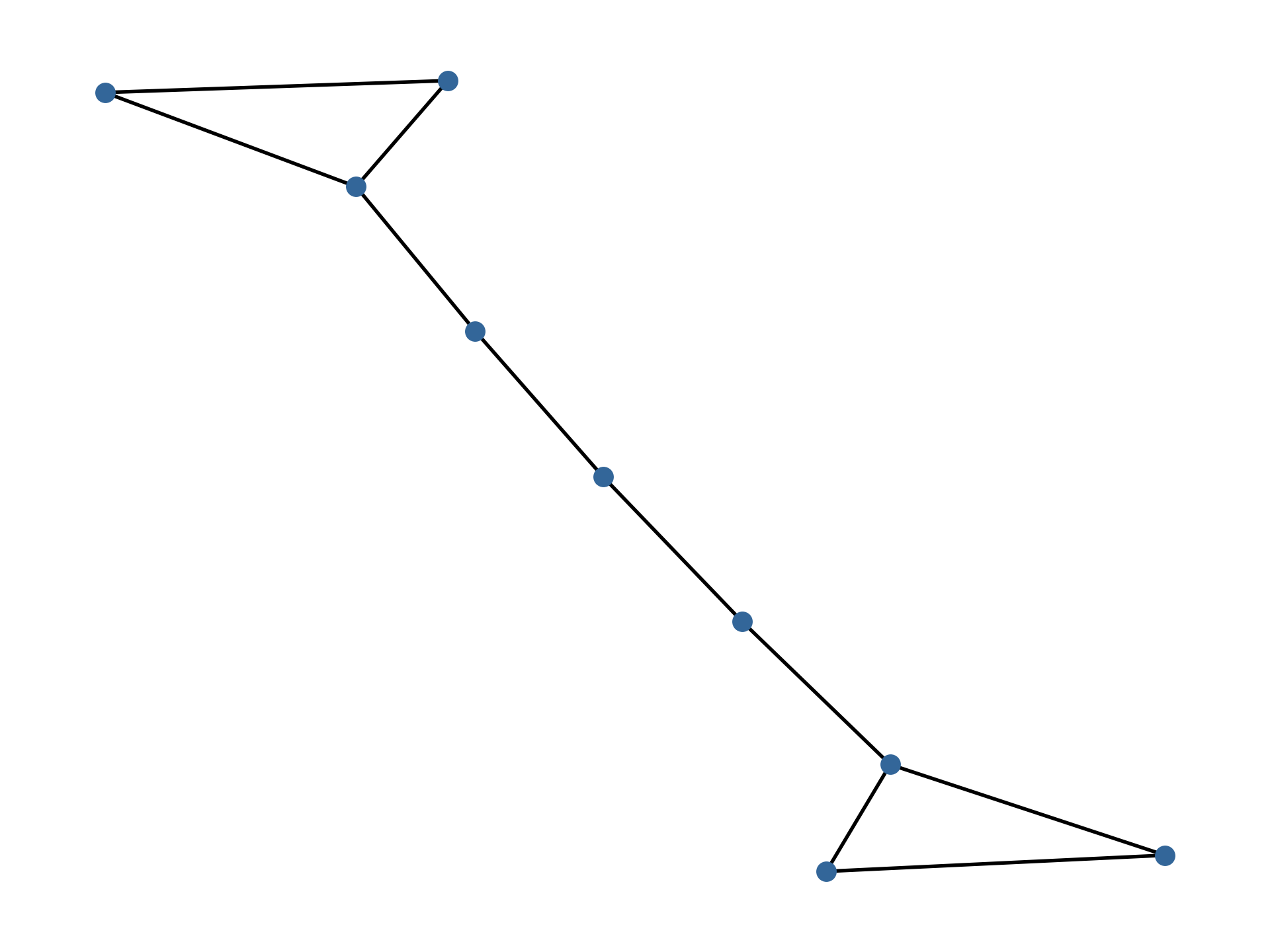} &   \includegraphics[width=30mm]{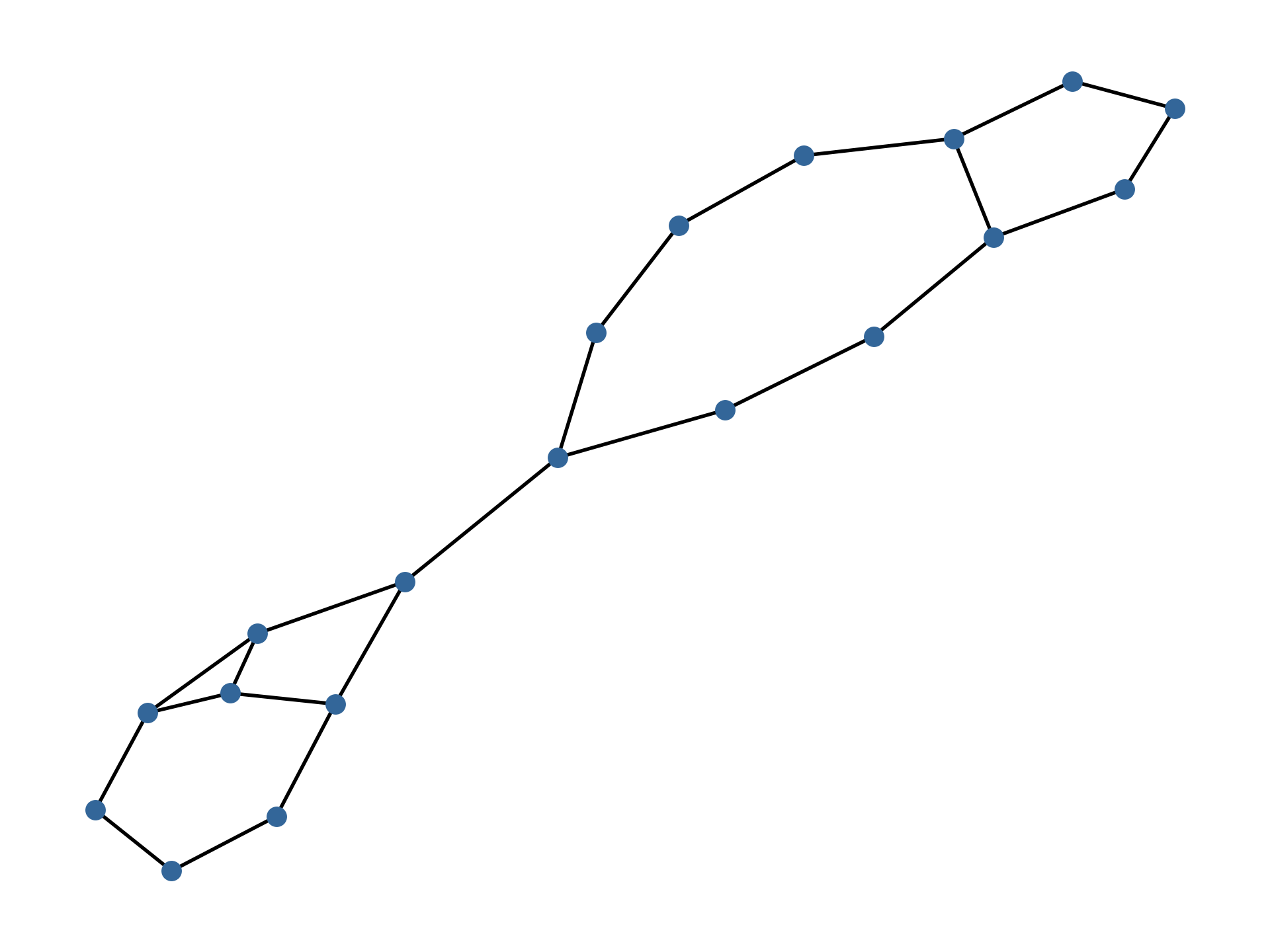}  &   \includegraphics[width=30mm]{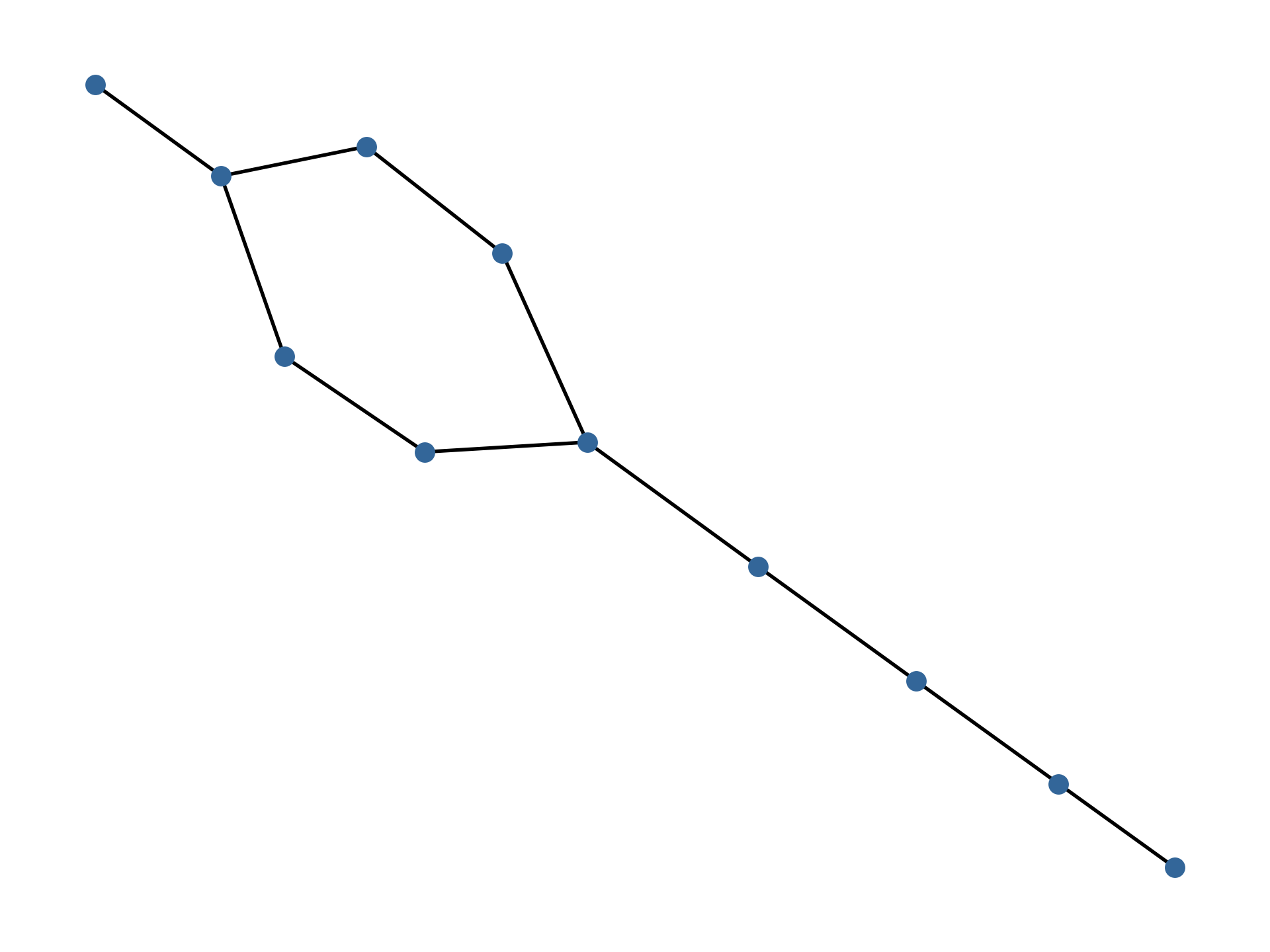} &   \includegraphics[width=30mm]{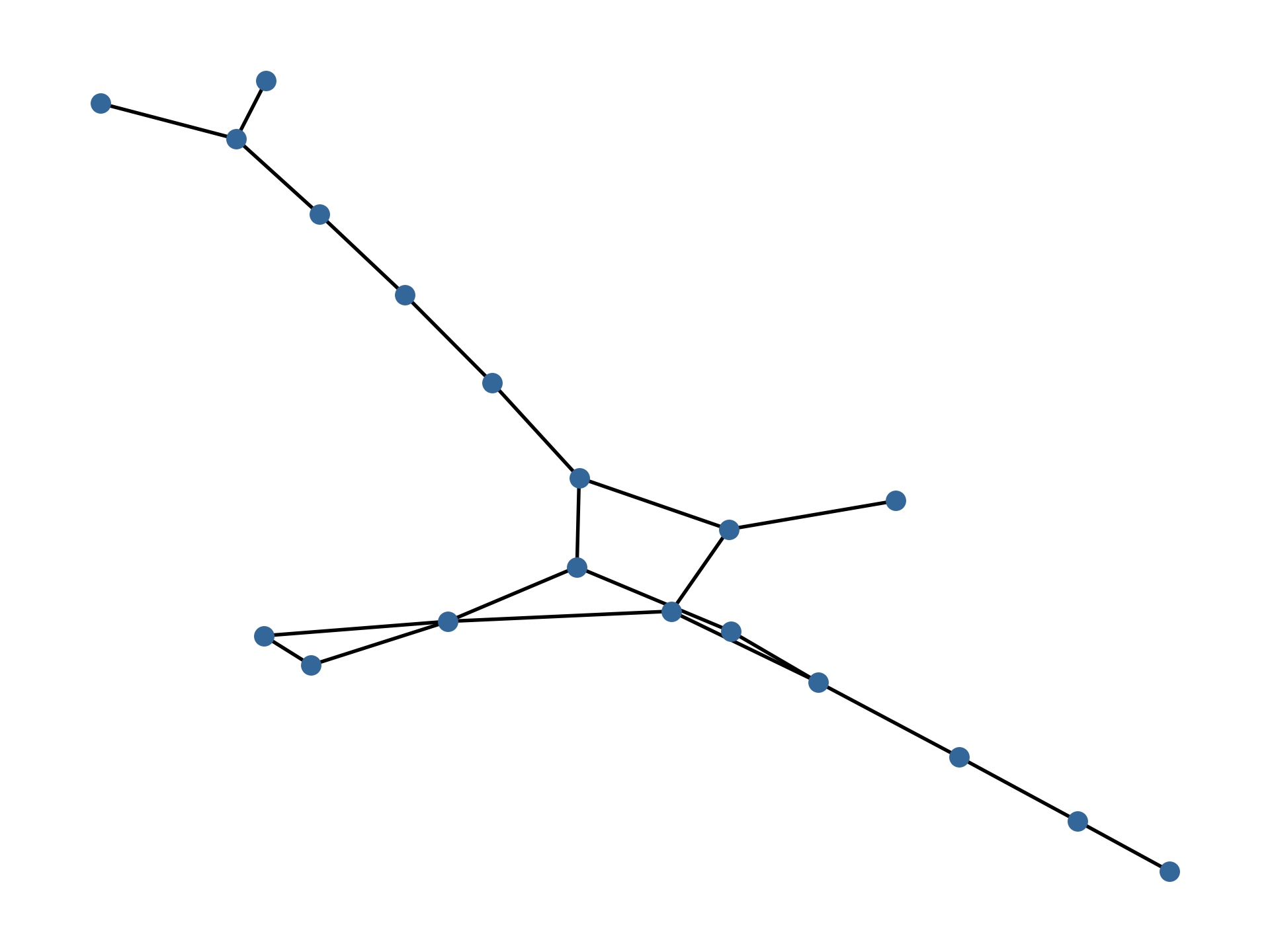}  \\
  \centering
    \begin{tabular}{l}
    \centering \scriptsize{GraphRNN}
  \end{tabular}  &   \includegraphics[width=30mm]{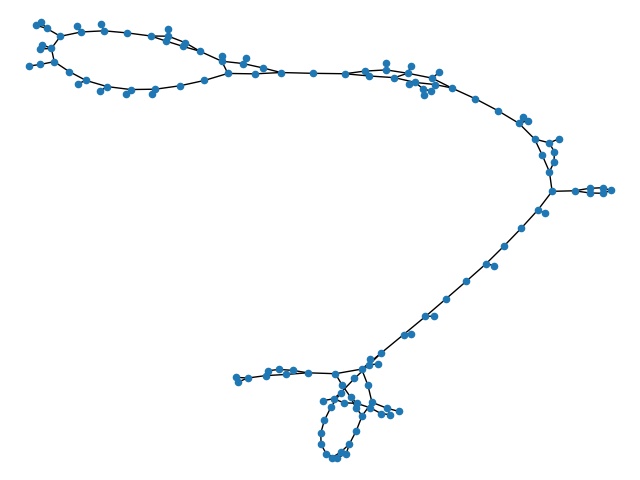} &   \includegraphics[width=30mm]{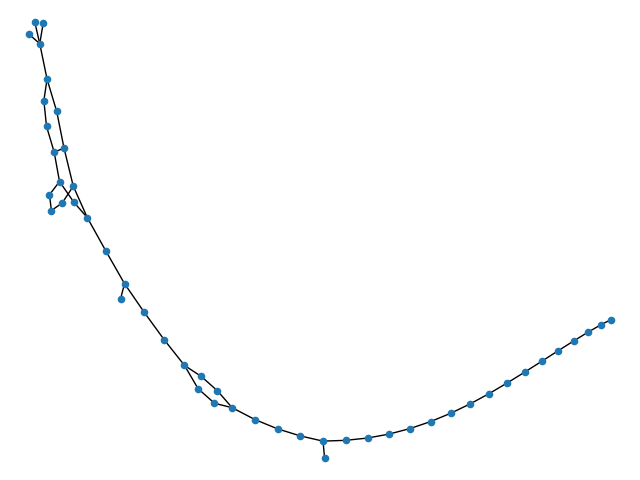}  &   \includegraphics[width=30mm]{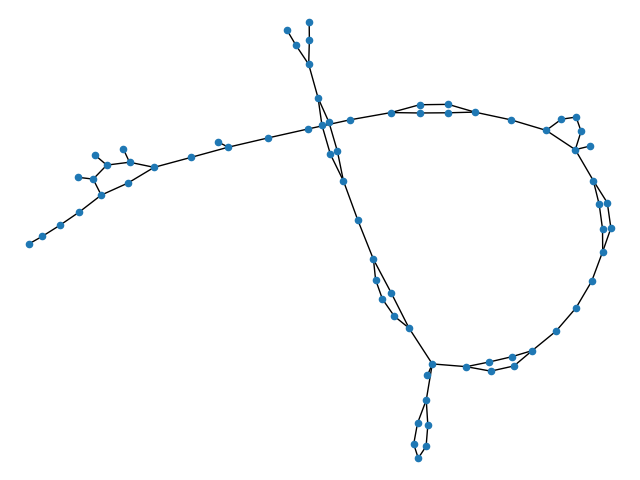} &   \includegraphics[width=30mm]{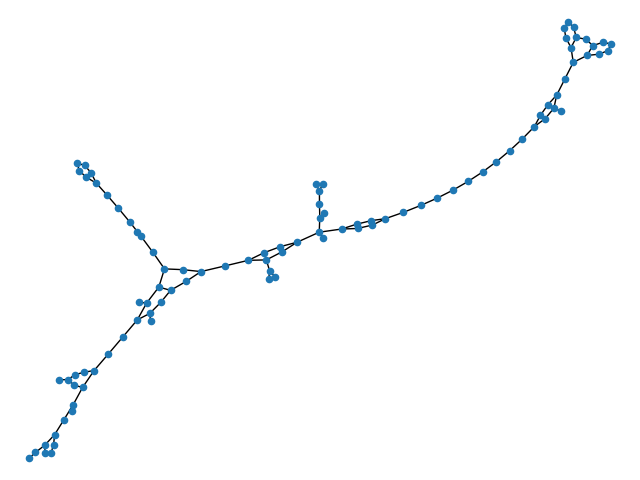}  \\
    \centering
    \begin{tabular}{l}
    \centering \pbox{15cm}{\scriptsize{GraphRNN-S}}
  \end{tabular}  &   \includegraphics[width=30mm]{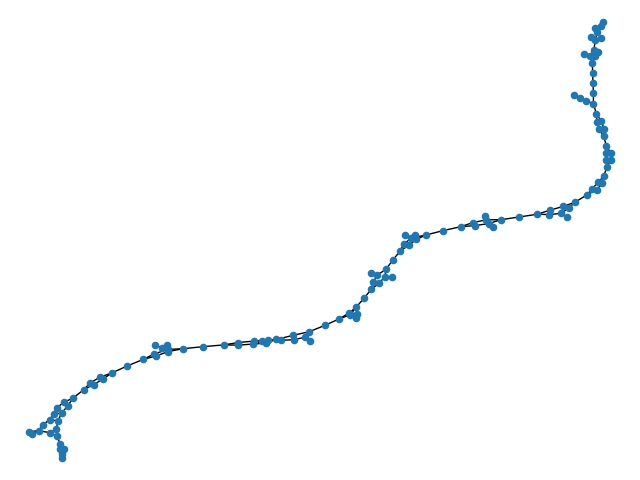} &   \includegraphics[width=30mm]{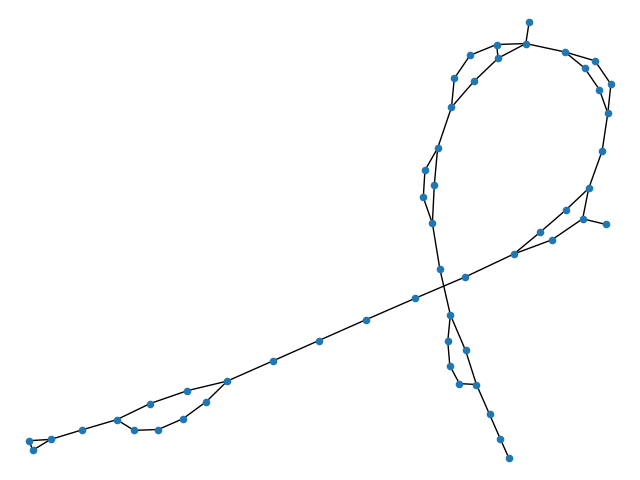}  &   \includegraphics[width=30mm]{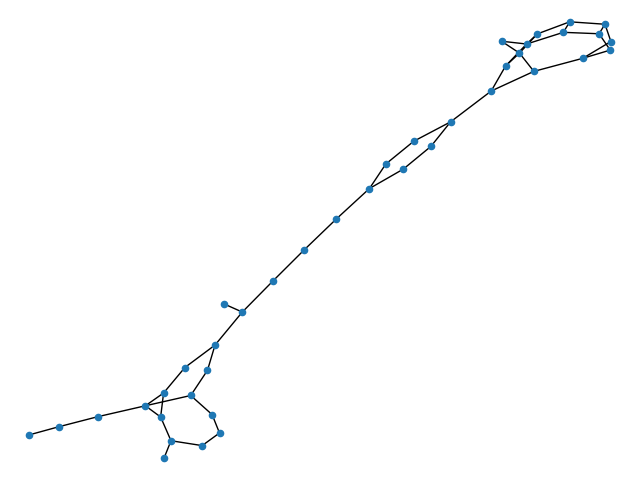} &   \includegraphics[width=30mm]{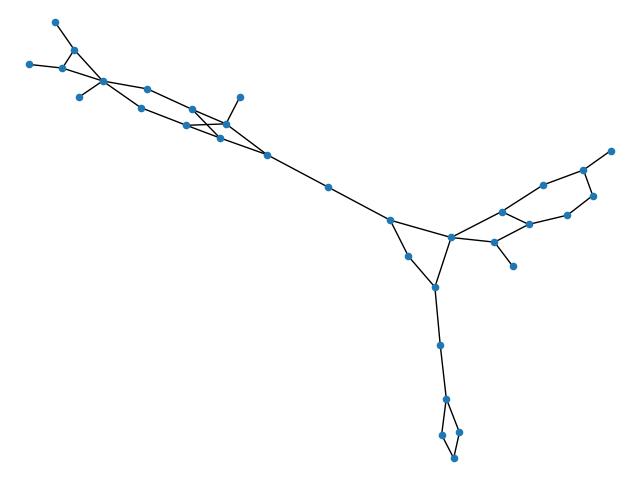}  \\
\end{tabular}
\caption{Visualization of generated \textbf{ogbg-molbbbp} graphs by  benchmark GGMs and the micro-macro modeling effect. The first block shows four randomly selected graphs from the test set. The first and second rows in the second block show samples generated by GraphVAE and GraphVAE\ourModelAcronym~models respectively. The bottom block shows graphs generated  by benchmark GGMs. Graphs generated with micro-macro modeling, GraphVAE\ourModelAcronym, make a noticeable improvement in comparison to GraphVAE. For each model we visually select and visualize the most similar generated samples to the test set.
}
\label{fig:OGBVisualization2}
\end{figure}

Here we extend the visual  examination of the micro-macro modeling  and
benchmark GGMs from Figure~\ref{fig:qualitative_comparision_grid}. The AB methodology for evaluating the micro-macro modeling  is to use the same VGAE architecture, here GraphVAE section \ref{GraphVAE-architecture},  and applying micro-macro modeling. 
Figures \ref{fig:gridVisualization}, \ref{fig:lobster-Visualization2}, \ref{fig:TriGridVisualization2}, \ref{fig:OGBVisualization2} and \ref{fig:DDVisualization2} provide visual comparisons of benchmark GGMs and the effect of the micro-macro modeling approach. For each model, we generate 20 samples and visually select and plot the most similar ones to the test set. The first block in each of the figures shows randomly selected target graphs from the test set. The second block compares the effect of the micro-macro modeling on the GraphVAE and the last block samples  graphs generated by the benchmark GGMs.


\subsection{Lesion studies on MM objective components}
This section drills downs into the different components of the MM objective and the importance of KL-divergence penalty.
The KL-divergence importance is studied by training the GraphVAE with $\beta$-VAE  and different values for hyperparameter $\beta$ which controls the relative importance of the KL-divergence penalty. The result shows that while the KL-divergence weighting affects the model performance, it cannot replace macro modeling. See the first block of table~\ref{table:lesion}.
The second block of table~\ref{table:lesion} investigates the effects of each target statistic in isolation. As shown, no single statistic has the power of all three combined. We also observe that different statistics are more important for different datasets.
 \begin{table}
  \caption{Sensitivity Analysis and Lesion Studies on MM objective components. The first block shows the GraphVAE trained with $\beta$-VAE \cite{higgins2016beta} and different $\beta$. $\beta$-VAE has a limited effect and cannot compensate for the lack of macro components. The second block investigates the effects of each target statistic in isolation. No single statistic has the power of all three combined.}
  \label{table:lesion}
  \centering
  \resizebox{\textwidth}{!}{
  \begin{tabular}{lcccccccccc}
    \toprule
    \multirow{2}{3.5em}{\textbf{Method}} &  \multicolumn{5}{c}{\textbf{Lobster}} &  \multicolumn{5}{c}{\textbf{ogbg-molbbbp}}\\
    & \small{Deg.} & \small{Clus.} & \small{Orbit}& \small{Spect} & \small{Diam.}& \small{Deg.} & \small{Clus.} & \small{Orbit}& \small{Spect} & \small{Diam.} \\
   \midrule
   GraphVAE ($\beta=1$) & $0.081$ & $0.739$  & $0.372$ & $0.056$ & $0.129$  & $0.028$ & $0.442$ & $0.047$ & $0.015$& $0.055$\\

   GraphVAE ($\beta=2$) & $0.039$ & $0.324$  & $0.262$ & $0.031$ & $0.021$ & $0.022$ & $0.425$ & $0.032$ & $0.015$ & $0.028$\\
   GraphVAE ($\beta=4$) & $0.068$ & $0.485$  & $0.472$ & $0.065$ & $0.147$ & $0.044$ &  $0.560$ & $0.063$ & $0.019$ & $0.046$\\
   GraphVAE ($\beta=8$) & $0.080$ & $0.432$  & $0.465$ & $0.065$ & $0.085$ & $0.127$ & $0.736$ & $0.250$ & $0.034$ & $0.055$\\
   GraphVAE ($\beta=16$) & $0.024$ & $0.314$  & $0.295$ & $0.024$ & $\textbf{0.068}$ & $0.243$ & $0.834$ & $0.537$ & $0.051$ & $0.109$\\
   GraphVAE ($\beta=32$) & $0.099$ & $0.513$  & $0.594$ & $0.045$ & $0.148$ & $0.392$ & $0.975$ & $0.896$ & $0.103$ & $0.266$\\
   \midrule
   GraphVAE-\ourModelAcronym & $\textbf{2}e^\textbf{{-4}}$ & $\textbf{0}$ & ${\textbf{0.008}}$ & ${0.017}$ & ${0.187}$ & $\textbf{0.001}$ & $\textbf{0.005}$ & $\textbf{8e}^\textbf{-4}$ & $\textbf{0.005}$&$\textbf{0.018}$ \\
   GraphVAE-Triangle-Count & $0.009$ & $\textbf{0}$  & $0.052$ & $\textbf{0.016}$ & $0.209$ & $0.382$ & $0.962$ & $0.846$& $0.076$ & $0.116$\\
   GraphVAE-$s$-step & $0.015$ & $0.255$  & $0.092$ & $0.025$ & $0.097$ & $0.399$ & $0.943$ & $0.866$ &$0.097$ & $0.336$\\
   GraphVAE-Degree-Hist & $0.025$ & $0.336$  & $0.342$ & $0.228$ & $0.078$ & $0.349$ & $0.950$ & $0.811$ & $0.080$ & $0.128$\\
    \bottomrule
  \end{tabular}%
 }
\end{table} 
 \subsection{Variance parameter}
Table \ref{table:variances} shows the values of the  learned variance parameter $\kweight_{\kthFeature}^{2}$  by the optimal $\sigma$-VAE approach for each of the datasets. The variance of a graph statistic can be interpreted as quantifying the empirical uncertainty of a graph statistic. Taking for example the {\em Triangle count} statistic, the learned variance   for the Lobster, Grid, and ogbg-molbbbp datasets is very small because there are almost no triangles in these datasets. For the protein dataset, the learned variance is comparatively large, which indicates the number of triangles has a wider range of values in this dataset.
\begin{table}[h]
 \caption{Values of the  learned variance parameter,  $\kweight_{\kthFeature}^{2}$, for each graph statistic. The variance values can be interpreted as quantifying the empirical uncertainty of a graph statistic.}
    \centering
	\label{table:variances}
	\resizebox{14cm}{!}{
    \begin{tabular}{lccccccccc}
    \hline
        {Dataset} & $2$-step & $3$-Step  & $4$-step   & $5$-step   & $6$-step  &  Degree hist.  & Triangle count \\ 
        \toprule
        {Triangle Grid} & $3.93e^{-5}$ & $4.11e^{-5}$ & $4.39e^{-5}$ & $4.53e^{-5}$ & $4.63e^{-5}$ & $0.21$ &  $4.94$ \\ 
        {Lobster} & $1.18e^{-5}$ & $9.85e^{-6}$ & $9.61e^{-6}$ & $9.49e^{-6}$ & $9.45e^{-6}$ & $3.89e^{-5}$  & $6.87e^{-6}$ \\ 
        {Grid} & $2.90e^{-5}$ & $2.16e^{-5}$ & $2.02e^{-5}$ & $1.92e^{-5}$ & $1.86e^{-5}$ & $0.02$ &  $6.85e^{-6}$ \\ 
        {Protein} & $2.09e^{-5}$ & $1.47e^{-5}$ & $1.41e^{-5}$ & $1.39e^{-5}$ & $1.38e^{-5}$ & $0.13$  & $158.82$ \\ 
        {ogbg-molbbbp} & $3.94e^{-5}$ & $2.82e^{-5}$ & $2.74e^{-5}$ & $2.61e^{-5}$ & $2.54e^{-5}$ & $7.84e^{-4}$  & $1.40e^{-4}$ \\
            \bottomrule
    \end{tabular}}
\end{table}
\subsection{Statistics-based comparison with benchmark GGMs}

Table~\ref{table:statMertics-SOTA} shows the benchmark results of statistics-based evaluation. On synthetic graphs, the GraphVAE\ourModelAcronym~ scores are superior to or  competitive with the BiGG and GRAN scores. On the real-world graphs, the GraphVAE\ourModelAcronym~ scores are    competitive with the BiGG and GRAN scores, and superior to the other benchmarks.

\begin{table}[h]
         \caption{Statistics-based comparison with benchmark GGMs. For a named evaluation graph statistic, each column reports the MMD between the test graphs  and the generated graphs. The best result is in bold and the second best is underlined.
  }
          \label{table:statMertics-SOTA}
    \begin{subtable}[h]{1\textwidth}
    \caption{{Synthetic Graphs} }
    \label{table:statMertics-SOTA-syn}
     \resizebox{\textwidth}{!}{
  \begin{tabular}{lccccccccccccccc}
    \toprule
    \multirow{2}{3.5em}{\textbf{Method}} &  \multicolumn{5}{c}{\textbf{Triangle Grid}} &
    \multicolumn{5}{c}{\textbf{Lobster}} &
    \multicolumn{5}{c}{\textbf{Grid}} 
    \\
    & \small{Deg.} & \small{Clus.}& \small{Orbit} & \small{Spect} & \small{Diam.} & \small{Deg.} & \small{Clus.} &  \small{Orbit} & \small{Spect}& \small{Diam.}& \small{Deg.} & \small{Clus.} &  \small{Orbit} & \small{Spect}& \small{Diam.} \\  

  \midrule

      50/50 split  &$3e^{-5}$  & $0.002$& $8e^{-5}$ & $0.004$ & $0.014$   & $0.002$ & $0$  & $0.002$ & $0.005$& $0.032$ &   $1e^{-5}$ & $0$  & $2e^{-5}$ & $0.004$ & $0.014$\\
    \midrule
    GraphVAE\ourModelAcronym  &$\textbf{0.001}$ & $\textbf{0.093}$& $\textbf{0.001}$& $\textbf{0.013}$& $\textbf{0.133}$    & $\textbf{2}e^{-4}$ & $\textbf{0}$ & $\underline{0.008}$ & $\underline{0.017}$ & $\underline{0.187}$ & $\textbf{5}e^\textbf{{-4}}$ & $\textbf{0}$ &  $\textbf{0.001}$ &  $\underline{0.014}$ & $\textbf{0.065}$ \\

    \midrule
     GraphRNN-S~\cite{you2018graphrnn}  &  $0.053$ & $1.094$ & $0.121$ & $0.033$ & $1.124$ & $0.016$ & $0.319$ & $0.285$ & $0.045$& $0.242$  &  $0.014$ & $0.003$ & $0.090$ & $0.112$ & $\underline{0.128}$ \\
    GraphRNN~\cite{you2018graphrnn}  & $\underline{0.033}$ & $1.167$ & $0.107$ & $0.030$ & $\underline{1.121}$  & $0.004$ & $\textbf{0}$ & $0.033$ & $0.035$ & $0.384$ &  $0.013$ & $0.166$ & $0.019$ & $0.111$ & $0.460$  \\
  GRAN~\cite{DBLP:conf/nips/LiaoLSWHDUZ19} & $0.134$ & $0.678$  &  $0.673$ & $0.184$ & $1.133$& $0.005$& $\underline{0.304}$  & $0.331$ & $0.043$ & $0.446$  &   $0.003$ & $1e^{-4}$ & $0.007$  & $\textbf{0.012}$ & $0.281$ \\
     BiGG~\cite{dai2020scalable}  &  $\textbf{0.001}$ & $\underline{0.107}$  & $\underline{0.004}$ &$\underline{0.020}$ & $1.123$  & $\underline{0.001}$  & $\textbf{0}$ & $\textbf{6}e^{-4}$ & $\textbf{0.012}$ & $\textbf{0.101}$ &$\underline{0.002}$ & $\underline{{3}e^{-5}}$ & $\underline{0.003}$ & $0.018$ & $0.328$ \\
    \bottomrule
  \end{tabular}%
  }  
    \end{subtable}
    \centering
    \begin{subtable}[h]{.7\textwidth}
      \caption{Real Graphs} 
      \label{table:statMertics-SOTA-real}
    \resizebox{\textwidth}{!}{
  \begin{tabular}{lccccccccccc}
    \toprule
    \multirow{2}{3.5em}{\textbf{Method}} &  \multicolumn{5}{c}{\textbf{Protein}} &
    \multicolumn{5}{c}{\textbf{ogbg-molbbbp}}
    \\
    & \small{Deg.} & \small{Clus.} & \small{Orbit} & \small{Spect} & \small{Diam.}& \small{Deg.} & \small{Clus.} &  \small{Orbit} & \small{Spect} & \small{Diam.}  \\  
   \midrule
       50/50 split  & $4e^{-5}$ & $0.004$ & $5e^{-4}$ & $4e^{-4}$  & $0.003$ & $2e^{-4}$ & $2e^{-5}$ & $9e^{-5}$ & $5e^{-4}$ & $0.002$ \\
    \midrule
  GraphVAE\ourModelAcronym    & $\underline{0.006}$ & $\textbf{0.059}$ & $\underline{0.152}$ & $\underline{0.007}$ & $\underline{0.091}$ & $\textbf{0.001}$ & ${0.005}$ & ${8e}^{-4}$ & $\textbf{0.005}$ & $\textbf{0.018}$\\
  \midrule
  GraphRNN-S~\cite{you2018graphrnn} & $0.046$ & $0.324$ & $0.316$ & $0.028$  & $0.302$  & $0.016$ & $0.572$ & $0.006$ & $0.045$  & $0.199$ \\
  GraphRNN~\cite{you2018graphrnn}& $0.012$ & $0.123$ & $0.264$ & $0.018$  & $0.342$ & $\underline{0.002}$& $\textbf{9}e^{-4}$ & $\underline{{4}e^{-4}}$ & $0.135$ & $0.495$ \\
    GRAN~\cite{DBLP:conf/nips/LiaoLSWHDUZ19}  & $\textbf{0.003}$ & $\textbf{0.059}$ & $\textbf{0.053}$ & $\textbf{0.004}$  & $\textbf{0.009}$ & $0.008$ & $0.353$ & $0.013$ & $0.056$  & $0.317$ \\
  BiGG~\cite{dai2020scalable}  & $0.007$  & $\underline{0.099}$ & $0.316$ & $0.012$ & $0.181$ &  $0.003$ & $\underline{0.001}$ & $\textbf{5}e^{-5}$ & $\underline{0.007}$  & $\underline{0.033}$ \\
    \bottomrule
  \end{tabular}%
  }%
     \end{subtable}
\end{table}

\subsection{Comparison of GGMs on train and generation time }
Table \ref{table:time-SOTA} compares benchmark GGMs and VGAEs on generation and train time for the main datasets. The benchmark GGMs require substantially more generation time than GraphVAEs. While MM modeling slows down training for the GraphVAE, the training time is still less than the benchmarks. 

\begin{table}[h]
  \caption{Comparison of benchmark GGMs on the {\em train and generation time}.  Train and Generation show average training time per epoch and average generation time per batch respectively.  The best result is in bold and the second best is underlined.
  }
  \label{table:time-SOTA}
  \centering
  \resizebox{\textwidth}{!}{
  \begin{tabular}{lcccccccccc}
    \toprule
    \multirow{2}{3.5em}{\textbf{Method}} &  \multicolumn{2}{c}{\textbf{Grid}} &
    \multicolumn{2}{c}{\textbf{Lobster}} &
    \multicolumn{2}{c}{\textbf{Triangle Grid}} &  \multicolumn{2}{c}{\textbf{Protein}} &
    \multicolumn{2}{c}{\textbf{ogbg-molbbbp}} \\
    & \small{{Train (s)}} & \small{Generation (s)}  & \small{{Train (s)}} & \small{Generation (s)}   & \small{{Train (s)}} & \small{Generation (s)}  & \small{{Train (s)}} & \small{Generation (s)}   & \small{{Train (s)}} & \small{Generation (s)}  \\
    \midrule
    GraphVAE  & $\textbf{0.28}$ & $\textbf{0.00}$  & $\textbf{0.11}$ & $\textbf{0.00}$ & $\textbf{0.24}$ & $\textbf{0.00}$ & ${\textbf{2.36}}$ & $\textbf{0.00}$ & $\textbf{1.11}$ & $\textbf{0.00}$ \\
    GraphVAE\ourModelAcronym  & $\underline{0.49}$ & $\textbf{0.00}$  & $\underline{0.15}$ & $\textbf{0.00}$& $\underline{0.40}$ & $\textbf{0.00}$  & $\underline{4.87}$ & $\textbf{0.00}$ &  $2.70$ & $\textbf{0.00}$ \\
  GraphRNN-S~\cite{you2018graphrnn} & $10.02$ & $32.20$  & $4.09$ & $24.21$ & $82.32$ & $12.62$ & $369.44$ & $110.23$ & $2.40$ & $37.92$\\
      GraphRNN~\cite{you2018graphrnn}& $16.16$ & $294.53$  & $4.66$ & $72.68$ & $296.76$ & $16.33$ & $864.1$ & $236.24$ & $\underline{2.02}$ & $30.89$ \\
    GRAN~\cite{DBLP:conf/nips/LiaoLSWHDUZ19} & $7.62$ & $22.13$  & $1.12$ & ${1.34}$ & $12.61$ & $29.27$ & ${9.51}$ & $44.19$ & $4.62$ & $37.60$ \\
        BiGG~\cite{dai2020scalable} & $97.75$ & $\underline{2.00}$  & $18.02$ & $\underline{0.31}$ & $82.86$ & $\underline{2.59}$ & ${130.28}$ & $\underline{2.93}$  & $23.14$ & $\underline{0.11}$\\
    \bottomrule
  \end{tabular}%
 } 
\end{table}

\subsection{Details on extended experiments}
\label{sec:ExtendedExp}
This section extends the experiments on real graph sets by evaluating the micro-macro modeling on MUTAG, PTC, and QM9 datasets.

\begin{itemize}
    \item MUTAG is a dataset of 188 mutagenic aromatic and heteroaromatic nitro compounds \cite{xu2018powerful}.
    \item PTC is a dataset of 344 chemical compounds that reports the carcinogenicity of male and female rats \cite{xu2018powerful}.
    \item QM9  is a large dataset containing about 134k organic molecules \cite{ramakrishnan2014quantum}.  
\end{itemize}
  
Following \cite{DBLP:conf/ai/ZahirniaSLSN22}  we withhold the edges/nodes labels to obtain homogenized datasets in which two nodes are connected if there is an edge between them in the original graph. Table \ref{table:new_MM} shows statistics-based and GNN-based evaluation of GraphVAE with and without  micro-macro modeling and comparison of micro-macro modeling on the train and generation time. Table \ref{table:new-GGM} compares GraphVAE-\ourModelAcronym~with benchmark GGMs. It was not feasible to train all auto-regressive benchmarks on QM9, so we report results only for the GraphVAE architecture with and without micro-macro modeling. 
 GraphVAE and GraphVAE\ourModelAcronym~are trained for $20,000$ epochs except for QM9, a large dataset with more than $134k$ graphs, which are trained for $100$ epochs. The experiments follow the setting in the main body of the paper, for details see section \ref{GraphVAE-architecture}.
 \paragraph{Impact on GraphVAE.}
Table \ref{table:new_MM} shows a very large improvement from micro-macro modeling for MMD RBF of generated graphs, especially for PTC the MMD RBF is reduced by more than one order of magnitude. The diversity of generated graphs, F1 PR, also substantially increased, The magnitude of  the increase is up to 53\%. For QM9, also the reality of generated graphs slightly improved. GraphVAE already had a strong performance on QM9, compared to the other datasets. To calculate the ideal score,  we used two randomly selected subsets of QM9 dataset, each including $5k$  graphs, because statistic-based evaluation metrics are slow to compute~\cite{thompson2022evaluation}. 

\begin{table}[h]
         \caption{Evaluation of micro-macro modeling for GraphVAEs on MUTAG, PTC, and QM9 datasets. 
  }
          \label{table:new_MM}
    \begin{subtable}[h]{1\textwidth}
    \caption{{GNN-based evaluation of micro-macro modeling.} }
    \label{table:new_MM_NN}
     \resizebox{\textwidth}{!}{
  \begin{tabular}{lcccccc}
    \toprule
    \multirow{2}{3.5em}{\textbf{Method}} &  \multicolumn{2}{c}{\textbf{MUTAG}} &
    \multicolumn{2}{c}{\textbf{PTC}} &
    \multicolumn{2}{c}{\textbf{QM9 }}   \\
    & \small{{MMD RBF}} & \small{F1 PR} & \small{{MMD RBF}} & \small{F1 PR} & \small{{MMD RBF}} & \small{F1 PR} \\
  \midrule
      50/50 split  & $0.03\pm0.00$ & $98.58\pm0.00$ & $0.04\pm0.00$ & $98.58\pm0.00$ & $0.010\pm0.00$ & $99.90\pm0.20$    \\
    \midrule
    GraphVAE  & ${0.09\pm0.02}$ & ${78.38}\pm{10.50}$  & ${0.53}\pm{0.13}$ & ${31.96}\pm{16.00}$ & ${0.024}\pm{0.01}$ & $\textbf{97.28}\pm\textbf{0.03}$ \\
    GraphVAE\ourModelAcronym  & $\textbf{0.07}\pm\textbf{0.01}$ & $\textbf{86.63}\pm\textbf{10.59}$  & $\textbf{0.04}\pm\textbf{0.01}$ & $\textbf{84.40}\pm\textbf{5.60}$ &
    $\textbf{0.019}\pm\textbf{0.00}$ & ${96.16}\pm{0.01}$ \\
    \bottomrule
  \end{tabular}%
  }  
    \end{subtable}
    \centering

\begin{subtable}[h]{1\textwidth}
    \caption{{Statistics-based evaluation of micro-macro modeling.} }
    \label{table:new_MM_stat}
     \resizebox{\textwidth}{!}{
  \begin{tabular}{lcccccccccccccccc}
    \toprule
    \multirow{2}{3.5em}{\textbf{Method}} &  \multicolumn{5}{c}{\textbf{MUTAG}} &
    \multicolumn{5}{c}{\textbf{PTC}} &  \multicolumn{5}{c}{\textbf{QM9}}
    \\
    & \small{Deg.} & \small{Clus.} & \small{Orbit} & \small{Spect} & \small{Diam.}& \small{Deg.} & \small{Clus.} &  \small{Orbit} & \small{Spect} & \small{Diam.} & \small{Deg.} & \small{Clus.} &  \small{Orbit} & \small{Spect} & \small{Diam.} \\  
   \midrule
       50/50 split  & $3e^{-4}$ & $0$ & $1e^{-5}$ & $0.005$  & $0.013$ & $1e^{-4}$ & $9e^{-5}$ & $8e^{-5}$ & $0.002$ & $0.013$ & $5e^{-5}$& $4e^{-4}$& $4e^{-4}$& $7e^{-5}$& $2e^{-6}$ \\
    \midrule
    GraphVAE   & ${0.005}$ & ${0.126}$ & ${0.003}$ & $\textbf{0.019}$ & ${0.055}$ & ${0.197}$ & ${0.757}$ & $0.562$ & ${0.036}$ & ${0.143}$ & ${0.007}$ & $\textbf{0.002}$ & $\textbf{0.003}$& ${0.004}$ &$0.005$ \\
  GraphVAE\ourModelAcronym    & $\textbf{0.001}$ & $\textbf{0}$ & $\textbf{1}e^{\textbf{-4}}$ & $\textbf{0.019}$ & $\textbf{0.015}$ & $\textbf{0.020}$ & $\textbf{3}e^{\textbf{-4}}$ & $\textbf{0.003}$ & $\textbf{0.018}$ & $\textbf{0.043}$ & $\textbf{0.005}$ & $\textbf{0.002}$ & $\textbf{0.003}$& $\textbf{0.003}$ &$\textbf{0.004}$\\
    \bottomrule
  \end{tabular}%
  }  
    \end{subtable}
    \begin{subtable}[h]{.8\textwidth}
    \caption{{Comparison of  micro-macro modeling on the train and generation time {\em per-epoch} and {\em per-batch} respectively} }
    \label{table:new_MM_time}
     \resizebox{\textwidth}{!}{
  \begin{tabular}{lcccccc}
    \toprule
    \multirow{2}{3.5em}{\textbf{Method}} &  \multicolumn{2}{c}{\textbf{MUTAG}} &
    \multicolumn{2}{c}{\textbf{PTC}} &
    \multicolumn{2}{c}{\textbf{QM9 }}   \\
    & \small{{Train (s)}} & \small{Generation (s)} & \small{{Train (s)}} & \small{Generation (s)} & \small{{Train (s)}} & \small{Generation (s)} \\
  \midrule

    GraphVAE  & $\textbf{0.07}$ & $\textbf{3e}^\textbf{{-4}}$  & $\textbf{0.23}$ & $\textbf{5e}^\textbf{-4}$  & $\textbf{0.38}$ & $\textbf{3e}^\textbf{-4}$\\
    GraphVAE\ourModelAcronym  & $0.15$ & $\textbf{3e}^\textbf{-4}$  & $0.32$ & $\textbf{5e}^\textbf{-4}$ & $0.75$ & $\textbf{3e}^\textbf{-4}$ \\
    \bottomrule
  \end{tabular}%
  }  
    \end{subtable}
\end{table}

\paragraph{GraphVAE\ourModelAcronym~vs. Benchmark GGMs.} As table \ref{table:new-GGM-NN} shows, GraphVAE\ourModelAcronym~beats the baselines on the MMD RBF,  and is very competitive on the F1 PR, though worse than BiGG. In addition, in our experiments, GraphVAE\ourModelAcronym~is much faster than the auto-regressive baselines in generation time, see \ref{table:new-GGM-time}.
\begin{table}[h]
         \caption{Micro-macro modeling comparison with benchmark GGMs on MUTAG, and PTC datasets. The benchmark methods were infeasible on QM9.  The best result is in bold and the second best is underlined.
  }
          \label{table:new-GGM}
    \begin{subtable}[h]{.8\textwidth}
    \caption{GNN-based comparison of micro-macro modeling with benchmark GGMs.}
    \label{table:new-GGM-NN}
     \resizebox{\textwidth}{!}{
  \begin{tabular}{lcccc}
    \toprule
    \multirow{2}{3.5em}{\textbf{Method}} &  \multicolumn{2}{c}{\textbf{MUTAG}} &
    \multicolumn{2}{c}{\textbf{PTC}}    \\
    & \small{{MMD RBF}} & \small{F1 PR} & \small{{MMD RBF}} & \small{F1 PR}  \\
  \midrule
      50/50 split  & $0.03\pm0.00$ & $98.58\pm0.00$ & $0.04\pm0.00$ & $98.58\pm0.00$ \\
    \midrule

    GraphVAE\ourModelAcronym  & $\textbf{0.07}\pm\textbf{0.01}$ & ${86.63}\pm{10.59}$  & $\textbf{0.04}\pm\textbf{0.01}$ & $\underline{84.40\pm{5.60}}$   \\
  \midrule 
  GraphRNN-S~\cite{you2018graphrnn} & $0.83\pm0.14$ & $55.25\pm22.62$ & $0.66\pm0.15$ & $34.50\pm18.12$ \\
      GraphRNN~\cite{you2018graphrnn} &$1.64\pm0.05$ & $0.99\pm0.00$&  $0.88\pm0.15$ & $32.26\pm0.05$   \\
    GRAN~\cite{DBLP:conf/nips/LiaoLSWHDUZ19} &  $\underline{0.29\pm0.08}$ & $\underline{93.24\pm3.68}$ & $\underline{0.17\pm0.02}$ & $81.20\pm7.14$  \\
        BiGG~\cite{dai2020scalable} & ${0.56\pm0.00}$ & $\textbf{98.08}\pm\textbf{0.00}$  & $\textbf{0.04}\pm\textbf{0.00}$ & $\textbf{98.11}\pm\textbf{1.66}$  
        \\
    \bottomrule
  \end{tabular}%
  }  
    \end{subtable}
    \centering
        \begin{subtable}[h]{\textwidth}
        \caption{Statistics-based comparison of micro-macro modeling with benchmark GGMs.}
    \resizebox{\textwidth}{!}{
  \begin{tabular}{lccccccccccc}
    \toprule
    \multirow{2}{3.5em}{\textbf{Method}} &  \multicolumn{5}{c}{\textbf{MUTAG}} &
    \multicolumn{5}{c}{\textbf{PTC}}
    \\
    & \small{Deg.} & \small{Clus.} & \small{Orbit} & \small{Spect} & \small{Diam.}& \small{Deg.} & \small{Clus.} &  \small{Orbit} & \small{Spect} & \small{Diam.}  \\  
   \midrule
       50/50 split  & $3e^{-4}$ & $0$ & $1e^{-5}$ & $0.005$  & $0.013$ & $1e^{-4}$ & $9e^{-5}$ & $8e^{-5}$ & $0.002$ & $0.013$ \\
    \midrule
  GraphVAE\ourModelAcronym    & $\underline{0.001}$ & $\textbf{0}$ & $\textbf{1e}^{\textbf{-4}}$ & $\textbf{0.019}$ & $\textbf{0.015}$ & ${0.020}$ & $\textbf{3e}^{\textbf{-4}}$ & $0.003$ & $\underline{0.018}$ & $\underline{0.043}$\\
  \midrule
  GraphRNN-S~\cite{you2018graphrnn} & $0.006$ & $\underline{5e^{-4}}$ & $0.002$ & $0.105$  & $1.157$  & $0.022$ & $0.254$ & $0.035$ & $0.057$  & $0.270$ \\
  GraphRNN~\cite{you2018graphrnn}& $0.006$ & $0.210$ & ${\underline{{8}e^{-4}}}$ & $0.070$  & $0.819$ & $\underline{0.005}$& $0.003$ & $\underline{0.002}$ & $0.075$ & $0.397$ \\
    GRAN~\cite{DBLP:conf/nips/LiaoLSWHDUZ19}  & $\textbf{6}e^{-4}$ & $0.015$ & ${0.007}$ & ${0.053}$  & ${0.685}$ & $0.013$ & $0.137$ & $0.006$ & $0.034$  & $0.194$ \\
  BiGG~\cite{dai2020scalable}  & $0.004$  & $\textbf{0}$ & $0.002$ & $\underline{0.040}$ & $\underline{0.293}$ &  $\textbf{1e}^\textbf{-4}$ & $\underline{0.002}$ & $\textbf{3e}^{-5}$ & $\textbf{0.016}$  & $\textbf{0.015}$ \\
    \bottomrule
  \end{tabular}%
  }%
     \end{subtable}
\begin{subtable}[h]{1\textwidth}
    \caption{Comparison of micro-macro modeling on the train and generation time, {\em per-epoch} and {\em per-batch} respectively, with benchmark GGMs.}
    \label{table:new-GGM-time}
    \centering
     \resizebox{.7\textwidth}{!}{
  \begin{tabular}{lcccc}
    \toprule
    \multirow{2}{3.5em}{\textbf{Method}} &  \multicolumn{2}{c}{\textbf{MUTAG}} &
    \multicolumn{2}{c}{\textbf{PTC}}    \\
    & \small{{Train (s)}} & \small{Generation (s)} & \small{{Train (s)}} & \small{Generation (s)} \\
  \midrule

    GraphVAE\ourModelAcronym  & $\textbf{0.15}$ & $\textbf{3e}^\textbf{-4}$  & $\textbf{0.32}$ & $\textbf{5e}^\textbf{-4}$  \\
  \midrule 
  GraphRNN-S~\cite{you2018graphrnn} & $1.18$ & ${5.77}$  & $2.12$ & $21.62$  \\
      GraphRNN~\cite{you2018graphrnn} & $1.38$ & $5.97$  & $2.08$ & $26.15$  \\
    GRAN~\cite{DBLP:conf/nips/LiaoLSWHDUZ19} & $\underline{0.88}$ & $24.63$  & $\underline{0.61}$ & $35.58$ \\ 
        BiGG~\cite{dai2020scalable} & $5.20$ & $\underline{0.08}$  & $7.66$ & $\underline{0.07}$   \\
    \bottomrule
  \end{tabular}%
  }  
    \end{subtable}

\end{table}

\subsection{System architecture}
\label{sec:SystemArchitecture}
The code for all models is run on the same system, an Intel(R) Core(TM) i9-9820X CPU 3.30GHz and Nvidia TITAN RTX GPU with TU102-core. Because of package compatibility issues, GraphRNN(-S) is run on an Intel(R) Core(TM) i7-5820K CPU 3.30GHz and a GM200 GeForce GTX TITAN X. 

\subsection{Complexity bounds}
Table~\ref{table:costOFcomponents}  gives complexity bounds for the different default statistics we use in this paper. Table~\ref{table:SOTA-Complexity} presents computational complexity of GGMs.
The bounds are based on the literature and the analysis presented in this paper.

\begin{table}[h]
  \caption{Complexity of MM-\elbo Components}
  \label{table:costOFcomponents}
  \centering
  \resizebox{11cm}{!}{
  \begin{tabular}{lccc}
    \toprule
   Component & Time Complexity &  Space Complexity & Property \\
      \midrule 
    Edge Reconstruction Probability & $O(\nodeNum^2)$ & $O(\nodeNum^2)$ &   Permutation Equivariant\\
        \midrule 
   Triangle Count & $O(\nodeNum^3)$ & $O(\nodeNum^2)$  & Permutation Invariant\\
   Degree histogram & $O(\nodeNum^2)$ & $O(\nodeNum^2)$  &  Permutation Invariant\\
   S-Step  transition probability & $O(\nodeNum^3)$ & $O(\nodeNum^2)$ &  Permutation Equivariant\\
 
    \bottomrule
  \end{tabular}%
 } 
\end{table}

\begin{table}[h]
  \caption{Graph Generative Models complexity comparison}
  \label{table:SOTA-Complexity}
  \centering
  \resizebox{14cm}{!}{
  \begin{tabular}{lcccc}
    \toprule
   Model & Train (Computational Complexity) &  Graph Generation (Computational Complexity) & Auto-Regressive Decision Steps   \\
      \midrule 
   GraphVAE~\cite{DBLP:conf/icann/SimonovskyK18} & $O(\nodeNum^4)$ & $O(\nodeNum^2)$& $O(1)$ \\
      GraphVAE & $O(\nodeNum^2)$ & $O(\nodeNum^2)$& $O(1)$ & \\
   GraphVAE\ourModelAcronym & $O(\nodeNum^3)$ & $O(\nodeNum^2)$& $O(1)$  \\
   BiGG~\cite{dai2020scalable} & $O(min((|E|+\nodeNum)\log\nodeNum, ~\nodeNum^2))$& $O(min((|E|+\nodeNum)\log\nodeNum, ~\nodeNum^2))$& $\log \numnodes$ \\
   GRAN~\cite{DBLP:conf/nips/LiaoLSWHDUZ19} & $O(\nodeNum^2)$ & $O(\nodeNum^2)$ &$\nodeNum$ \\
   GraphRNN~\cite{you2018graphrnn} & $O(\nodeNum^2)$ &$O(\nodeNum^2)$ &$|E|\nodeNum$ \\
    \bottomrule
  \end{tabular}%
 } 
\end{table}

\subsection{Societal impact}
\label{sec:SocietalImpact}
Graph generation could have both positive and negative Societal impact, depending on the application domain. On the positive side, graphs can represent molecules, and graph modeling supports medical discovery. On the negative side, network analysis ,as a field, has potential to increase and misuse control over network participants. For example, to motivate surveillance violations of privacy in targeting recommendations, or identify users through their social links. However, these harms can be mitigated by strengthening privacy protections during data collection. Furthermore, network analysis can provide significant societal benefit, for example, by highlighting the existence and situation of marginalized communities and understanding the flow of influence and (mis)information in social networks. The main contribution of our work supports more beneficial uses by enhancing the understanding of global network structure, as opposed to surveillance and the potentially harmful targeting of individuals. 

\subsection{Code overview}
The implementation is provided at  \url{https://github.com/kiarashza/GraphVAE-MM}. main.py includes the training pipeline and also micro-macro  objective functions implementation. 
Source codes for loading real graph datasets and generating synthetic graphs are included in data.py. All the Python packages used in our experiments are provided in environment.yml.
Generated graph samples for each of the datasets are provided in the "ReportedResult/" directory, both in the pickle and png format. This directory also includes the log files and hyperparameters details used to train the GraphVAE\ourModelAcronym~on each of the datasets.

\end{document}